\documentclass[10pt,twocolumn,letterpaper]{article}

\usepackage{wacv}              
\usepackage{amsmath}
\usepackage{graphicx}
\usepackage{caption}
\usepackage{subcaption}
\usepackage{multirow}
\usepackage{makecell}
\usepackage{boldline}
\usepackage{amssymb}
\usepackage{pifont}
\usepackage{dsfont}
\usepackage{wrapfig}
\usepackage{algorithm}
\usepackage{algorithmic}

\usepackage{tikz}
\usepackage{pgfplots}
\pgfplotsset{compat=1.18}
\usepackage{subcaption}

\definecolor{mypink}{rgb}{0.858, 0.188, 0.478}

\definecolor{wacvblue}{rgb}{0.21,0.49,0.74}
\usepackage[pagebackref,breaklinks,colorlinks,allcolors=wacvblue]{hyperref}

\def\wacvPaperID{2994} 
\def\confName{WACV}
\def\confYear{2026}

\title{Stabilizing Direct Training of Spiking Neural Networks: Membrane Potential Initialization and Threshold-robust Surrogate Gradient}

\author{
Hyunho Kook\quad
Byeongho Yu\quad
Jeong Min Oh\quad
Eunhyeok Park\\
Pohang University of Science and Technology (POSTECH), Republic of Korea\\
{\tt\small \{kookhh0827, bhyu418, ojm010130, eh.park\}@postech.ac.kr}
}

\begin{document}
\maketitle
\begin{abstract}
Recent advancements in the direct training of Spiking Neural Networks (SNNs) have demonstrated high-quality outputs even at early timesteps, paving the way for novel energy-efficient AI paradigms. However, the inherent non-linearity and temporal dependencies in SNNs introduce persistent challenges, such as temporal covariate shift (TCS) and unstable gradient flow with learnable neuron thresholds. In this paper, we present two key innovations: \textbf{MP-Init (Membrane Potential Initialization)} and \textbf{TrSG (Threshold-robust Surrogate Gradient)}. MP-Init addresses TCS by aligning the initial membrane potential with its stationary distribution, while TrSG stabilizes gradient flow with respect to threshold voltage during training. Extensive experiments validate our approach, achieving state-of-the-art accuracy on both static and dynamic image datasets. The code is available at: \href{https://github.com/kookhh0827/SNN-MP-Init-TRSG}{\textcolor{mypink}{https://github.com/kookhh0827/SNN-MP-Init-TRSG}}
\end{abstract}
    
\section{Introduction}
\label{sec:introduction}


Spiking Neural Networks (SNNs) transmit information through discrete spikes over time, enabling efficient, event-driven processing~\cite{cao:2015, esser:2016, neftci:2018, roy:2019, cramer:2022, rathi:2023}, especially on neuromorphic hardware~\cite{merolla:2014, akopyan:2015, davies:2018}. Recent advances in direct training~\cite{wu:2018, wu:2019} have significantly improved SNN accuracy with few timesteps, enhancing efficiency. Yet, SNNs still lag behind deep neural networks, and their complex dynamics often hinder performance on large-scale datasets~\cite{review_guo:2023, review_zhou:2024}. Thus, further innovations are crucial to ensure reliability and accuracy for real-world adoption.

One of the primary challenges in SNN training is \textbf{Temporal Covariate Shift (TCS)}~\cite{bntt:2020}, where internal covariate shifts in activations accumulate over time. While TEBN~\cite{tebn:2022} and TAB~\cite{tab:2024} attempt to address this issue by introducing timestep-specific normalization parameters, they disrupt parameter homogeneity across time, increasing both model complexity and computational cost. More importantly, our analysis reveals that these methods fail to mitigate TCS in other critical components of SNNs, particularly the membrane potential. Since the membrane potential directly determines neuronal output, neglecting TCS in this component can lead to significant accuracy degradation.

The second challenge lies in \textbf{training the internal parameters}~\cite{fang:2021, wang:2022, rathi:2023} of Leaky Integrate-and-Fire (LIF) neurons, which are the threshold voltage ($V_{\text{thr}}$) and leakage term ($\tau$). Despite their crucial role in LIF neuron behavior, they are typically fixed or updated via gradient descent under strict constraints. We reveal that existing gradient flow heavily depends on the scale of $V_{\text{thr}}$, and unconstrained updates during training could lead to significant instability. Surprisingly, none of the existing studies has explicitly recognized this issue.

In this work, we introduce a comprehensive approach to overcoming these limitations. Our first contribution, \textbf{MP-Init (Membrane Potential Initialization)}, stems from our analysis of the root cause of TCS in SNNs. By addressing TCS in membrane dynamics, MP-Init ensures consistent neuron behavior across timesteps without using timestep-dependent parameters. Our second contribution, \textbf{TrSG (Threshold-Robust Surrogate Gradient)}, is a refined version of surrogate gradient that stabilizes the gradient flow with respect to the threshold $V_{\text{thr}}$. This modification allows for stable gradient-based updates of all internal parameters, leading to superior convergence.

We provide a rigorous mathematical analysis of our methods and demonstrate through extensive experiments that they significantly improve SNN performance. Our approach enables effective SNN training, achieving state-of-the-art results on both static and dynamic datasets.
\section{Related Work}
\label{sec:related}

\subsection{Addressing Temporal Covariate Shift}
The TCS problem was first identified in BNTT~\cite{bntt:2020} and later addressed by methods such as TEBN~\cite{tebn:2022} and TAB~\cite{tab:2024}, which adapt normalization layers on a per-timestep basis. While these methods alleviate TCS in activations, they increase model complexity with additional parameters and, more critically, overlook the persistent shift in membrane potential, a primary source of TCS during inference (see Section~\ref{sec:root}). 

IMP+LTS~\cite{shen:2024} proposed training the initial membrane potential for every neuron to enhance representational power. However, it does not directly address TCS and incurs a large parameter overhead scaling with the number of neurons $O(N)$. By contrast, our approach initializes membrane potentials at the layer level $O(L)$ ($L \ll N$), directly targeting TCS while remaining lightweight.

More recently, MPS~\cite{ding:2025} reinterpreted SNNs from an ensemble-learning perspective, arguing that excessive differences in membrane potential distributions across timesteps destabilize subnetworks. Their smoothing and guidance strategies improve temporal consistency and yield accuracy gains. However, these methods modify the LIF update rule itself and require additional smoothing coefficients and guidance losses, meaning both training and inference diverge from the standard LIF neurons. In contrast, our approach preserves the original inference pipeline: MP-Init adds negligible overhead during training while inference remains compatible with neuromorphic hardware~\cite{davies:2018}.


\subsection{Learning in Spiking Neural Networks}
Surrogate Gradient (SG) descent enables end-to-end training of SNNs by approximating the non-differentiable firing process~\cite{lee:2016, wu:2018, wu:2019}. Some studies primarily focus on the impact of hyperparameters and the shape of surrogate derivatives~\cite{zenke:2021, li:2021}, while recent advancements introduce differentiable surrogate gradient functions~\cite{li:2021} and Temporal Efficient Training (TET) losses~\cite{deng:2022}. Other approaches regulate membrane potential by adding loss terms~\cite{guo:2022, guo:2023} or applying batch normalization~\cite{guo:2023_2}. Meanwhile, methods that modify the dynamics of LIF neurons~\cite{fang:2023, yao:2022, huang:2024} have been explored, though they could be incompatible with existing neuromorphic hardware~\cite{hunsberger:2015, review_guo:2023}.

Recent studies have also explored training internal parameters, such as the threshold voltage and leakage term, via gradient descent~\cite{fang:2021, wang:2022, rathi:2023}. For instance, PLIF~\cite{fang:2021} optimizes the leakage term, LTMD~\cite{wang:2022} adjusts the threshold voltage, and DIET-SNN~\cite{rathi:2023} learns both. However, these works provide limited analysis of gradient flow related to these learnable parameters, leaving questions about training stability unanswered. In contrast, we find that when the threshold voltage becomes trainable, surrogate gradient methods can exhibit instability, leading to misaligned, exploding, or vanishing gradients, as discussed in Section~\ref{sec:type1_type2}. To overcome these issues, we propose a novel SG formulation that remains stable when learning the threshold voltage, ensuring robust convergence.  

\section{Preliminaries}
\label{sec:preliminary}

\subsection{The LIF Neuron Model}

LIF model~\cite{eshraghian:2021} is a well-known spiking neural model that emulates key behaviors of biological neurons, including firing spikes and decaying membrane potential. The LIF neuron’s behavior in discrete steps \(t \in \mathbb{N}\) is described as the following equations:

\begin{equation}
\label{eq:lif_discrete_potential}
    M^{l}[t] = \left( 1 - \frac{1}{\tau^{l}} \right) U^{l}[t-1] + \frac{1}{\tau^{l}} I^{l}_{\text{in}}[t],
\end{equation}

\begin{equation}
\label{eq:lif_discrete_firing}
    S^{l}[t] = \mathcal{H}\left( M^{l}[t] - V^{l}_{\text{thr}} \right),
\end{equation}

\begin{equation}
\label{eq:lif_discrete_reset}
    U^{l}[t] = \begin{cases}
        M^{l}[t] - V^{l}_{\text{thr}} \cdot S^{l}[t], & \text{(soft reset)} \\
        M^{l}[t] \cdot (1 - S^{l}[t]). & \text{(hard reset)}
\end{cases}
\end{equation}

In these equations, \( l \) represents the layer index within the network, and \(\mathcal{H}\) denotes the Heaviside step function. The presynaptic input current is defined as \( I^{l}_{\text{in}}[t] = W_{l-1}^{l} O^{l-1}[t] \), where \( W_{l-1}^{l} \) is the synaptic weight matrix connecting layer \( l-1 \) to layer \( l \), and \( O^{l-1}[t] \) represents the output of the previous layer. In direct training, this output typically corresponds to the binary spike itself, i.e., \( O^{l}[t] = S^{l}[t] \)~\cite{wu:2018}. The leakage constant \(\tau^{l}\) can be absorbed into the synaptic weights~\cite{eshraghian:2021}, leading to a rescaling \( W_{l-1}^{l}:=\frac{1}{\tau^{l}}W_{l-1}^{l} \) in \cref{eq:lif_discrete_potential}. The choice between a soft and hard reset in Eq.~\ref{eq:lif_discrete_reset} influences the membrane potential dynamics after firing. Under a soft reset~\cite{rueckauer:2016}, the membrane potential decreases by the threshold voltage \(V^{l}_{\text{thr}}\). In contrast, a hard reset~\cite{wu:2018} resets the membrane potential to zero. Notably, our two novel ideas are effective in both cases, leading to improved quality.

\subsection{Exclusion of Silent Neurons}
In our analysis of TCS, we focus only on \emph{active neurons}, i.e., those that fire at least once during each simulation. Previous work~\cite{yin:2022} shows that a large portion of neurons in SNNs remain subthreshold throughout a simulation (i.e., \emph{silent}). They are irrelevant to our analysis, as they neither contribute to information propagation nor exhibit statistical properties similar to those of active neurons. Consequently, we exclude silent neurons when analyzing the statistics of the membrane potential.
\section{Membrane Potential Initialization}
\label{sec:first_method}

TCS~\cite{bntt:2020, tebn:2022, tab:2024} limits the performance of SNNs during inference, particularly when batch normalization (BN) is employed. Despite BN's capability to normalize activations during training, the fixed normalization statistics during inference result in activation drift across timesteps, causing accuracy degradation. Previous methods~\cite{tebn:2022, tab:2024} have tried to address this issue through timestep-dependent normalization, but these approaches remain superficial as they fail to resolve the underlying cause: drift in the membrane potential of LIF neurons.

To tackle this issue, we propose \textbf{MP-Init (Membrane Potential Initialization)}, an efficient solution that directly targets TCS in the membrane potential. MP-Init ensures stable activation distributions throughout inference by initializing the membrane potential of spiking neurons in each layer to match their stationary distribution. This eliminates the need for timestep-dependent normalization and enhances final performance by aligning training and inference statistics.

\begin{figure}[t!]
  \centering
  \begin{subfigure}{0.495\linewidth}
    \includegraphics[width=0.95\linewidth]{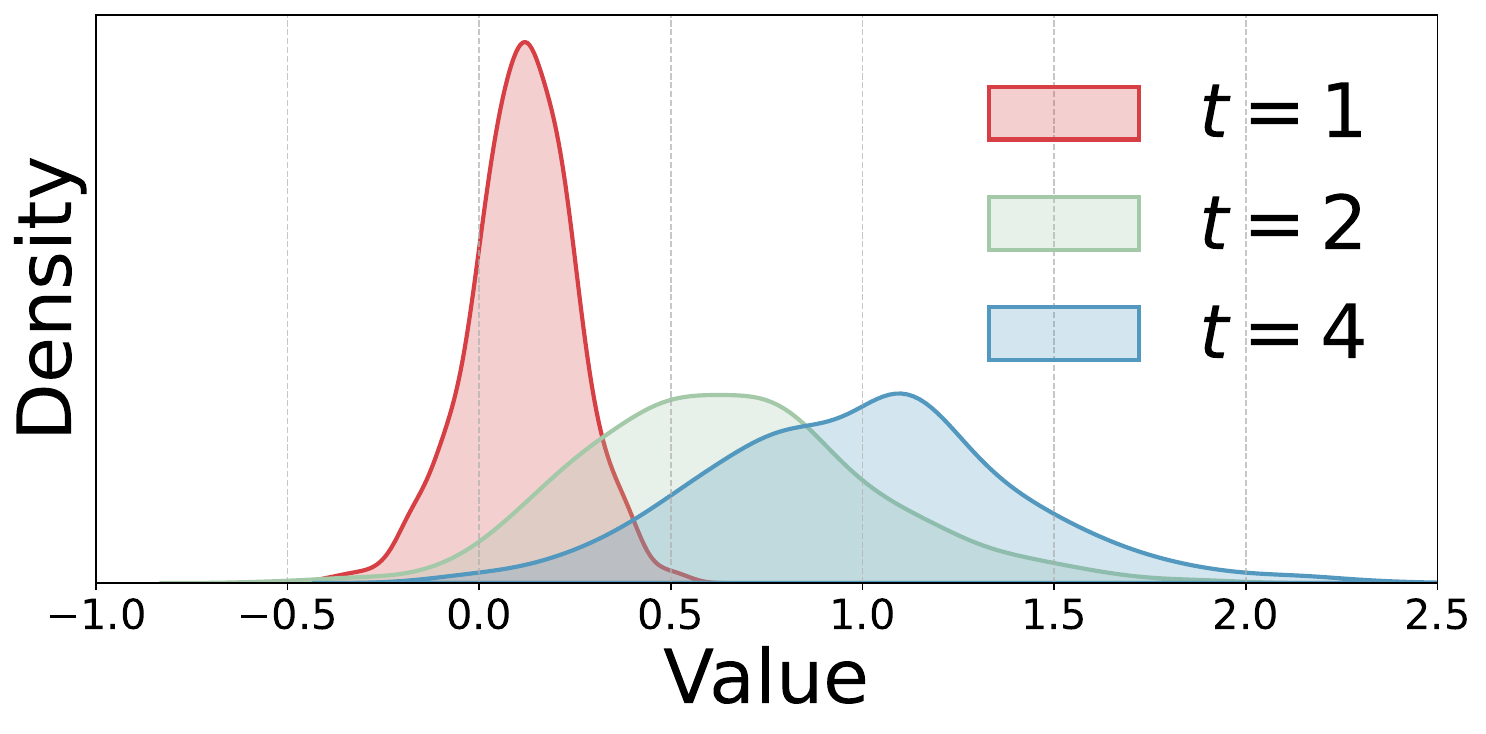}
    \caption{tdBN}
    \label{fig:membrane_potential_a}
  \end{subfigure}
  \hfill
  \begin{subfigure}{0.495\linewidth}
    \includegraphics[width=0.95\linewidth]{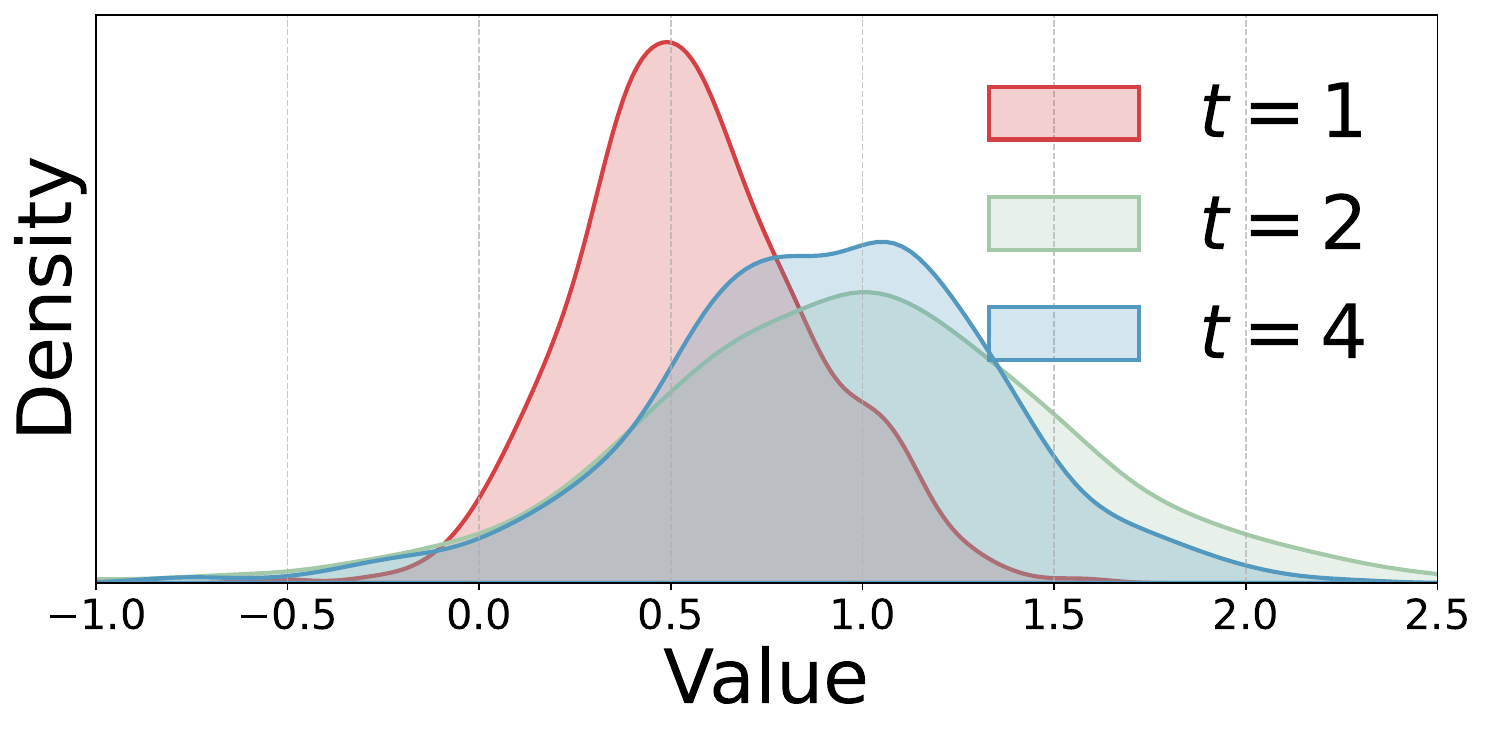}
    \caption{TEBN}
  \end{subfigure}
  \begin{subfigure}{0.495\linewidth}
    \includegraphics[width=0.95\linewidth]{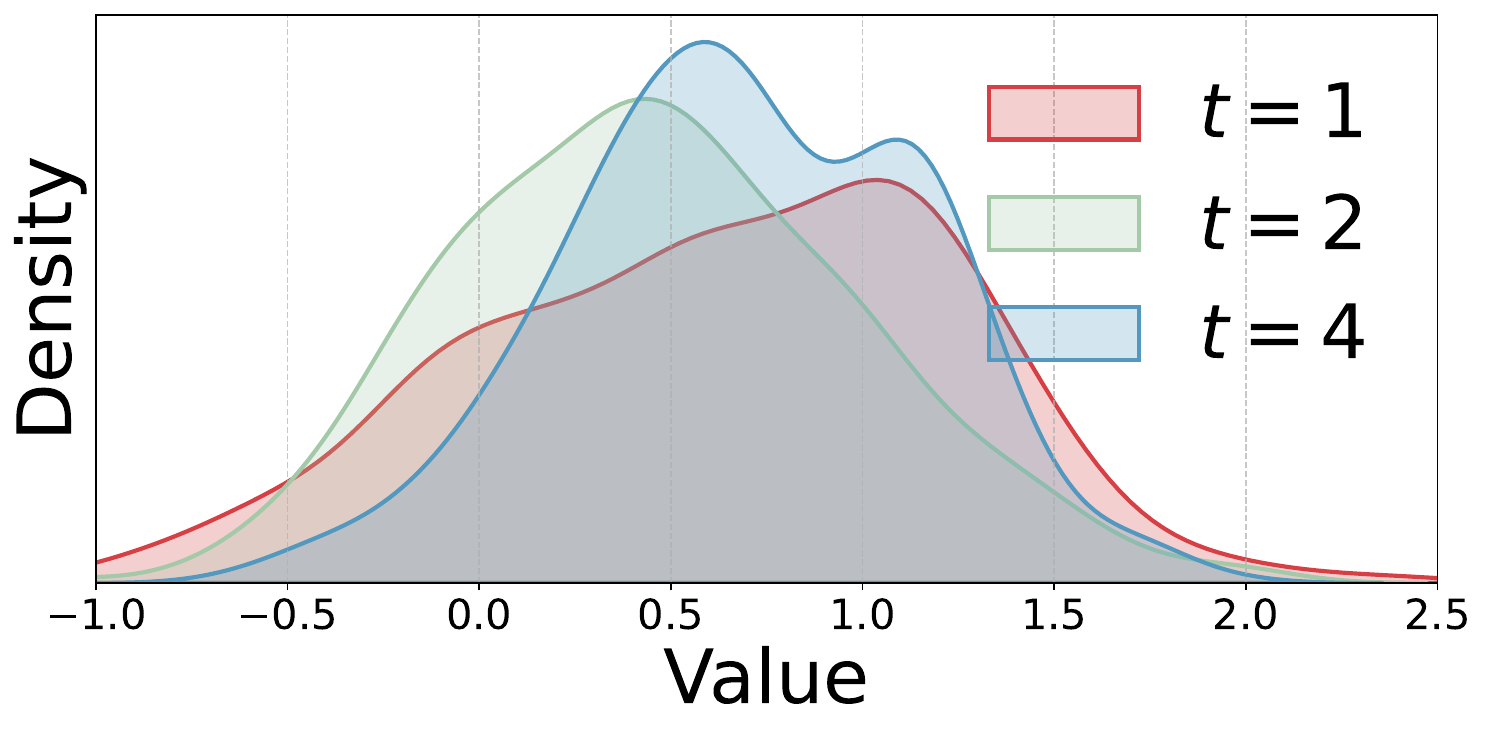}
    \caption{TAB}
  \end{subfigure}
  \hfill
  \begin{subfigure}{0.495\linewidth}
    \includegraphics[width=0.95\linewidth]{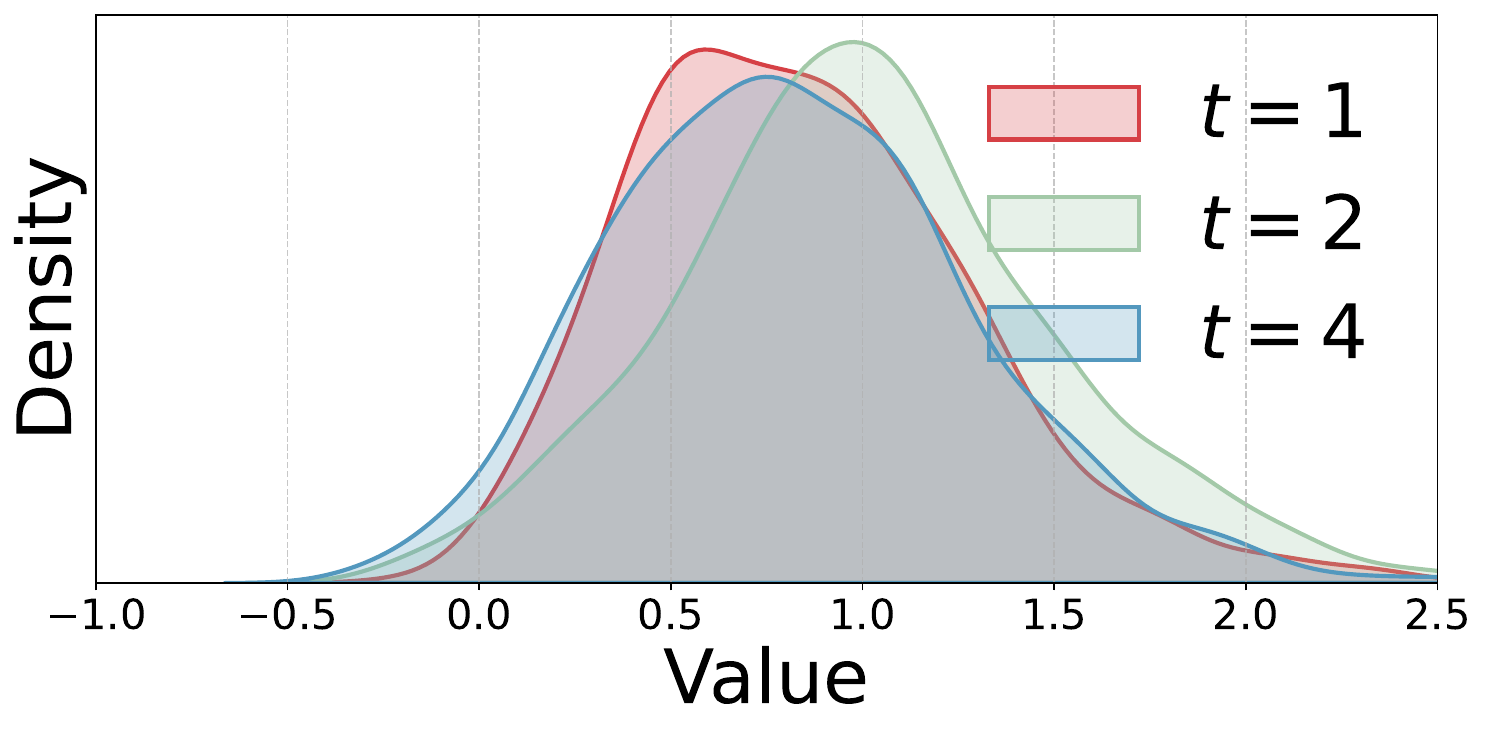}
    \caption{MP-Init}
  \end{subfigure}
  \caption{Membrane potential distribution of the third layer's first spiking layer of ResNet-19 on CIFAR100}
  \label{fig:combined_membrane}
\end{figure}

\begin{figure}[t!]
  \centering
  \begin{subfigure}{0.495\linewidth}
    \includegraphics[width=0.95\linewidth]{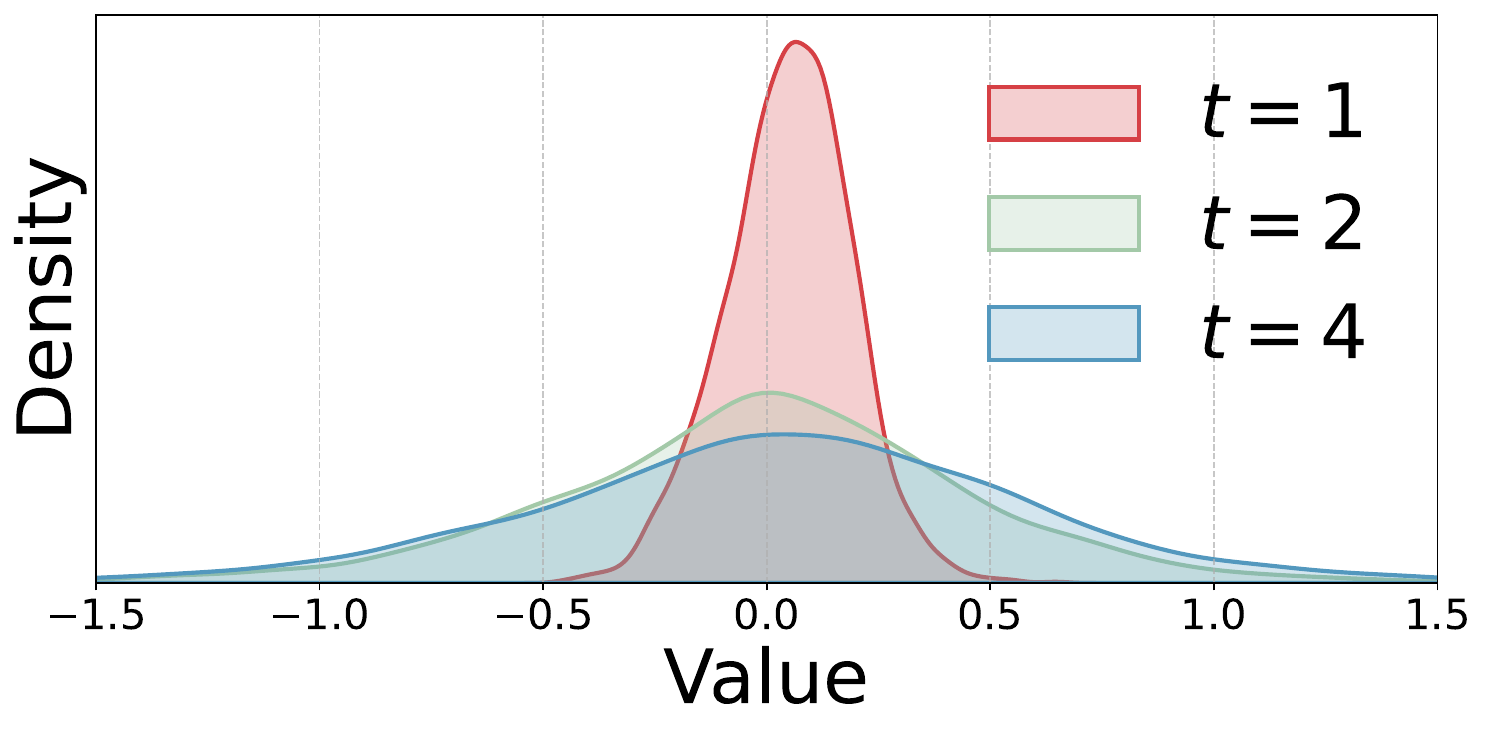}
    \caption{tdBN}
    \label{fig:presynaptic_input_a}
  \end{subfigure}
  \hfill
  \begin{subfigure}{0.495\linewidth}
    \includegraphics[width=0.95\linewidth]{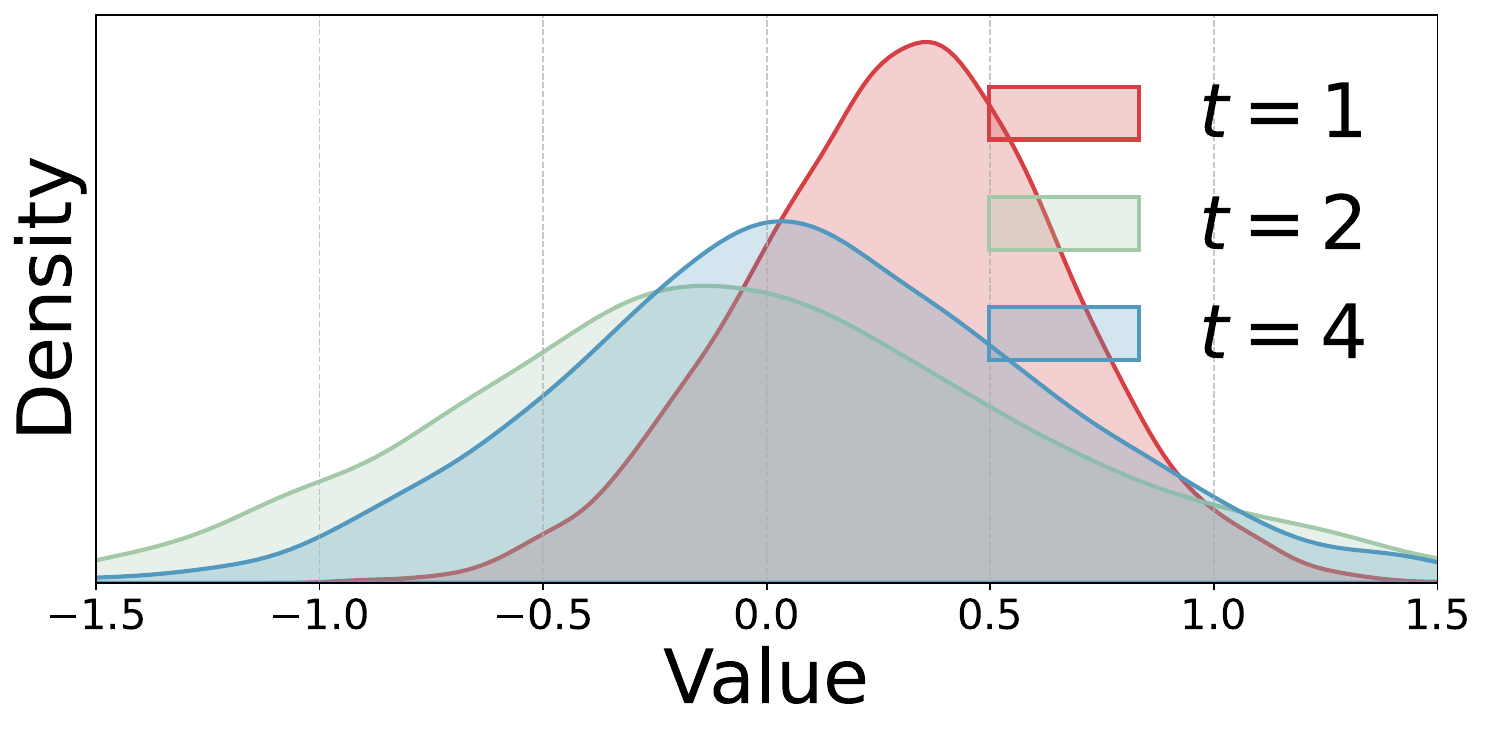}
    \caption{TEBN}
  \end{subfigure}
  \begin{subfigure}{0.495\linewidth}
    \includegraphics[width=0.95\linewidth]{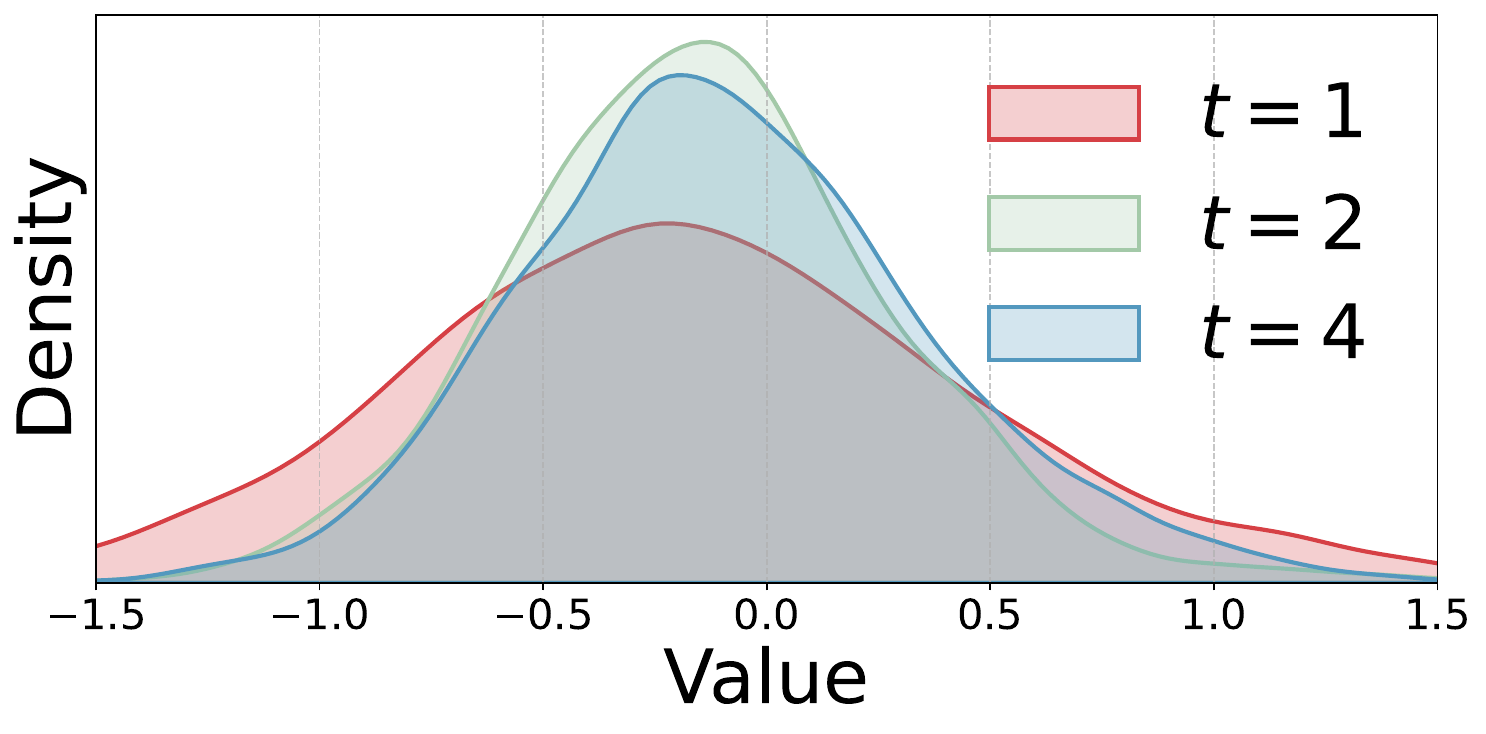}
    \caption{TAB}
  \end{subfigure}
  \hfill
  \begin{subfigure}{0.495\linewidth}
    \includegraphics[width=0.95\linewidth]{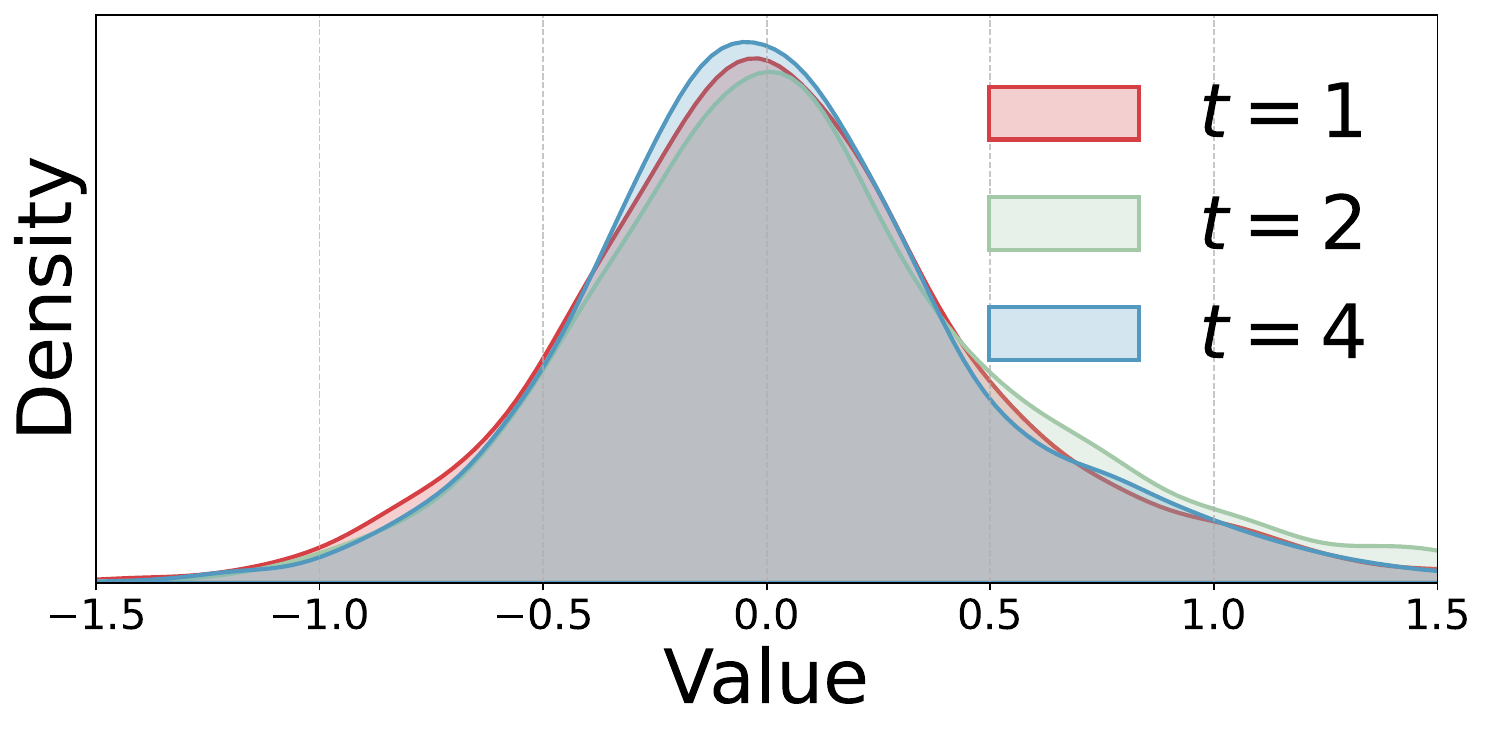}
    \caption{MP-Init}
  \end{subfigure}
  \caption{Activation distribution of the convolutional layer after the third layer's first spiking layer of ResNet-19 on CIFAR100}
  \label{fig:combined_activation}
\end{figure}

\subsection{Root Cause of TCS}\label{sec:root}

We first identify that TCS in activations fundamentally originates from TCS in membrane potentials. According to \cref{eq:lif_discrete_firing}, spike outputs depend directly on membrane potentials. If membrane potentials significantly drift over timesteps, the resulting activation distribution shifts correspondingly.

We empirically visualize the membrane potential at each timestep to validate this claim. As illustrated in \cref{fig:combined_membrane}, the membrane potential in tdBN, which does not account for the TCS problem, exhibits a clear temporal drift, leading to TCS in activation (\cref{fig:combined_activation}). While TEBN and TAB attempt to stabilize the distribution by timestep-dependent normalizations, the membrane potential itself continues to undergo transient drift, preserving TCS in activation.

\subsection{Convergence of Membrane Potential}
\label{sec:first_method_convergence}
While we have empirically demonstrated the existence of TCS in the membrane potential, a theoretical perspective can provide a deeper understanding. In this section, we present the core principles explaining why the membrane potential drifts over time.

We prove that the layer-wise distribution of membrane potential converges to a distribution possibly differing from its initial state as time progresses. This was proven on continuous-time LIF models with stable input statistics~\cite{dumont:2017}. Here, we extend this idea to the discrete-time model and show it has the same property. Let's start with the following assumption:

\textbf{Assumption 1.}
Assume $\{I_{\text{in}}[t]\}_{t\ge1}$ be a sequence drawn from an i.i.d. distribution. Furthermore, suppose there exist real constants $u^- < u^+$ where the membrane potential $U[t]$ of active neurons within each layer remains within $[u^-, u^+]$ for all $t$.

This i.i.d.\ assumption is reasonable, as inputs ${I_{\text{in}}[t]}_{t\ge1}$ are repeatedly drawn from a dataset or a normalized layer. Although soft reset LIF dynamics do not strictly prevent unbounded growth, well-trained SNNs rarely exhibit it in active neurons. The empirical validation of this assumption is provided in the supplementary material (\cref{subsec:validation_of_assumption_1}). Based on Assumption 1, we present the following theorem:

\textbf{Theorem 1.}
Under Assumption 1, the sequence $\{U[t]\}_{t \ge 0}$ forms a Markov chain with transition kernel $P$. This chain satisfies Doeblin’s minorization condition, ensuring the existence of a unique stationary distribution $\pi$. Moreover, the Total Variation (TV) distance between $U[t]$ and $\pi$ decreases exponentially.

Detailed proof is provided in supplementary material (\cref{subsec:proof_of_theorem_1}). This theorem indicates that TCS in the membrane potential is inevitable when the initial membrane potential differs from its stationary distribution. Traditionally, the membrane potential is initialized to zero, i.e., \(U[0] = 0\), but this zero initialization is a primary contributor to TCS. Therefore, to mitigate TCS, we must reduce the difference between the initial membrane potential and its stationary distribution.

\subsection{Membrane Potential Initialization}

Theorem 1 implies that eliminating the mismatch between the initial state and this stationary distribution is crucial. However, directly computing \(\pi\) is intractable; hence, we propose a more practical approach to initialize the membrane potential with a \textbf{per-layer constant} that approximates the mean of \( \pi \), as supported by the following lemma:

\textbf{Lemma 1.} Among all constants \( c \in \mathbb{R} \), \( c = E[\pi] \) minimizes expected square difference \(E[(\pi-c)^{2}]\).

The proof is provided in supplementary material (\cref{subsec:proof_of_lemma_1}). This lemma suggests that initializing the membrane potential to the mean of its stationary distribution is optimal for eliminating drift and minimizing variance. However, the next challenge lies in approximating \(E[\pi]\).

One simple solution is leveraging Theorem 1, which states that a membrane potential distribution exponentially converges to \(\pi\) over time. Therefore, we approximate \(\pi\) using the membrane potential at the final timestep (\(U[T]\)), which is the closest to \(\pi\). Based on this, we estimate the per-layer statistics of \(E[\pi]\) using \(E[U[T]]\).

To implement this efficiently, we compute the average membrane potential \(U[T]\) at the final timestep \(T\) for each layer during training. This average is then used to update a running mean \(r\), which is initialized to zero at the beginning of training. Before feeding each batch into the network (during both training and inference), we initialize the membrane potential of every layer to this running mean, i.e., \(U[0] = r\). This technique is called \textbf{MP-Init}, which requires minimal extra memory or computation.

As illustrated in \cref{fig:combined_membrane,fig:combined_activation}, MP-Init effectively resolves both TCS in membrane potential and activation, resulting in consistent distributions over time. Importantly, MP-Init is straightforward to implement and does not require modifications to BN layers or involve complex training pipelines, unlike methods such as TEBN~\cite{tebn:2022} or TAB~\cite{tab:2024}. This simplicity and ease of implementation significantly enhance the usability of SNNs across various applications without altering existing recipes. 

\section{Threshold-robust Surrogate Gradient}
\label{sec:second_method}

\begin{figure}[t!]
    \centering
    \includegraphics[width=0.99\linewidth]{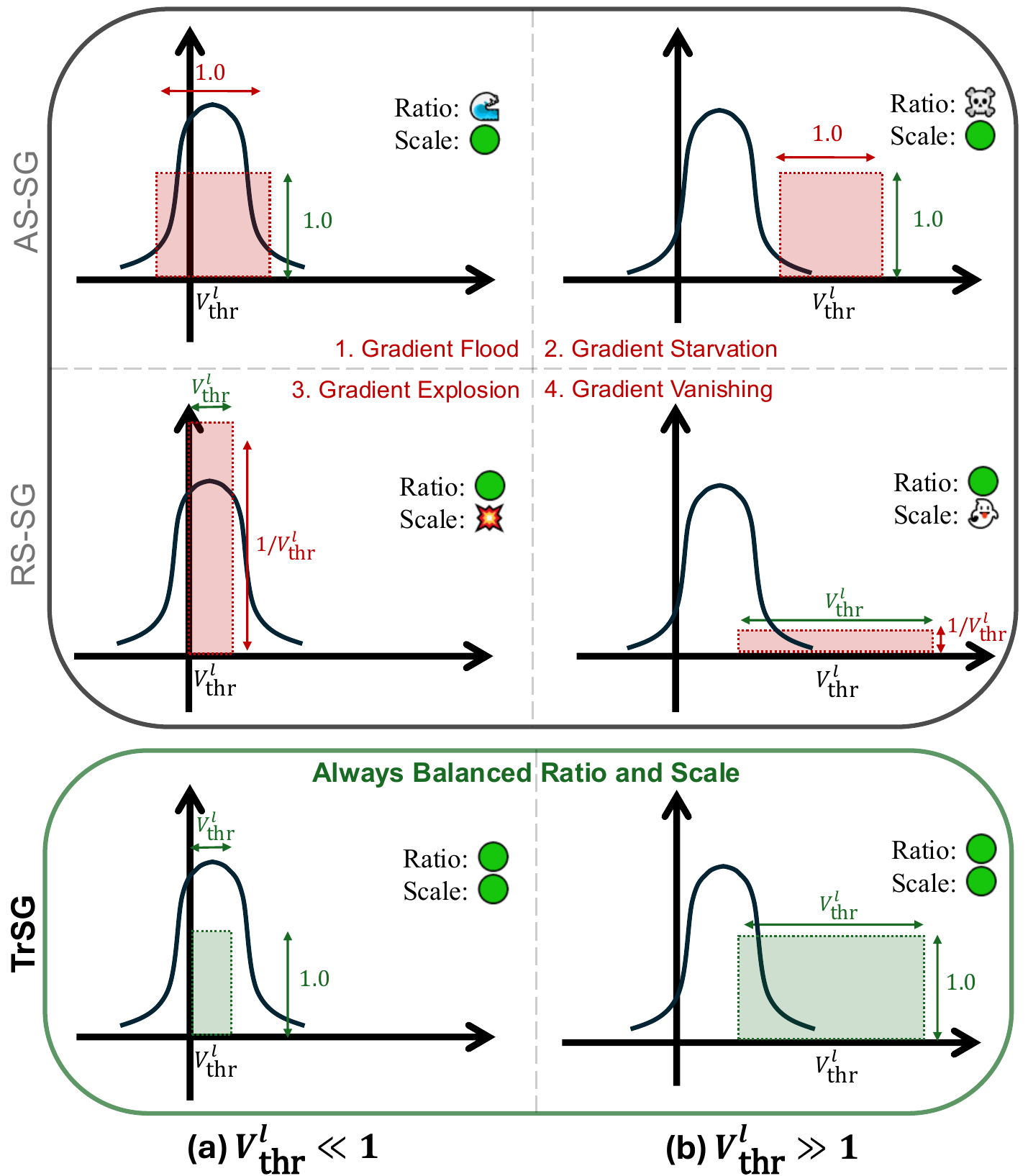}
    \caption{\textbf{AS-SG vs. RS-SG vs. TrSG with a rectangular surrogate gradient ($\gamma=1$).}
    The curve depicts the membrane potential distribution, while the colored boxes represent the surrogate gradient region: the horizontal span indicates the active gradient window, and the vertical extent indicates its magnitude. Panels (a) and (b) compare the behaviors when the threshold is small ($V_{\text{thr}}^{l}\!\ll\!1$) and large ($V_{\text{thr}}^{l}\!\gg\!1$), respectively. \emph{AS-SG} uses a fixed window of width $\gamma$, which leads to \textit{gradient flood} or \textit{starvation}. \emph{RS-SG} normalizes the window by $V_{\text{thr}}^{l}$ but scales the magnitude as $1/V_{\text{thr}}^{l}$, causing \textit{explosion} or \textit{vanishing}. \textbf{TrSG} multiplies the threshold forward during training, canceling the $1/V_{\text{thr}}^{l}$ factor and keeping both window and magnitude balanced across thresholds.}
    \label{fig:second_main}
\end{figure}

Surrogate Gradient (SG)~\cite{wu:2018} enables gradient-based optimization of SNNs by replacing the non-differentiable spiking function with a differentiable surrogate near the threshold. This allows effective gradient descent in SNNs, often improving performance in early timesteps. However, our analysis shows that SG used for internal neuronal parameters~\cite{fang:2021,wang:2022,rathi:2023}, particularly \(V_{\text{thr}}\), makes training highly sensitive to the \emph{scale} of \(V_{\text{thr}}\). While careful tuning can yield strong accuracy, this dependency exposes a fundamental limitation of current SG-based training.

In this section, we identify, \emph{to our knowledge for the first time}, that prior SG designs exhibit strong sensitivity of gradient flow to \(V_{\text{thr}}\). To address this, we introduce \textbf{TrSG (Threshold-robust SG)}, which maintains stable gradients regardless of the threshold value. TrSG is simple to implement, adds negligible overhead, and is compatible with a wide range of surrogate functions.

\subsection{Background: Surrogate Gradient Descent}
\label{sec:background_srgd}

As in Eq.~\ref{eq:lif_discrete_firing}, an LIF neuron emits a spike when the membrane potential crosses the threshold. Let \(H(x)\) denote the Heaviside step. Then the spike can be written as
\[
S^l[t] \;=\; H\!\bigl(x(M^l[t],V^l_{\text{thr}})\bigr),
\]
where the \emph{argument} \(x(\cdot)\) encodes how the membrane potential \(M^l[t]\) and the threshold \(V^l_{\text{thr}}\) are combined (e.g., \(x=M^l[t]-V^l_{\text{thr}}\)). 
Because \(H(\cdot)\) is non-differentiable and \(H'(\cdot)\) is a Dirac delta function, SG training replaces \(H\) with a smooth surrogate \(f\) whose derivative locally mimics \(H'\):
\[
H'(x)\;\approx\; f'(x)\quad\text{near }x=0.
\]
Applying the chain rule to the spike yields the working gradient used in training,
\begin{equation}
\label{eq:def_of_srgd}
\frac{\partial S^l[t]}{\partial M^l[t]}
\;=\;
H'\!\bigl(x\bigr)\,\frac{\partial x}{\partial M^l[t]}
\;\approx\;
f'\!\bigl(x\bigr)\,\frac{\partial x}{\partial M^l[t]},
\end{equation}
so the \emph{choice of \(x(\cdot)\)} is crucial—especially when \(V^l_{\text{thr}}\) is learnable—because it determines both the active region of nonzero gradients and the scaling of their magnitudes.

Upon reviewing the literature, SG-based methods fall into two families according to how they define \(x(M^l[t],V^l_{\text{thr}})\): \emph{Absolute-Scale SG (AS-SG)} and \emph{Relative-Scale SG (RS-SG)}. This distinction becomes significant once the threshold is trained.

\subsection{AS-SG vs. RS-SG}
\label{sec:type1_type2}

To make the scaling effects explicit, we analyze a representative surrogate with a rectangular derivative
\[
f'(x) \;=\; \frac{1}{\gamma}\,\mathds{1}\!\Bigl(|x|<\tfrac{\gamma}{2}\Bigr),
\]
where $\gamma>0$ controls the width of the region where gradients flow. The phenomena we highlight are \emph{not} specific to this choice; any surrogate exhibits the same issues.

For a better illustration, Fig.~\ref{fig:second_main} visualizes a membrane potential distribution under two different threshold scales: small ($V_{\text{thr}}^{l}\!\ll\!1$) and large ($V_{\text{thr}}^{l}\!\gg\!1$).

\paragraph{1.\;AS-SG: Absolute-Scale Surrogate Gradient.}
Many works~\cite{zheng:2021, deng:2022, fang:2021, wang:2022} take
\begin{equation}
\label{eq:abs_diff_x}
x \;=\; M^l[t] - V^l_{\text{thr}},
\end{equation}
which gives
\begin{equation}
\label{eq:abs_diff_grad}
f'\!\bigl(M^l[t]-V^l_{\text{thr}}\bigr)
\;=\;
\frac{1}{\gamma}\,\mathds{1}\!\Bigl(\bigl|M^l[t]-V^l_{\text{thr}}\bigr|<\tfrac{\gamma}{2}\Bigr),
\end{equation}
and thus
\begin{equation}
\label{eq:abs_diff_sgrad}
\frac{\partial S^l[t]}{\partial M^l[t]}
\;\approx\;
f'\!\bigl(M^l[t]-V^l_{\text{thr}}\bigr).
\end{equation}
Here the active window \(\{M^l[t]\in(V^l_{\text{thr}}-\tfrac{\gamma}{2},\,V^l_{\text{thr}}+\tfrac{\gamma}{2})\}\) has a \emph{fixed} width \(\gamma\), and the magnitude is fixed, independent of \(V^l_{\text{thr}}\). When the \(V^l_{\text{thr}}\) is small, the window becomes excessively wide relative to the membrane potential spread, so gradients keep flowing even far from the true firing boundary (\emph{gradient flood}). Conversely, when the \(V^l_{\text{thr}}\) is large, the fixed-width window covers only a tiny fraction of the distribution, and most gradients vanish (\emph{gradient starvation}). 

\paragraph{2.\;RS-SG: Relative-Scale Surrogate Gradient.}
Other works normalize by the threshold; either by \(\gamma=V^l_{\text{thr}}\)~\cite{meng:2023} or by defining
\(
x \;=\; \frac{M^l[t]}{V^l_{\text{thr}}} - 1
\)
as in~\cite{rathi:2023}. In this case,
\begin{equation}
\label{eq:rel_diff_grad}
f'\!\Bigl(\tfrac{M^l[t]}{V^l_{\text{thr}}}-1\Bigr)
\;=\;
\frac{1}{\gamma}\,\mathds{1}\!\Bigl(\Bigl|\tfrac{M^l[t]}{V^l_{\text{thr}}}-1\Bigr|<\tfrac{\gamma}{2}\Bigr),
\end{equation}
so the active window \(\{M^l[t]\in((1-\tfrac{\gamma}{2})V^l_{\text{thr}},\,(1+\tfrac{\gamma}{2})V^l_{\text{thr}})\}\) scales with \(V^l_{\text{thr}}\). 
However, the chain rule introduces a sensitivity of the gradient magnitude to the threshold:
\begin{equation}
\label{eq:rel_diff_sgrad}
\frac{\partial S^l[t]}{\partial M^l[t]}
\;\approx\;
\frac{1}{V^l_{\text{thr}}}\,
f'\!\Bigl(\tfrac{M^l[t]}{V^l_{\text{thr}}}-1\Bigr).
\end{equation}
Thus, small thresholds amplify gradients excessively (\emph{gradient explosion}), whereas large thresholds suppress them severely (\emph{gradient vanishing}). 

\subsection{Threshold-robust Surrogate Gradient}
\label{sec:ta_srgd}

We propose \textbf{TrSG}, a simple modification that preserves the desirable \emph{relative} window while canceling the problematic \(1/V^l_{\text{thr}}\) factor in the gradient. During training we (i) keep the relative-scale argument,
\[
x \;=\; \frac{M^l[t]}{V^l_{\text{thr}}} - 1,
\]
and (ii) multiply the spike by the threshold on the forward path:
\begin{equation}
\label{eq:ta_srgd_core}
O^l[t] \;=\; V^l_{\text{thr}}\cdot S^l[t].
\end{equation}
Then,
\begin{equation}
\label{eq:ta_srgd_chainrule}
\frac{\partial O^l[t]}{\partial M^l[t]}
\;=\;
V^l_{\text{thr}}\;\frac{\partial S^l[t]}{\partial M^l[t]}
\;\approx\;
V^l_{\text{thr}}\cdot
\Bigl(\frac{1}{V^l_{\text{thr}}} f'(x)\Bigr)
\;=\;
f'(x),
\end{equation}
so the gradient magnitude becomes \emph{threshold-invariant} while the active window still adapts with \(V^l_{\text{thr}}\). 

\textbf{Inference remains binary.} The scaling is training-only: at test time we revert to \(S^l[t]\in\{0,1\}\) and absorb \(V^l_{\text{thr}}\) into the next layer via a simple rescaling \(W^{l+1}_{l}\leftarrow V^l_{\text{thr}}\,W^{l+1}_{l}\). 

In summary, \textbf{TrSG} unifies a relative-scale argument with threshold multiplication on the forward path, ensuring stable gradients across thresholds. This design eliminates the flood/starvation issue of AS-SG and the vanish/explode problem of RS-SG, as further demonstrated in \cref{subsec:how_trsg_impacts_training}.

\section{Experiment}

We evaluate \textbf{MP-Init} and \textbf{TrSG} on both static and event-based vision benchmarks, including CIFAR10, CIFAR100~\cite{cifar}, ImageNet~\cite{imagenet}, and DVS-CIFAR10~\cite{dvscifar10}, using diverse backbones such as ResNet-19~\cite{zheng:2021}, a 7-layer CNN~\cite{tebn:2022,tab:2024}, ResNet-34~\cite{resnet}, and SEW-ResNet-34~\cite{fang:2021_2}. Following tdBN~\cite{zheng:2021}, batch normalization is applied across both temporal and batch dimensions. All baselines are trained under their original configurations, and we strictly follow their reported preprocessing pipelines for fairness. Unlike some prior works, we avoid strong augmentations such as AutoAugment~\cite{autoaug} or Cutout~\cite{cutout} to ensure that observed gains come solely from our proposed methods. For every configuration, we run three independent trials and report the results as \(\mathbf{mean}\pm\mathbf{std}\) across runs. Detailed descriptions of architectures, preprocessing, and hyperparameters are provided in the supplementary material (\cref{sec:experimental_detail}).

\begin{table}[t]
\small
\centering
\begin{tabular}{cccc}
\toprule
\textbf{Dataset}     & \textbf{Method}    & \textbf{Timestep} & \textbf{Accuracy (\%)} \\
\toprule
\multirow{6}{*}{CIFAR100}    & tdBN      & \multirow{2}{*}{4} & 75.55 ± 0.06 \\ 
                             & + MP-Init &                    & \textbf{76.09 ± 0.04}    \\
\cline{2-4}
                             & TEBN      & \multirow{2}{*}{4} & 75.96 ± 0.68    \\
                             & + MP-Init &                    & \textbf{76.45 ± 0.24}    \\
\cline{2-4}
                             & TAB       & \multirow{2}{*}{4} & 76.25 ± 0.49    \\
                             & + MP-Init &                    & \textbf{77.24 ± 0.12}    \\ 
\hlineB{1.5}
\multirow{6}{*}{DVS-CIFAR10} & tdBN      & \multirow{2}{*}{10} & 76.60 ± 0.20    \\
                             & + MP-Init &                     & \textbf{77.37 ± 0.38}    \\
\cline{2-4}
                             & TEBN      & \multirow{2}{*}{10} & 75.17 ± 0.50    \\
                             & + MP-Init &                     & \textbf{76.70 ± 0.78}    \\
\cline{2-4}
                             & TAB       & \multirow{2}{*}{4} & 76.43 ± 0.25    \\
                             & + MP-Init &                     & \textbf{76.93 ± 0.38}    \\ 
\bottomrule
\end{tabular}
\caption{Validation of performance improvements from MP-Init using ResNet-19 on CIFAR100 and DVS-CIFAR10 datasets.}
\label{tab:mpinit_acc_impr}
\end{table}

\begin{figure}[t]
  \centering
  \begin{subfigure}{0.32\linewidth}
    \centering
    \includegraphics[width=0.995\linewidth]{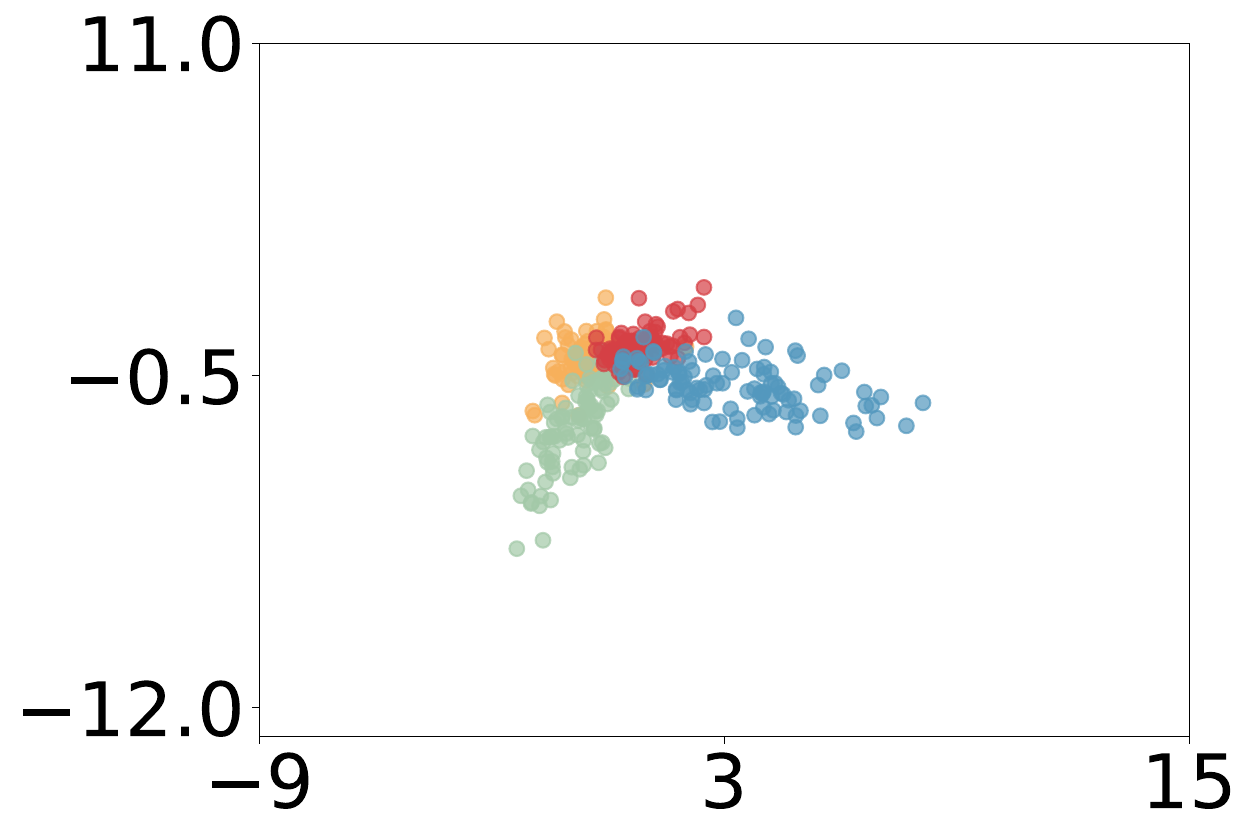}
    \caption{t = 1}
  \end{subfigure}
  \begin{subfigure}{0.32\linewidth}
    \centering
    \includegraphics[width=0.995\linewidth]{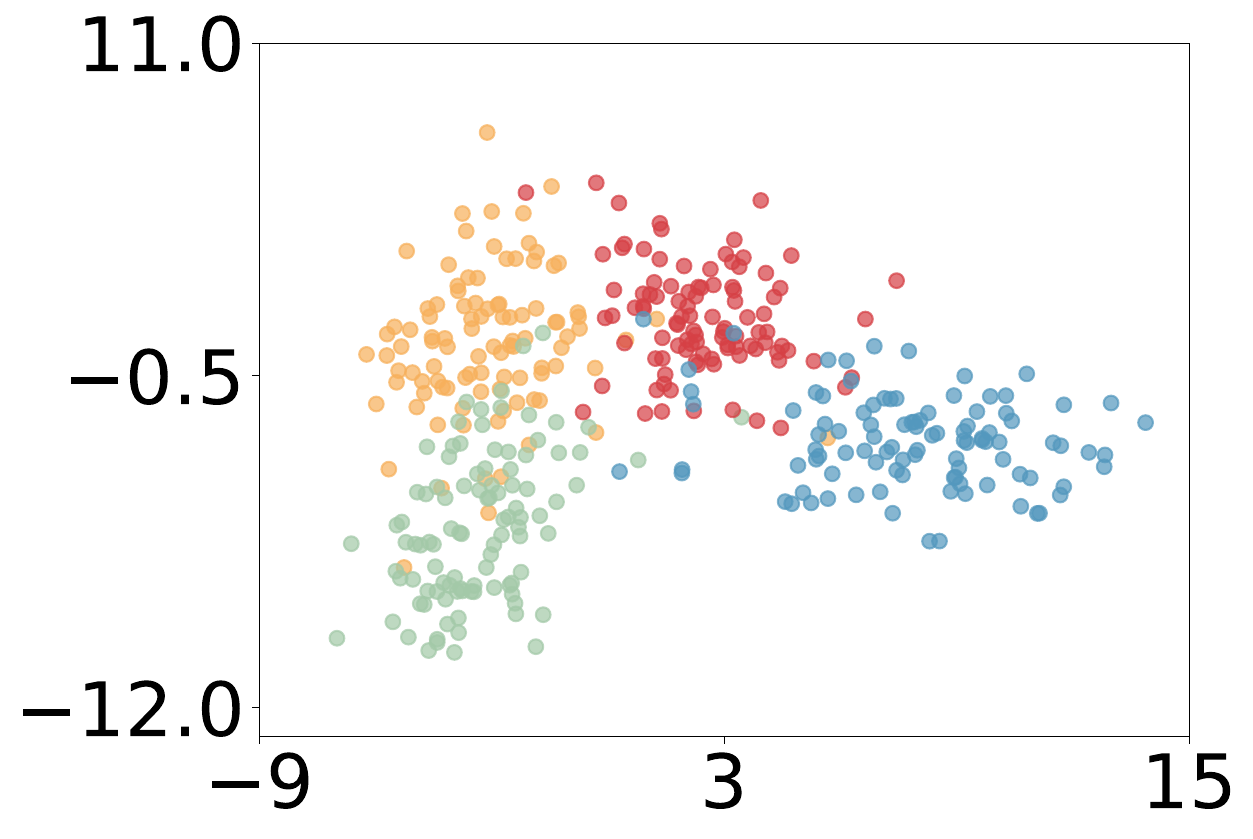}
    \caption{t = 2}
  \end{subfigure}
  \begin{subfigure}{0.32\linewidth}
    \centering
    \includegraphics[width=0.995\linewidth]{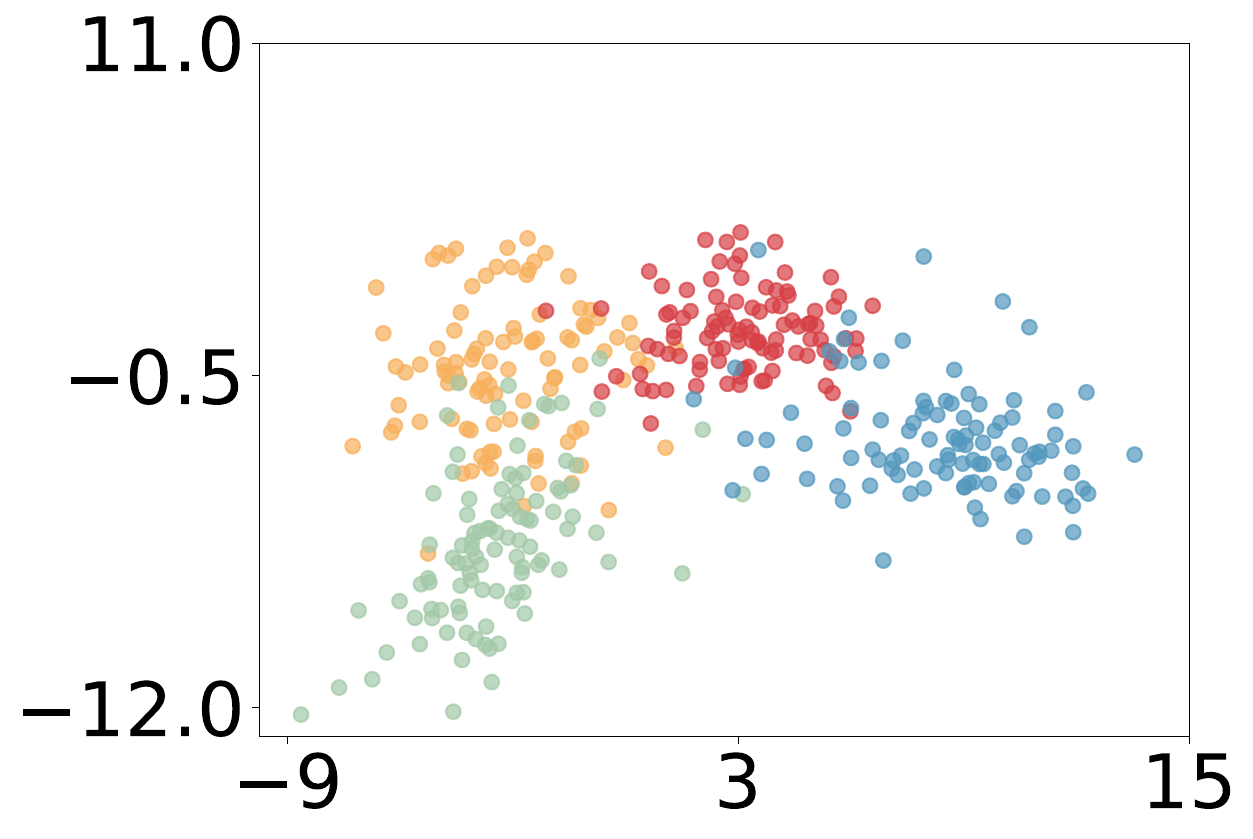}
    \caption{t = 4}
  \end{subfigure}
  \caption{PCA visualization of final logits w/o MP-Init.}
  \label{fig:feature_split_wo_mpinit}
\end{figure}

\begin{figure}[t]
  \centering
  \begin{subfigure}{0.32\linewidth}
    \centering
    \includegraphics[width=0.995\linewidth]{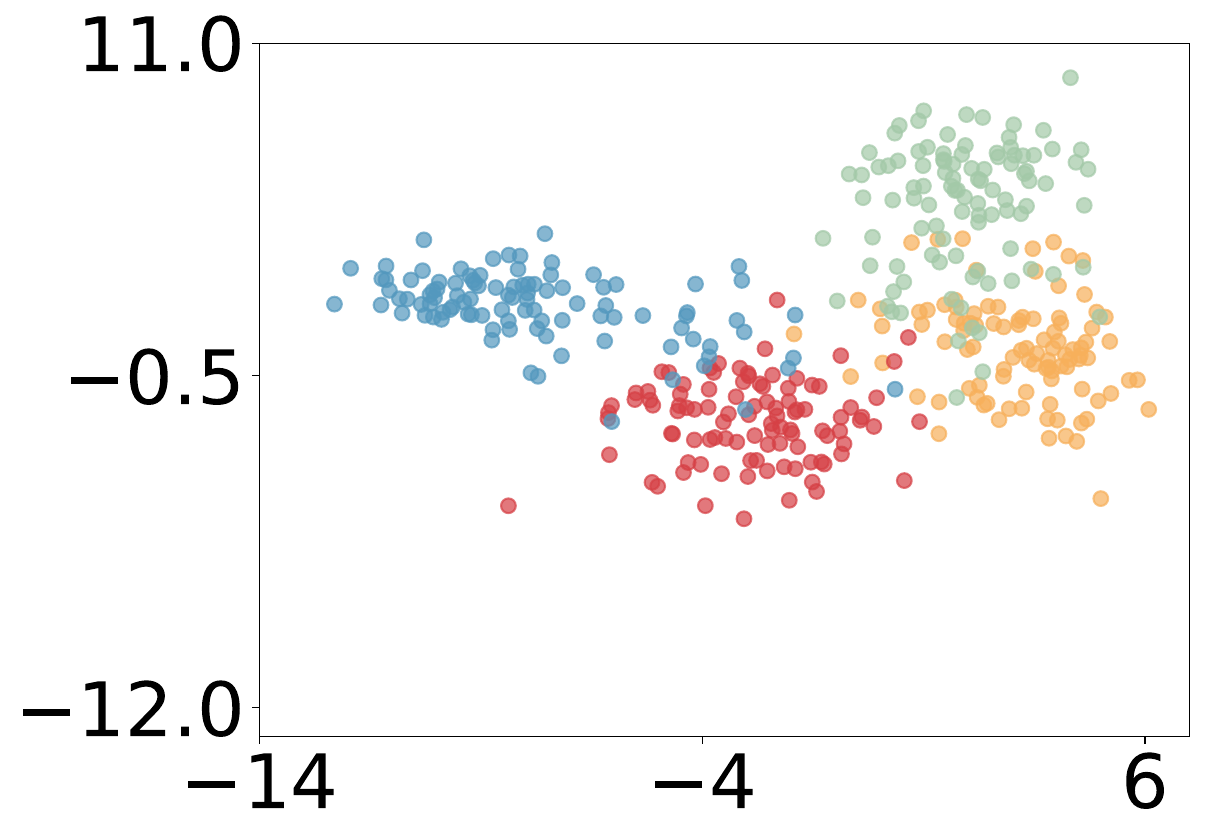}
    \caption{t = 1}
  \end{subfigure}
  \begin{subfigure}{0.32\linewidth}
    \centering
    \includegraphics[width=0.995\linewidth]{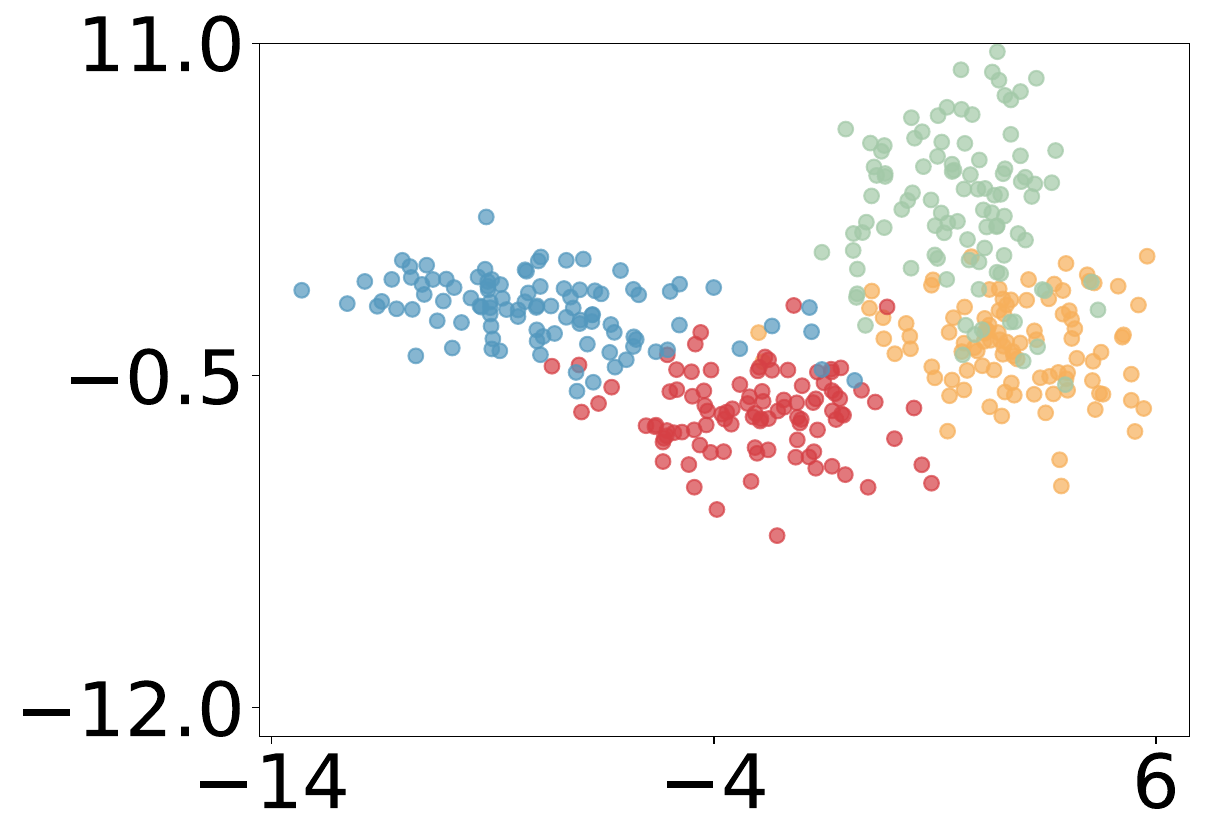}
    \caption{t = 2}
  \end{subfigure}
  \begin{subfigure}{0.32\linewidth}
    \centering
    \includegraphics[width=0.995\linewidth]{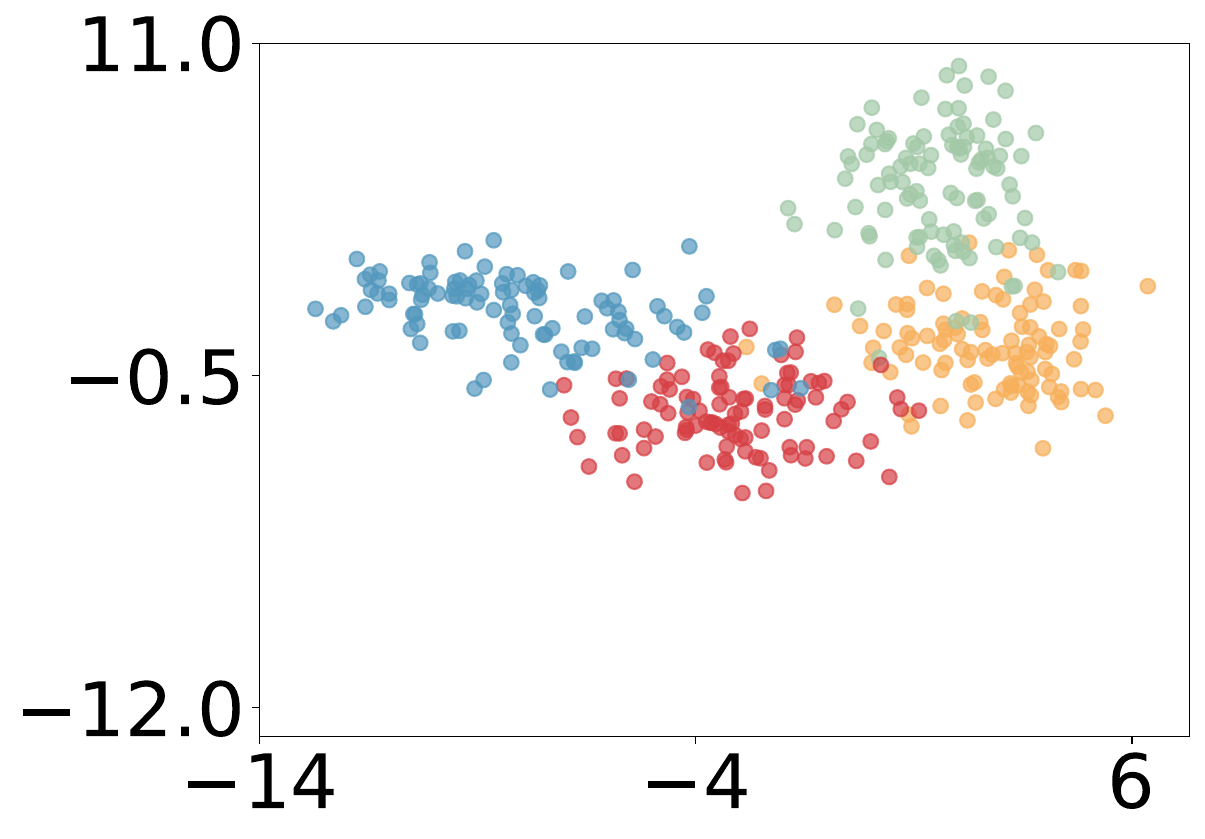}
    \caption{t = 4}
  \end{subfigure}
  \caption{PCA visualization of final logits with MP-Init.}
  \label{fig:feature_split_w_mpinit}
\end{figure}

\subsection{Analysis of MP-Init}
\label{subsec:analysis_of_mpinit}

\paragraph{Effectiveness of MP-Init}

We begin by assessing whether MP-Init effectively enhances the final performance of SNNs. To do this, we train ResNet-19 on CIFAR100 and DVS-CIFAR10 datasets and measure accuracy with and without MP-Init, as shown in Table~\ref{tab:mpinit_acc_impr}. We observe consistent performance improvements with MP-Init, not only in tdBN, which does not account for TCS, but also in TCS-aware methods such as TEBN and TAB. This consistency suggests that previous approaches may only partially address the TCS issue, whereas MP-Init offers a more fundamental solution.

\begin{figure}[t]
  \centering
  \begin{subfigure}{0.495\linewidth}
    \centering
    \includegraphics[width=0.99\linewidth]{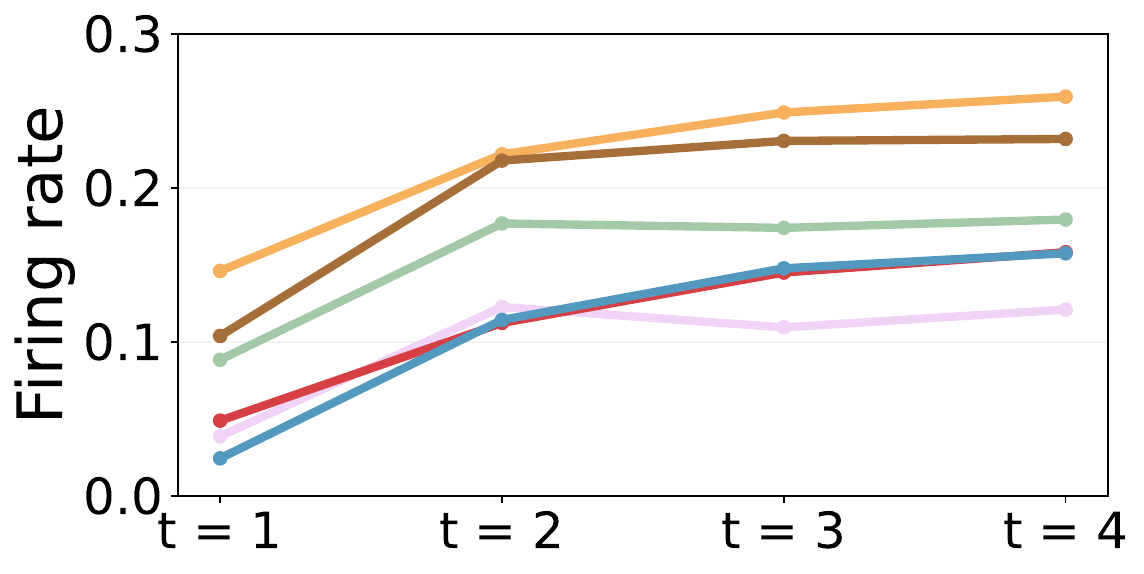}
    \caption{w/o MP-Init}
  \end{subfigure}
  \hfill
  \begin{subfigure}{0.495\linewidth}
    \centering
    \includegraphics[width=0.99\linewidth]{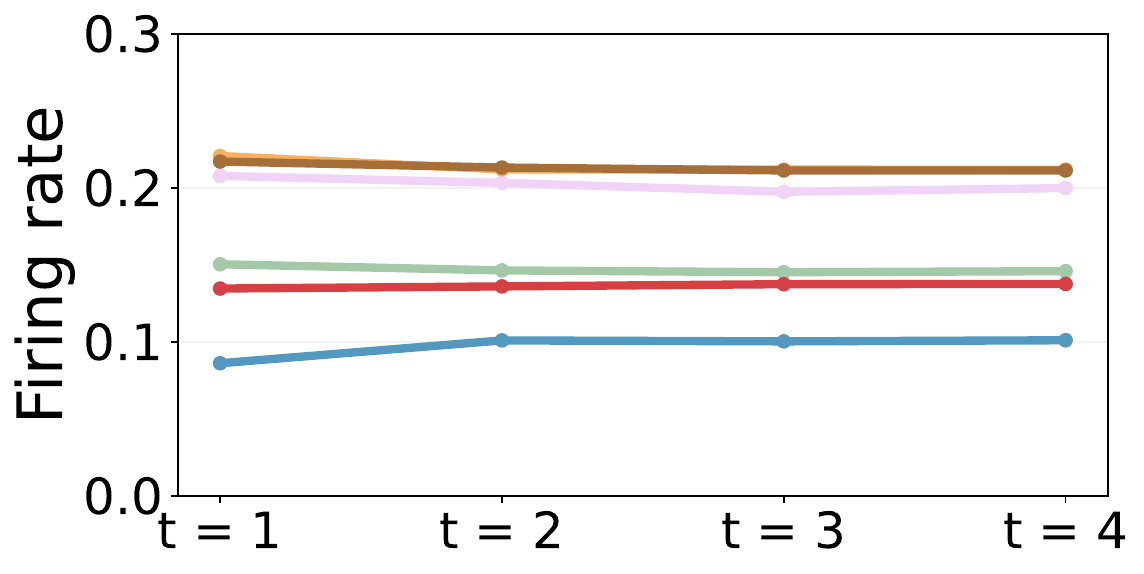}
    \caption{w/ MP-Init}
  \end{subfigure}
  \caption{Firing rate of all neurons at each timestep in arbitrarily selected 6 layers of ResNet-19 on CIFAR10.}
  \label{fig:firing_rate_at_timestep}
\end{figure}

\begin{figure}[t!]
  \centering
  \begin{subfigure}{0.495\linewidth}
    \centering
    \includegraphics[width=0.99\linewidth]{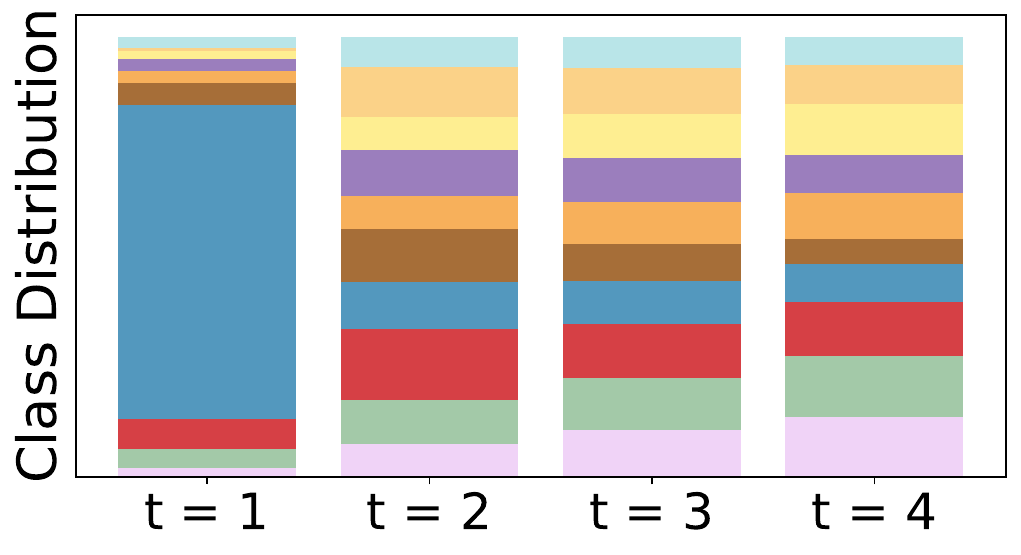}
    \caption{w/o MP-init}
  \end{subfigure}
  \hfill
  \begin{subfigure}{0.495\linewidth}
    \centering
    \includegraphics[width=0.99\linewidth]{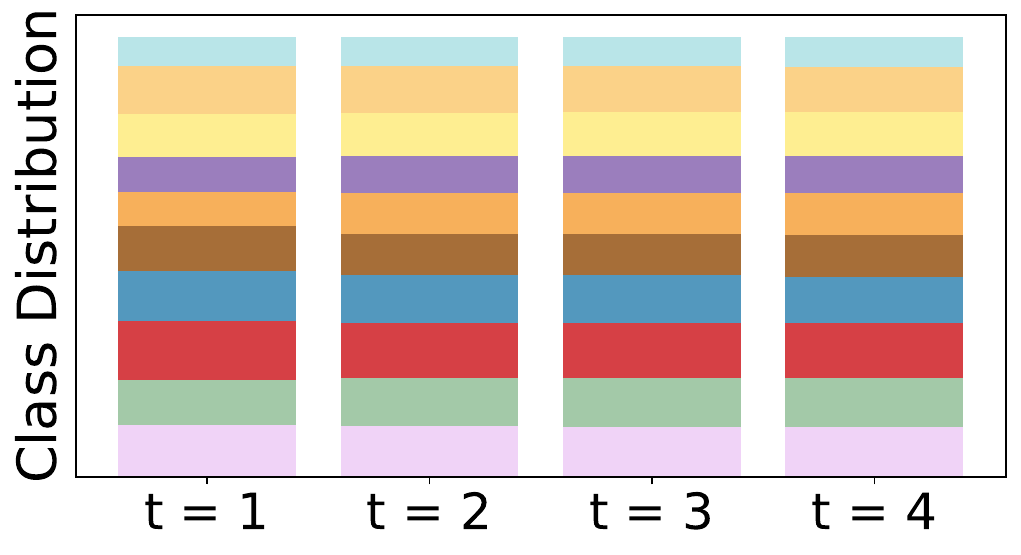}
    \caption{w/ MP-init}
  \end{subfigure}
  \caption{Classification  distributions over timesteps in ResNet-19 on CIFAR10.}
  \label{fig:classification}
\end{figure}

\begin{figure}[t]
  \centering
  \includegraphics[width=0.995\linewidth]{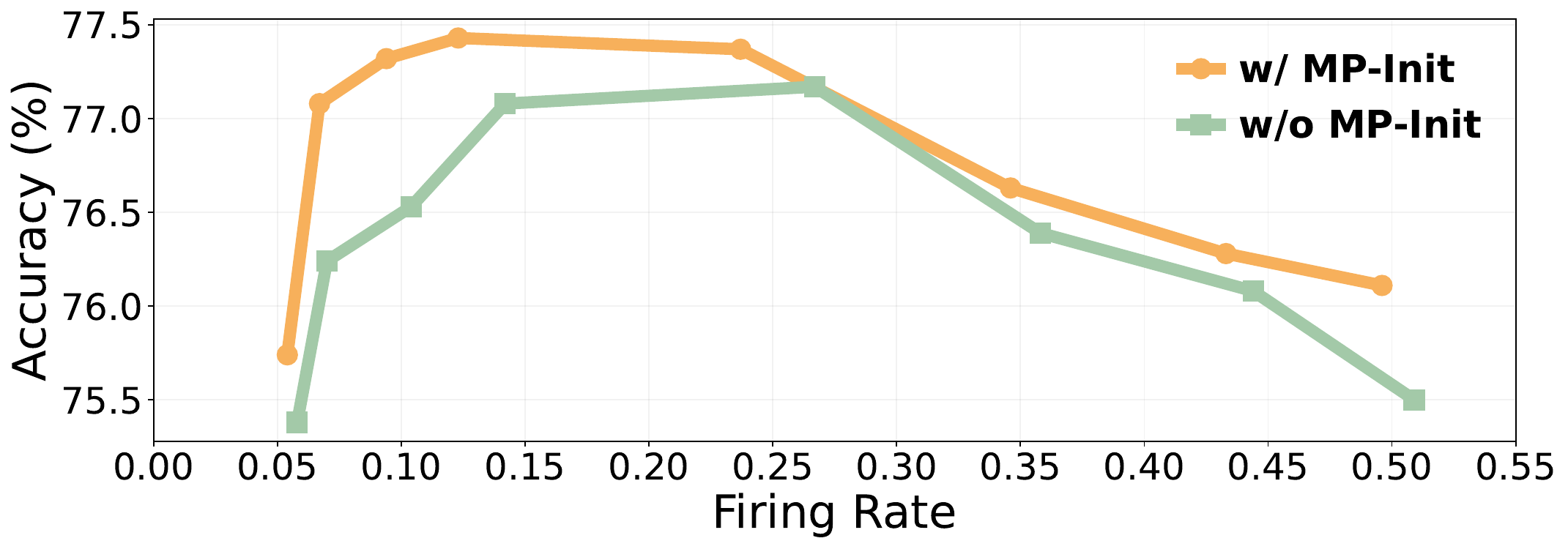}
  \caption{Applying a sparsity loss to control firing rates shows MP-Init forms a Pareto front across all firing rates.}
  \label{fig:firing_rate_vs_accuracy_mpinit}
\end{figure}

\begin{figure*}[t]
    \centering
    
    \begin{subfigure}[b]{0.48\textwidth}
        \centering
        \includegraphics[width=\linewidth]{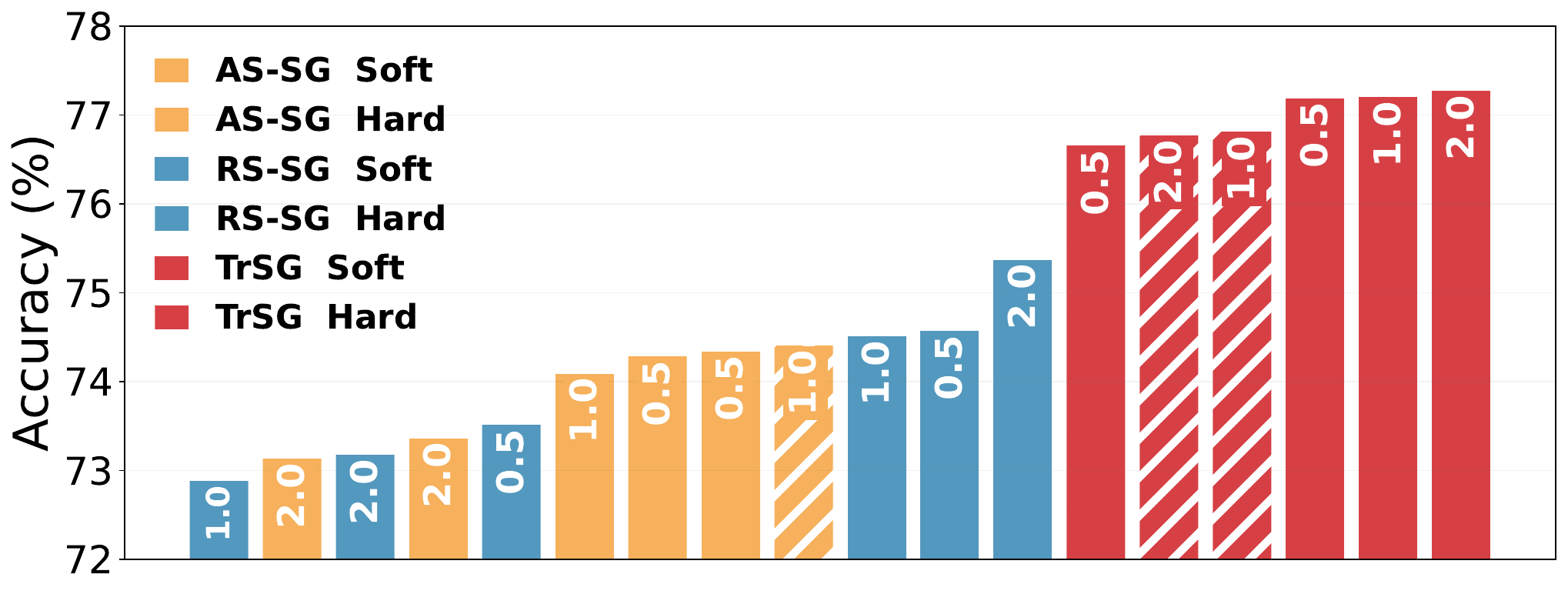}
        \caption{Rectangular-Shaped Surrogate Gradient}
        \label{fig:acc_rect}
    \end{subfigure}
    \hfill
    \begin{subfigure}[b]{0.48\textwidth}
        \centering
        \includegraphics[width=\linewidth]{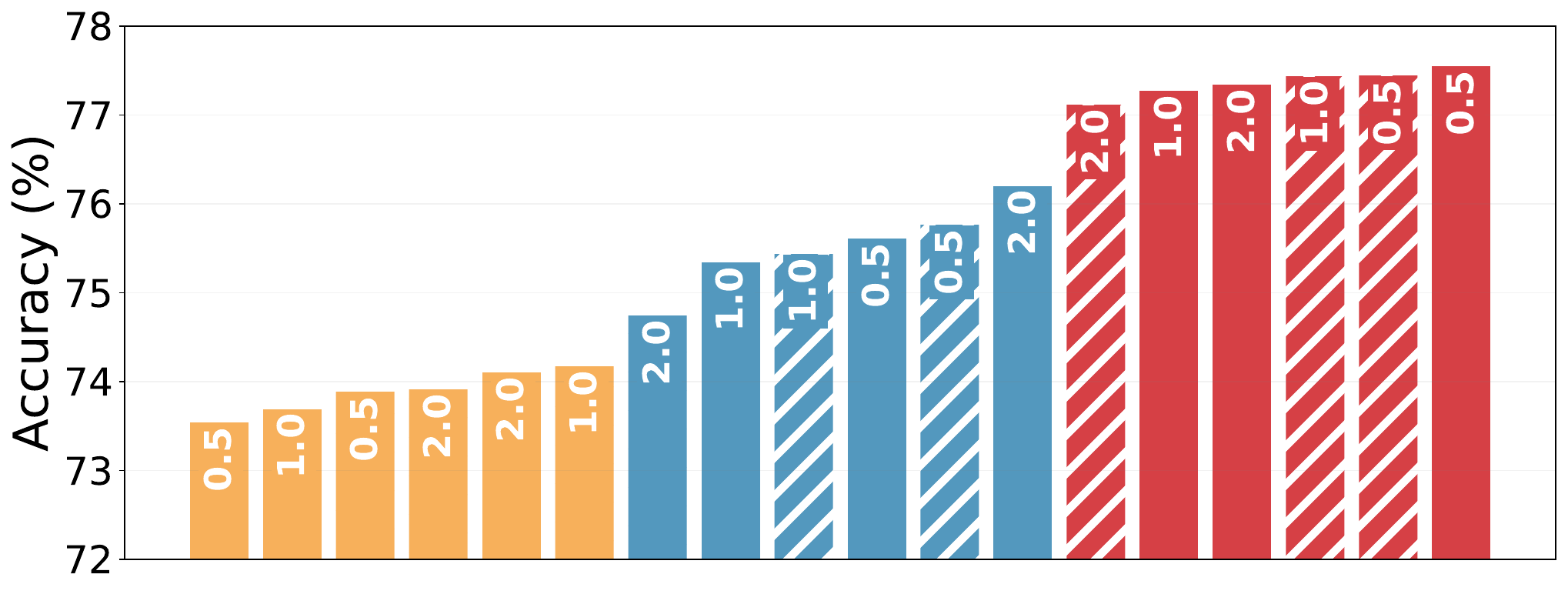}
        \caption{Triangular-Shaped Surrogate Gradient}
        \label{fig:acc_tri}
    \end{subfigure}
    
    \vspace{0.5em}

    \begin{subfigure}[b]{0.24\textwidth}
        \centering
        \includegraphics[width=\linewidth]{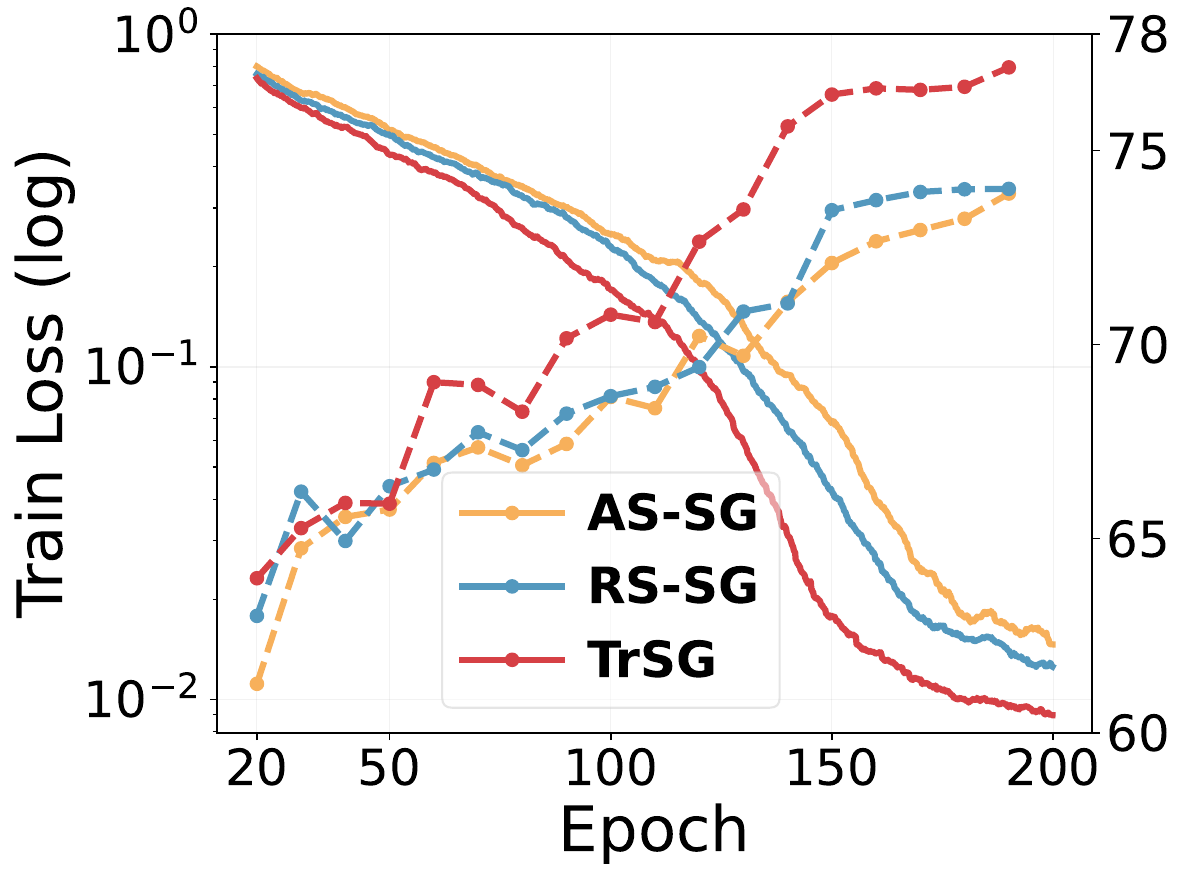}
        \caption{Rectangular (soft)}
        \label{fig:rect_soft}
    \end{subfigure}
    \begin{subfigure}[b]{0.24\textwidth}
        \centering
        \includegraphics[width=\linewidth]{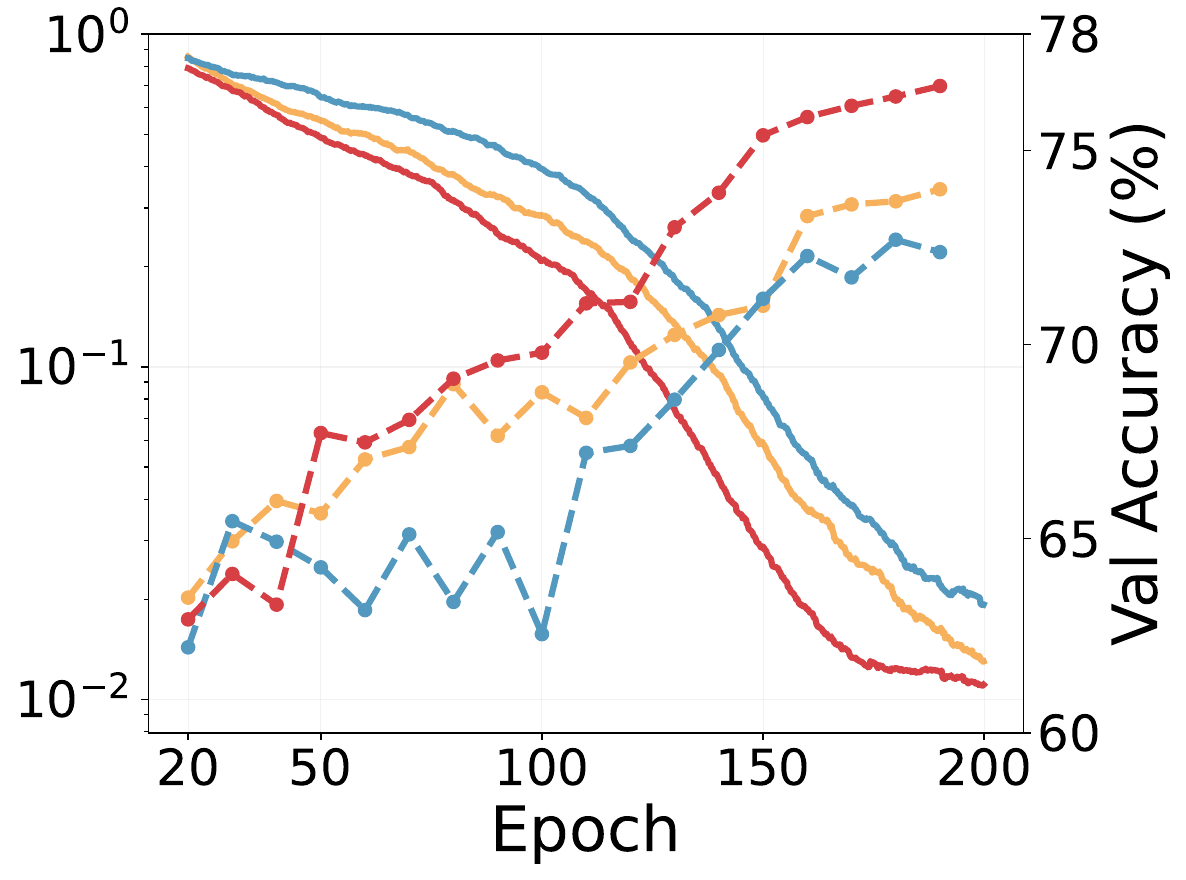}
        \caption{Rectangular (hard)}
        \label{fig:rect_hard}
    \end{subfigure}
    \hfill
    \begin{subfigure}[b]{0.24\textwidth}
        \centering
        \includegraphics[width=\linewidth]{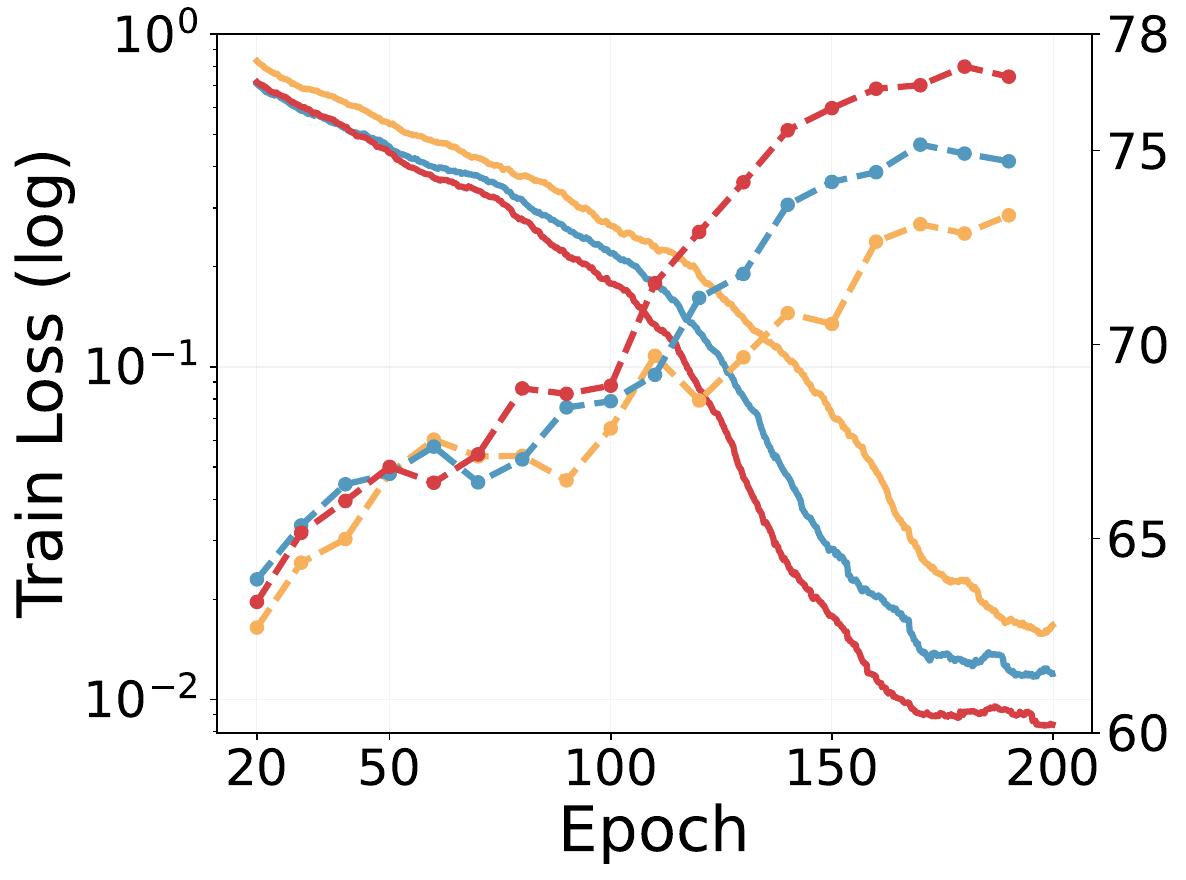}
        \caption{Triangular (soft)}
        \label{fig:tri_soft}
    \end{subfigure}
    \begin{subfigure}[b]{0.24\textwidth}
        \centering
        \includegraphics[width=\linewidth]{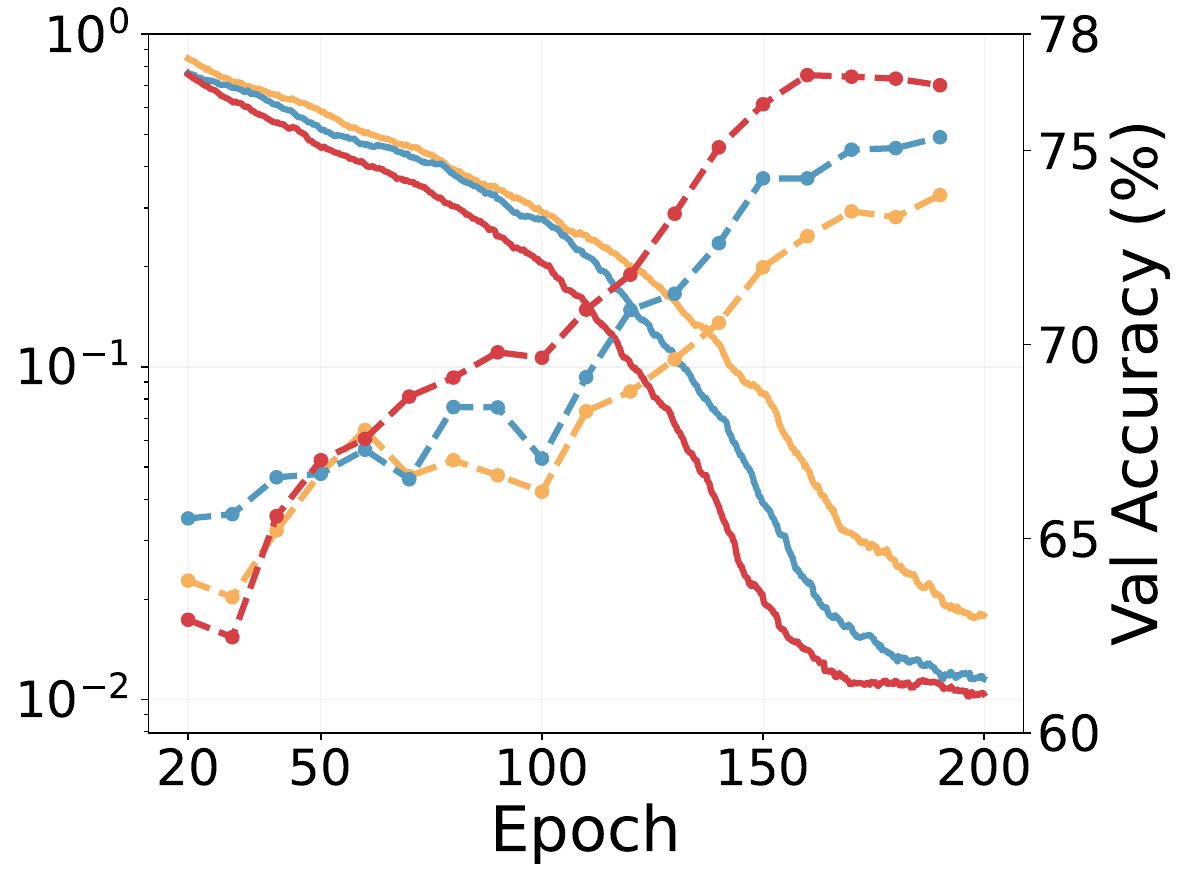}
        \caption{Triangular (hard)}
        \label{fig:tri_hard}
    \end{subfigure}
    \vspace{-5pt}
    \caption{
        \textbf{Comparison of SG approaches under different surrogate function shapes (rectangular and triangular), reset types (soft and hard).}
        \textbf{(Top Row)} shows final accuracy on CIFAR100 (ResNet-19, timestep = 4) 
        for rectangular and triangular surrogate gradient shapes at various initial $V_{\text{thr}}\in\{0.5,1.0,2.0\}$ and an initial $\tau=2.0$. 
        \textbf{(Bottom Row)} plots the training loss and validation accuracy curves with initial $\tau=2.0$ and $V_{\text{thr}}=1.0$. 
        In all experiments, \textbf{TrSG} consistently outperforms the AS-SG and RS-SG methods, demonstrating robustness to different initial values.
    }
    \label{fig:unified_fig}
    \vspace{-8pt}
\end{figure*}

\paragraph{Impact of TCS and Role of MP-Init}


To visualize TCS effects, we train ResNet-19 on CIFAR10 and project the final logits of four randomly selected classes onto two principal components (\cref{fig:feature_split_wo_mpinit,fig:feature_split_w_mpinit}). Without MP-Init, these classes remain entangled in early steps and only gradually separate, reflecting TCS-induced output bias. With MP-Init, class separation emerges from the first timestep. This aligns with the firing-rate dynamics in~\cref{fig:firing_rate_at_timestep}: MP-Init stabilizes neuron activations across timesteps, whereas the baseline accumulates activity over time. Class-distribution analysis (\cref{fig:classification}) further confirms this: early predictions without MP-Init skew heavily toward a certain class, requiring later steps to rebalance. MP-Init addresses this issue by promoting well-distributed class outputs from the start, enabling reliable output even in early timesteps.

\paragraph{Firing Rate vs. Accuracy}
A natural concern after \cref{fig:firing_rate_at_timestep} is whether MP-Init simply increases firing and thereby harms efficiency. To test this, we apply a regularization loss~\cite{yan:2022} for ResNet-19 on CIFAR100 that directly controls spike sparsity and evaluate performance across a wide range of firing rates (\cref{fig:firing_rate_vs_accuracy_mpinit}). The results show that MP-Init does not rely on excessive spiking: it consistently forms a Pareto frontier, achieving higher accuracy than the baseline at the same or even lower firing rates. Remarkably, at a firing rate as low as 0.05, MP-Init still outperforms all baseline settings. This indicates that MP-Init not only enhances accuracy but also enables more energy-efficient SNN operation.

\subsection{Analysis of TrSG}
\label{subsec:analysis_of_ta}

In this section, we compare three distinct ways of defining the SG: \textbf{AS-SG} (Absolute-Scale), \textbf{RS-SG} (Relative-Scale), and our proposed \textbf{TrSG} when training both \(V_{\text{thr}}\) and \(\tau\). To ensure fair evaluation, we use the same training recipes across all methods, as explained in the supplementary material (\cref{subsec:dataset}). To assess initialization sensitivity, we also provide results with different initial values of \(V_{\text{thr}}\) and \(\tau\).

\paragraph{Effectiveness of TrSG.}
We first evaluate three SG approaches using both \text{soft} and \text{hard} reset LIF neurons, along with two widely adopted SG shapes for $f_{sr}'(x)$: \textbf{(i) Rectangular}~\cite{zheng:2021,rathi:2023,wang:2022}: 
    \(
    f'_{sr}(x)=\tfrac{1}{\gamma}\,\mathds{1}\bigl(|x|<\tfrac{\gamma}{2}\bigr)
    \), \textbf{(ii) Triangular}~\cite{meng:2023,deng:2022,tab:2024}: 
    \(
    f'_{sr}(x)=\tfrac{1}{\gamma^2}\,\max\!\bigl(0,\gamma - |x|\bigr).
    \)
We train each configuration to learn both $V_{\text{thr}}$ and $\tau$ under varying initial $V_{\text{thr}}$ values. 
Figure~\ref{fig:unified_fig}\,(a)--(b) (top row) shows the final accuracies on the CIFAR100 dataset using ResNet-19.

Across all conditions, \textbf{TrSG} consistently outperforms both AS-SG and RS-SG, exhibiting robust performance with various initial values of $V_{\text{thr}}$. Overall, TrSG converges to better optima than previously used SG methods, reflecting its stable gradient flow across a wide threshold range.

\paragraph{Training Loss and Validation Accuracy.}
We further examine how the training loss and validation accuracy evolve over epochs. Figure \ref{fig:unified_fig}\,(c)--(f) (bottom row) shows that, for both soft and hard reset neurons, TrSG achieves faster convergence in training loss compared to AS-SG or RS-SG and maintains a higher validation accuracy throughout training. This improvement demonstrates TrSG’s ability to circumvent the gradient issues inherent in AS-SG and RS-SG. Thus, TrSG not only reaches higher final accuracy but also fosters a more stable learning process from the early stages.

A similar trend appears when varying the initial value of $\tau$, and when using different surrogate function shapes (e.g., Sigmoid), which can be found in the supplementary material (\cref{subsec:validation_of_atan,subsec:tau_var_supp}).

\begin{table}[h]
\small
\centering
\begin{tabular}{cccc}
\toprule
\textbf{Dataset} & \textbf{MP-Init} & \textbf{TrSG} & \textbf{Accuracy (\%)} \\
\toprule
\multirowcell{4}[0pt][c]{CIFAR100\\(T=4)} 
 & \ding{55} & \ding{55} & 75.55 ± 0.06 \\
 & \ding{51} & \ding{55} & 76.09 ± 0.04 \\
 & \ding{55} & \ding{51} & 77.15 ± 0.06 \\
 & \ding{51} & \ding{51} & \textbf{77.66 ± 0.15} \\
\midrule
\multirowcell{4}[0pt][c]{DVS-CIFAR10\\(T=10)} 
 & \ding{55} & \ding{55} & 76.60 ± 0.20 \\
 & \ding{51} & \ding{55} & 77.37 ± 0.20 \\
 & \ding{55} & \ding{51} & 80.83 ± 0.64 \\
 & \ding{51} & \ding{51} & \textbf{81.43 ± 0.40} \\
\bottomrule
\end{tabular}
\caption{Ablation study on CIFAR100 and DVS-CIFAR10 using ResNet-19.}
\label{tab:ablation_table}
\vspace{-10pt}
\end{table}

\begin{table*}[t!]
\centering
\small
\begin{tabular}{ccccc}
\toprule
\textbf{Dataset} & \textbf{Method} & \textbf{Network} & \textbf{Timestep} & \textbf{Accuracy (\%)} \\
\toprule
\multirow{5}{*}{CIFAR10} & tdBN~\cite{zheng:2021} & \multirow{5}{*}{ResNet-19} & 2 / 4 / 6 & 92.34 / 92.92 / 93.16 \\
& TET~\cite{deng:2022} & & 2 / 4 / 6 & 94.16 / 94.44 / 94.50 \\
& TEBN~\cite{tebn:2022} & & 2 / 4 / 6 & 94.57 / 94.70 / 94.71 \\
& TAB~\cite{tab:2024} & & 2 / 4 / 6 & 94.73 / 94.76 / 94.81 \\
& \textbf{Ours} & & 2 / 4 / 6 & \textbf{95.05 ± 0.15} / \textbf{95.34 ± 0.06} / \textbf{95.50 ± 0.16} \\
\midrule
\multirow{4}{*}{CIFAR100} & TET~\cite{deng:2022} & \multirow{4}{*}{ResNet-19} & 2 / 4 / 6 & 72.87 / 74.47 / 74.72 \\
& TEBN~\cite{tebn:2022} & & 2 / 4 / 6 & 75.86 / 76.13 / 76.41 \\
& TAB~\cite{tab:2024} & & 2 / 4 / 6 & 76.31 / 76.81 / 76.82 \\
& \textbf{Ours} & & 2 / 4 / 6 & \textbf{76.87 ± 0.16} / \textbf{77.66 ± 0.15} / \textbf{77.91 ± 0.32} \\
\midrule
\multirow{12}{*}{ImageNet} & tdBN~\cite{zheng:2021} & \multirow{4}{*}{ResNet-34} & 6 & 63.72 \\
& Dspike~\cite{li:2021} & & 6 & 68.19 \\
& TEBN~\cite{tebn:2022} & & 4 & 64.29 \\
& TAB~\cite{tab:2024} & & 4 & 67.78 \\
\addlinespace[1pt]
\cline{2-5}
\addlinespace[1pt]
& SEW-ResNet~\cite{fang:2021_2} & \multirow{6}{*}{SEW-ResNet-34} & 4 & 67.04 \\
& TET~\cite{deng:2022} & & 4 & 68.00 \\
& TEBN~\cite{tebn:2022} & & 4 & 68.28 \\
& IMP+LTS~\cite{shen:2024} & & 4 & 68.90 \\
& MPS~\cite{ding:2025} & & 4 & 69.03 \\
& EAGD~\cite{yang:2025} & & 4 & 68.12 \\

\addlinespace[1pt]
\cline{2-5}
\addlinespace[1pt]
& \multirow{2}{*}{\textbf{Ours}} & ResNet-34 & 4 / 6 & 67.15 ± 0.03 / \textbf{68.73 ± 0.09} \\
& & SEW-ResNet-34 & 4 & \textbf{69.67 ± 0.08} \\
\midrule
\multirow{8}{*}{DVS-CIFAR10} & tdBN~\cite{zheng:2021} & \multirow{3}{*}{ResNet-19} & 10 & 67.80 \\
& MPBN~\cite{guo:2023_2} & & 10 & 74.40 $\pm$ 0.20 \\
& RMP-Loss~\cite{guo:2023} & & 10 & 76.20 $\pm$ 0.20 \\
\addlinespace[1pt]
\cline{2-5}
\addlinespace[1pt]
& BNTT~\cite{bntt:2020} & \multirow{3}{*}{7-layer CNN} & 20 & 63.20 \\
& TEBN~\cite{tebn:2022} & & 10 & 75.10 \\
& TAB~\cite{tab:2024} & & 4 & 76.70 \\
\addlinespace[1pt]
\cline{2-5}
\addlinespace[1pt]
& \multirow{2}{*}{\textbf{Ours}} & ResNet-19 & 10 & \textbf{81.43 ± 0.40} \\
&  & 7-layer CNN & 4 & \textbf{79.27 ± 0.75} \\
\bottomrule
\end{tabular}
\caption{Comprehensive comparison of our methods on static and dynamic datasets.}
\label{tab:comprehensive_results}
\vspace{-5pt}
\end{table*}

\subsection{Ablation Study and Final Configuration}
\label{subsec:ablation_and_config}

Building on earlier findings, given their superior quality and stability, we select \textbf{rectangular-shaped} surrogate functions and \textbf{soft reset} neurons for subsequent experiments. 

Next, we validate the individual and combined effects of MP-Init and TrSG through an ablation study on CIFAR100 and DVS-CIFAR10. From Table~\ref{tab:ablation_table}, both MP-Init and TrSG clearly improve performance over the baseline. These results confirm their complementary contributions and justify adopting both MP-Init and TrSG as our default setup for subsequent experiments.

\subsection{Comparison with State-of-the-arts}

We compare our methods, \textbf{TrSG} and \textbf{MP-Init}, against state-of-the-art methods on both static datasets (CIFAR10/100, ImageNet) and dynamic datasets (DVS-CIFAR10), using standard LIF neurons and various backbones. Comprehensive results are summarized in Table~\ref{tab:comprehensive_results}.

On \textbf{CIFAR10 and CIFAR100}, our method achieves the best accuracy across all evaluated timesteps. On CIFAR10, we reach 95.50\% at 6 timesteps, improving over the previous best (94.81\% with TAB) by 0.69\%pt. On CIFAR100, our method obtains 77.91\% at 6 timesteps, surpassing the strongest baseline (76.82\% with TAB) by 1.09\%pt. Moreover, our results at fewer timesteps outperform baselines at larger timesteps, highlighting improved efficiency.

On \textbf{ImageNet}, our approach achieves 68.73\% top-1 accuracy with ResNet-34 at 6 timesteps. With SEW-ResNet-34, we further push performance to 69.67\% at only 4 timesteps, outperforming prior strong methods such as IMP+LTS~\cite{shen:2024} (68.90\%) and MPS~\cite{ding:2025} (69.03\%). Unlike these approaches, which require training additional neuron-specific parameters or complex optimization, our method introduces negligible overhead.

On the dynamic dataset \textbf{DVS-CIFAR10}, our method delivers 81.43\% accuracy with ResNet-19 at 10 timesteps, exceeding the previous best (76.20\% with RMP-Loss). With a lightweight 7-layer CNN, we also achieve 79.27\% at only 4 timesteps, demonstrating that the benefits of TrSG and MP-Init generalize broadly to both static and event-based data.

\section{Conclusion}
\label{sec:conclusion}

In this paper, we introduced two key methods, MP-Init and TrSG, to tackle core challenges in training SNNs. MP-Init reduces TCS by aligning membrane potentials with their stationary distribution. At the same time, TrSG stabilizes the training process by stabilizing gradient flow with respect to the threshold voltage, allowing us to achieve state-of-the-art accuracy on static and dynamic datasets. Together, they contribute to improving the stability, efficiency, and accessibility of SNN training with minimal overhead.
{
    \small
    \bibliographystyle{ieeenat_fullname}
    \bibliography{main}
}



\clearpage
\maketitlesupplementary

\section{General Details}

\subsection{Biological Plausibility of MP-Init}
A neuroscience paper~\cite{yoo:2023} shows that resting membrane potentials (RMPs) differ across distinct cortical regions. Specifically, neurons in the prefrontal cortex (PFC) show approximately 10 mV higher resting membrane potentials compared to neurons in the posterior parietal cortex (PPC), primarily driven by differences in recurrent synaptic strengths mediated by NMDA receptors. These regional differences imply that the stationary membrane potential distribution may inherently vary across neural populations and brain areas. Therefore, the layer-wise initialization of MP-Init is similar to the brain's intrinsic regional variability in membrane potentials, thus providing a biological plausibility and justification for our proposed MP-Init method.

\subsection{Limitations}
Despite the advancements presented in the paper, several limitations remain. MP-Init efficiently mitigates TCS with minimal overhead, but exploring alternative initialization strategies could provide further insights. While training internal parameters of LIF neurons may introduce challenges in deploying SNNs on neuromorphic hardware optimized for simpler neuron dynamics. Future work should focus on adapting these methods for broader architectures and exploring more hardware-compatible designs for real-world applications.

\begin{table}[h]
\centering
\small
\begin{tabular}{llc}
\toprule
\textbf{Dataset} & \textbf{Architecture} & \textbf{Accuracy (\%)} \\
\midrule
\multirow{7}{*}{DVS-CIFAR10} 
& Spikformer~\cite{zhou2023spikformer} & 80.90 \\
& Spikingformer~\cite{zhou2023spikingformer} & 81.30 \\
& S-Transformer~\cite{yao2023stransformer} & 80.00 \\
& SpikeFormer2~\cite{zhou:2024} & 78.90 \\
& \quad + \textbf{Ours} & \textbf{80.10} \\
& QKFormer~\cite{zhou:2024_2} & 83.80 \\
& \quad + \textbf{Ours} & \textbf{84.80} \\
\midrule
\multirow{7}{*}{ImageNet}
& Spikformer~\cite{zhou2023spikformer} & 70.24 \\
& Spikingformer~\cite{zhou2023spikingformer} & 72.45 \\
& S-Transformer~\cite{yao2023stransformer} & 72.28 \\
& SpikingResformer~\cite{shi:2024} & 74.34 \\
& \quad + \textbf{Ours} & \textbf{75.62} \\
& QKFormer~\cite{zhou:2024_2} & 78.80 \\
& \quad + \textbf{Ours} & \textbf{79.90} \\
\bottomrule
\end{tabular}
\caption{Transformer-based SNNs with and without our method. 
DVS-CIFAR10 is evaluated at 10 timesteps; ImageNet at 4 timesteps. 
Baselines are selected with comparable parameter budgets; our variant applies MP-Init + TrSG with all other settings held fixed.}
\label{tab:advanced_snn}
\end{table}

\begin{table}[h]
\centering
\small
\begin{tabular}{lc}
\toprule
\textbf{Architecture} & \textbf{COCO2017 (mAP)} \\
\midrule
EMS-ResNet10~\cite{su:2023} & 0.291 / 0.138 \\
\quad + \textbf{Ours} & \textbf{0.305} / \textbf{0.147} \\
\bottomrule
\end{tabular}
\caption{Object detection on COCO2017 with EMS-ResNet10 (timesteps=3). 
Baseline and ours are trained under the \emph{same} configuration for 100 epochs; values are mAP@0.5 / mAP@0.5:0.95.}
\label{tab:object_detection}
\end{table}

\subsection{Extension to Advanced Architectures and Tasks}

We further evaluate MP-Init and TrSG on \textbf{Transformer-based SNNs}. 
On \textbf{DVS-CIFAR10} (10 timesteps), our method improves SpikeFormer2 from 78.90\% to \textbf{80.10\%} and QKFormer from 83.80\% to \textbf{84.80\%}. 
On \textbf{ImageNet} (4 timesteps), SpikingResformer rises from 74.34\% to \textbf{75.62\%}, and QKFormer from 78.80\% to \textbf{79.90\%} (\cref{tab:advanced_snn}). 
These \textbf{+1.0–1.3 \%pt} gains indicate that MP-Init and TrSG transfer cleanly to advanced Transformer backbones without architectural changes.

Beyond classification, we validate task generality on \textbf{COCO2017} detection with EMS-ResNet10~\cite{su:2023}. 
Using the same training configuration (100 epochs) for both baseline and ours, MP-Init + TrSG improves mAP from 0.291 to \textbf{0.305} (mAP@0.5) and from 0.138 to \textbf{0.147} (mAP@0.5:0.95), as summarized in \cref{tab:object_detection}. 
Together, these results show that our methods are broadly effective across architectures and tasks while preserving the efficient inference pipeline of standard LIF SNNs.

\section{Implementation Details}
\label{sec:experimental_detail}
\subsection{Architecture}
We employ various network architectures to validate the effectiveness of MP-Init and TrSG. For the CIFAR-10/100~\cite{cifar} and ImageNet~\cite{imagenet} datasets, we utilize ResNet-19~\cite{zheng:2021} and ResNet-34~\cite{resnet}. On the DVS-CIFAR10 dataset~\cite{dvscifar10}, we apply a 7-layer CNN (64C3-AP2-128C3-128C3-AP2-256C3-256C3-AP2-1024FC-10FC) as outlined in TEBN~\cite{tebn:2022} and TAB~\cite{tab:2024}, in addition to ResNet-19. We integrate the temporal dimension into the batch dimension to ensure proper normalization, following the tdBN method~\cite{zheng:2021}.

In MP-Init and TrSG, the running mean of membrane potential \(E[U[T]]\), the threshold potential (\(V_{thr}\)), and the time decay factor (\(\tau\)) are layer-wise parameters. We also tried channel-wise versions of them, but found only a small difference in performance. We set the initial \(E[U[T]]\) to 0.0, \(V_{thr}\) to 1.0, and \(\tau\) to 2.0, which are commonly used in SNNs~\cite{deng:2022,tebn:2022,fang:2021_2,fang:2021,meng:2023,shen:2024,zhou:2024_2,shi:2024} without further tuning on networks/datasets for a fair comparison. The network directly processes the input through the first layer, which acts as the stem layer to generate spikes, while the final layer does not generate spikes~\cite{deng:2022}. To enhance generalization on the CIFAR-10/100 and DVS-CIFAR10 datasets, we incorporate dropout layers before each fully connected layer, consistent with the approach used in TEBN and TAB~\cite{tebn:2022,tab:2024}.

\subsection{Algorithm}
\label{sec:algorithm}

\paragraph{MP-Init: Running-mean initialization.}
To mitigate TCS, we initialize each layer’s membrane potential at the beginning of a simulation window using a running mean of the final potential from the previous window. The statistic is computed only from \emph{active neurons} (those with $S^{l}[t]>0$ at least once), so silent neurons do not bias the estimate. The running mean is updated \emph{during training only}; at inference it is fixed and reused. Unless otherwise stated, we set the EMA coefficient to $\beta=0.9$.

\begin{algorithm}[h]
\caption{MP-Init (per-layer initialization)}
\label{alg:mpinit}
\begin{algorithmic}[1]
\STATE Initialize per-layer running mean $\mu^{l}\!\leftarrow\!0$ for each layer $l$
\STATE \textbf{Training phase:}
\FOR{each batch with simulation window of length $T$}
  \STATE Set initial membrane $U^{l}[0] \leftarrow \mu^{l}$
  \STATE Initialize activity mask $\texttt{mask}^{l}\leftarrow 0$
  \FOR{$t=1$ to $T$}
    \STATE Update $M^{l}[t]$ by \cref{eq:lif_discrete_potential}, compute spikes $S^{l}[t]$ by \cref{eq:lif_discrete_firing}, and reset $U^{l}[t]$ by \cref{eq:lif_discrete_reset}
    \STATE $\texttt{mask}^{l} \leftarrow \texttt{mask}^{l} \lor (S^{l}[t] > 0)$
  \ENDFOR
  \STATE $\mu_{\text{batch}}^{l} \leftarrow \mathrm{mean}\!\big(U^{l}[T]\ \big|\ \texttt{mask}^{l}{=}1\big)$
  \STATE $\mu^{l} \leftarrow (1-\beta)\mu^{l} + \beta\,\mu_{\text{batch}}^{l}$
\ENDFOR
\STATE
\STATE \textbf{Inference phase:}
\STATE Set initial membrane $U^{l}[0] \leftarrow \mu^{l}$
\STATE Run forward dynamics with \cref{eq:lif_discrete_potential,eq:lif_discrete_firing,eq:lif_discrete_reset}
\end{algorithmic}
\end{algorithm}

\paragraph{Stable parameterization of \(\tau\) and \(V_{\text{thr}}\).}
To keep LIF parameters within valid ranges, we reparameterize
\[
\tau = \tfrac{1}{\sigma(w)} \in (1,\infty), \quad 
V_{\text{thr}} = \mathrm{Softplus}(k) \in (0,\infty),
\]
where \(w,k\) are unconstrained learnable variables and \(\sigma(\cdot)\) is the logistic sigmoid~\cite{fang:2021}. This yields stable gradients w.r.t.\ \((w,k)\) while ensuring feasible \((\tau,V_{\text{thr}})\).

\paragraph{Training vs.\ inference.}

\emph{MP-Init:} During training, we track spikes $S^{l}[t]$ to update $\mu^{l}$ as in \cref{alg:mpinit}. At inference, $\mu^{l}$ is no longer updated; it simply initializes $U^{l}[0]$, leaving the LIF forward rule unchanged.  

\emph{TrSG:} As detailed in \cref{sec:ta_srgd}, threshold multiplication is a \emph{training-only} trick; at test time inference remains binary ($S^{l}[t]\!\in\!\{0,1\}$), and $V_{\text{thr}}^{l}$ is absorbed into the next layer by rescaling $W^{l+1}_{l}\!\leftarrow\! V_{\text{thr}}^{l} W^{l+1}_{l}$. Thus, both MP-Init and TrSG keep the inference pipeline identical to standard LIF.

\paragraph{Practical note on efficiency.}
SpikingJelly~\cite{spikingjelly} currently lacks CUDA‑optimized kernels for jointly training \(V_{\text{thr}}\) and \(\tau\). We therefore implemented CUDA kernels supporting AS‑SG, RS‑SG, and TrSG, which accelerates end‑to‑end training substantially. For ResNet‑34 on ImageNet with 8$\times$A100‑80GB, the total wall‑clock for 120 epochs drops from \(\sim 32.5\) hours to \(\sim 16.7\) hours (about \(1.94\times\) speed‑up).

\subsection{Dataset}
\label{subsec:dataset}

\paragraph{CIFAR-10/100:} We train ResNet-19 with timesteps of 2, 4, and 6, using one RTX3090 GPU. The optimization is performed using the SGD optimizer with a weight decay of 5e-4 and a CosineAnnealing scheduler, starting with an initial learning rate of 0.02 and a batch size of 64. We apply random cropping to 32x32 pixels, random horizontal flips, and data normalization.

\paragraph{DVS-CIFAR10:} We train the 7-layer CNN and ResNet-19 with timesteps of 4 and 10, respectively, using one RTX3090 GPU. The training setup includes an SGD optimizer with a weight decay of 5e-4 and a CosineAnnealing scheduler, starting with an initial learning rate of 0.02 and a batch size of 64. The images are randomly cropped to 48x48 pixels and horizontally flipped, following the method described in~\cite{tebn:2022, tab:2024}.

\paragraph{ImageNet:} We train ResNet-34 with 6 timesteps and report results at multiple inference timesteps without additional fine-tuning. The optimizer is SGD with a weight decay of \(4\times10^{-5}\), paired with a CosineAnnealing scheduler that starts at 0.10. The batch size is 512, distributed across 8 NVIDIA A100-80GB GPUs with synchronized batch statistics. SEW-ResNet-34 follows the same procedure and hyperparameters.  Data augmentation includes random cropping and resizing to 224x224 pixels, random horizontal flips, and data normalization.

\subsection{Parameter Overhead}

If timestep-dependent parameters are introduced for each neuron, the parameter overhead would scale as \(O(L T N)\) at maximum, where \(L\), \(T\), and \(N\) represent the number of layers, timesteps, and neurons per layer, respectively. This represents the worst-case scenario, assuming every neuron at every timestep requires independent parameters. By leveraging layerwise statistics, we significantly reduce this complexity. The total parameter overhead is reduced to \(O(L)\), as \(V_{\text{thr}}\), \(\tau\), and the running mean of the membrane potential are single parameters at each layer.

In comparison, prior works such as TEBN~\cite{tebn:2022} and TAB~\cite{tab:2024} employ timestep-dependent parameters, resulting in an overhead of \(O(LT)\). IMP+LTS~\cite{shen:2024}, which trains the initial membrane potential per neuron, incurs an overhead of $O(N)$, where ($N>>L,T$). In comparison, our method provides superior efficiency and achieves higher accuracy simultaneously.

\section{Additional Details: MP-Init}
\label{sec:additional_detail_mp_init}

\begin{figure}[h]
    \centering
    \includegraphics[width=0.99\linewidth]{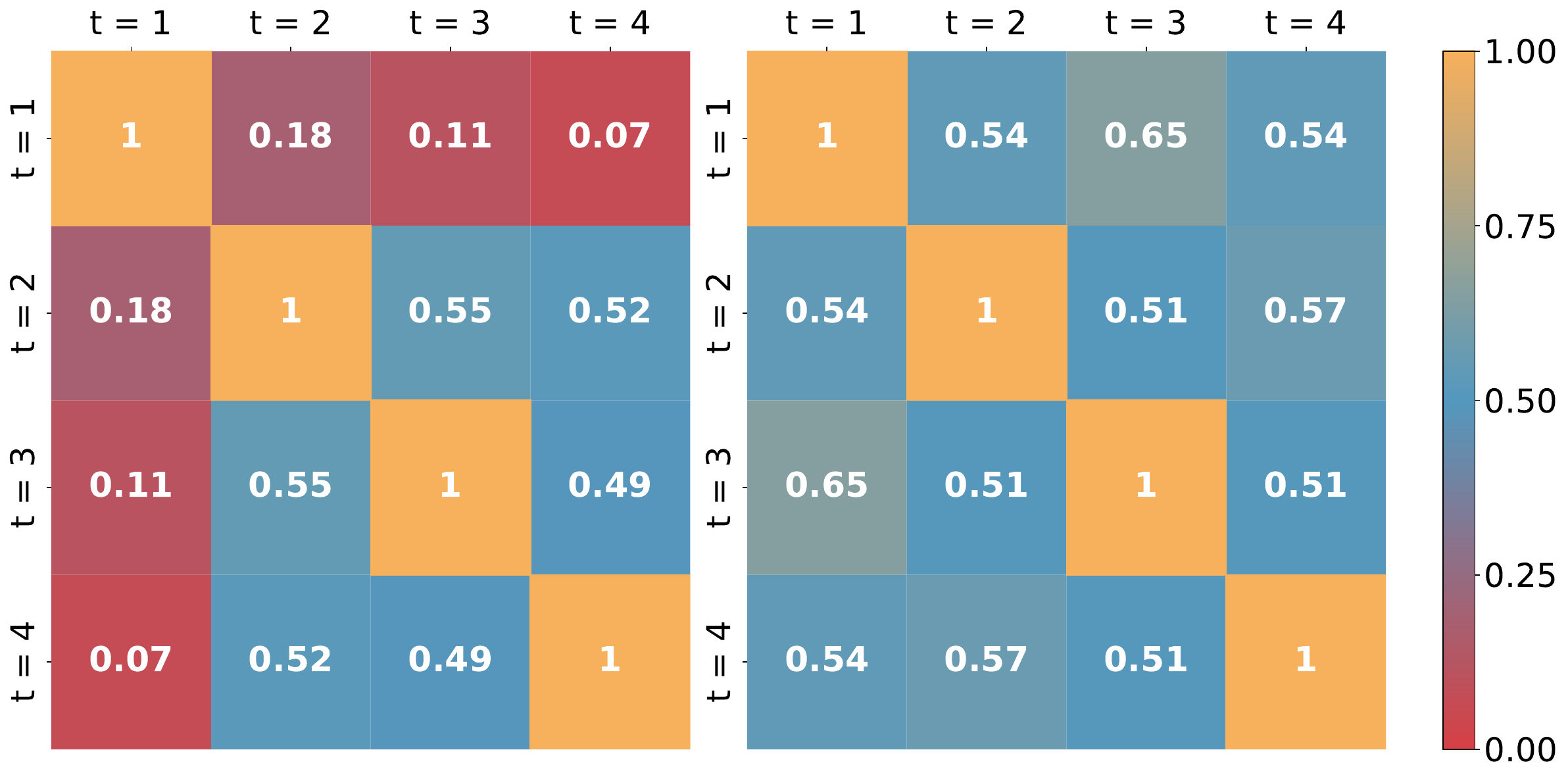}
    \caption{\textbf{Gradient conflicts across timesteps with and without MP-Init.} 
We visualize the cosine similarity of gradients across timesteps in ResNet-19 on CIFAR100. 
\emph{Left:} Without MP-Init, the gradient from $t=1$ strongly conflicts with later timesteps, reflecting the temporal inconsistency highlighted by MPS~\cite{ding:2025}. 
\emph{Right:} With MP-Init, conflicts are largely resolved and similarities improve, indicating that MP-Init aligns temporal ensemble members and stabilizes training.}
    \label{fig:gradient_conflict_tcs}
\end{figure}

\subsection{How does MP-Init impact training?}

TEBN~\cite{tebn:2022} and TAB~\cite{tab:2024} have shown that temporal covariate shift (TCS) indeed occurs, and our paper has already demonstrated that MP-Init effectively resolves it. However, prior studies did not provide clear evidence of \emph{how} TCS directly impairs training. Inspired by the ensemble interpretation of SNNs in MPS~\cite{ding:2025}, we revisited this question and conducted an experiment on ResNet-19 with CIFAR100. Specifically, we computed the gradients applied to a given layer at each timestep and measured the cosine similarity across timesteps. As shown in \cref{fig:gradient_conflict_tcs}, without MP-Init (left), the gradients from $t=1$ exhibit strong conflicts with later timesteps, indicating severe inconsistency across the temporal ensemble members. With MP-Init (right), these conflicts are largely alleviated, and the overall cosine similarity improves. This analysis provides the direct evidence that TCS not only degrades inference consistency but also hinders gradient alignment during training, and highlights how MP-Init effectively stabilizes optimization.

\begin{figure}[t]
  \centering
  \begin{subfigure}[b]{0.99\linewidth}
    \centering
    \includegraphics[width=\linewidth]{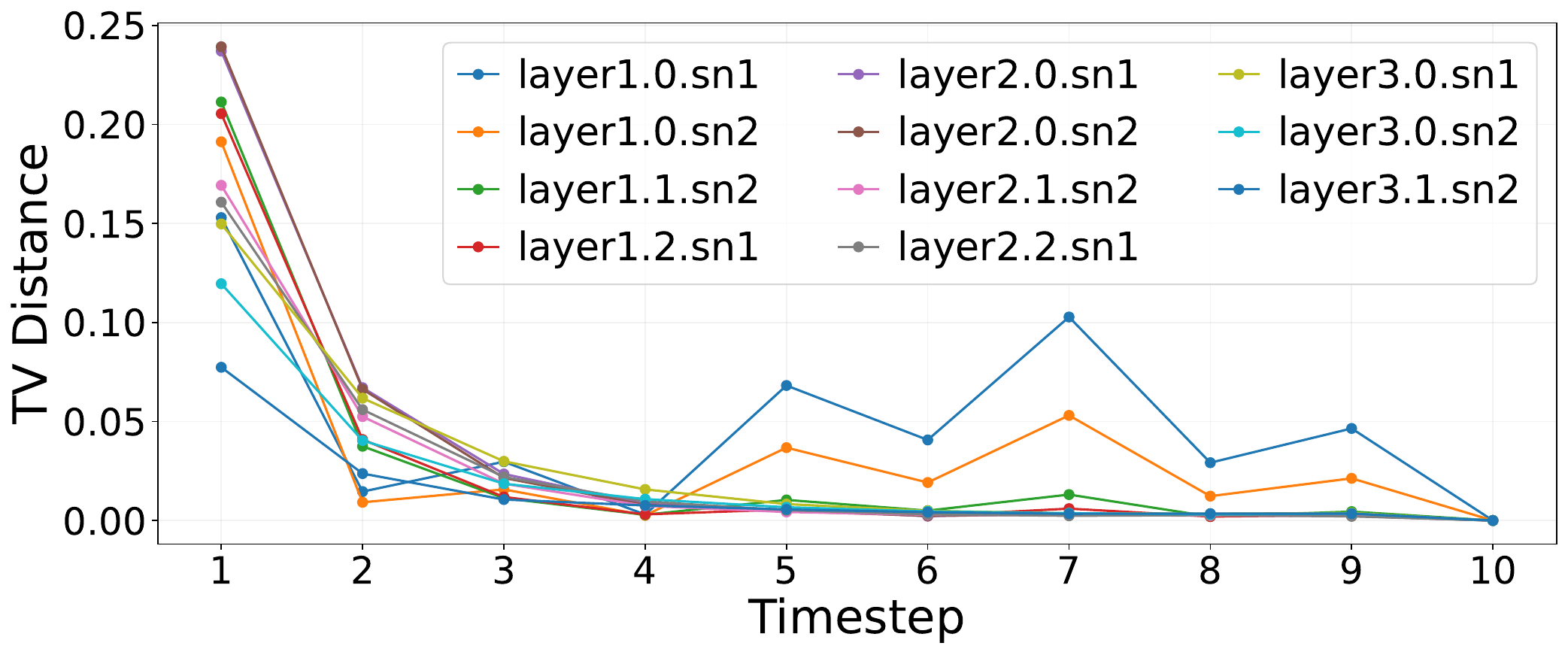}
    \caption{TV distance between input distributions at different timesteps. Each curve compares $t=1\ldots9$ against the reference at $t=10$.}
    \label{fig:tv-distance}
  \end{subfigure}
  \hfill
  \begin{subfigure}[b]{0.99\linewidth}
    \centering
    \includegraphics[width=\linewidth]{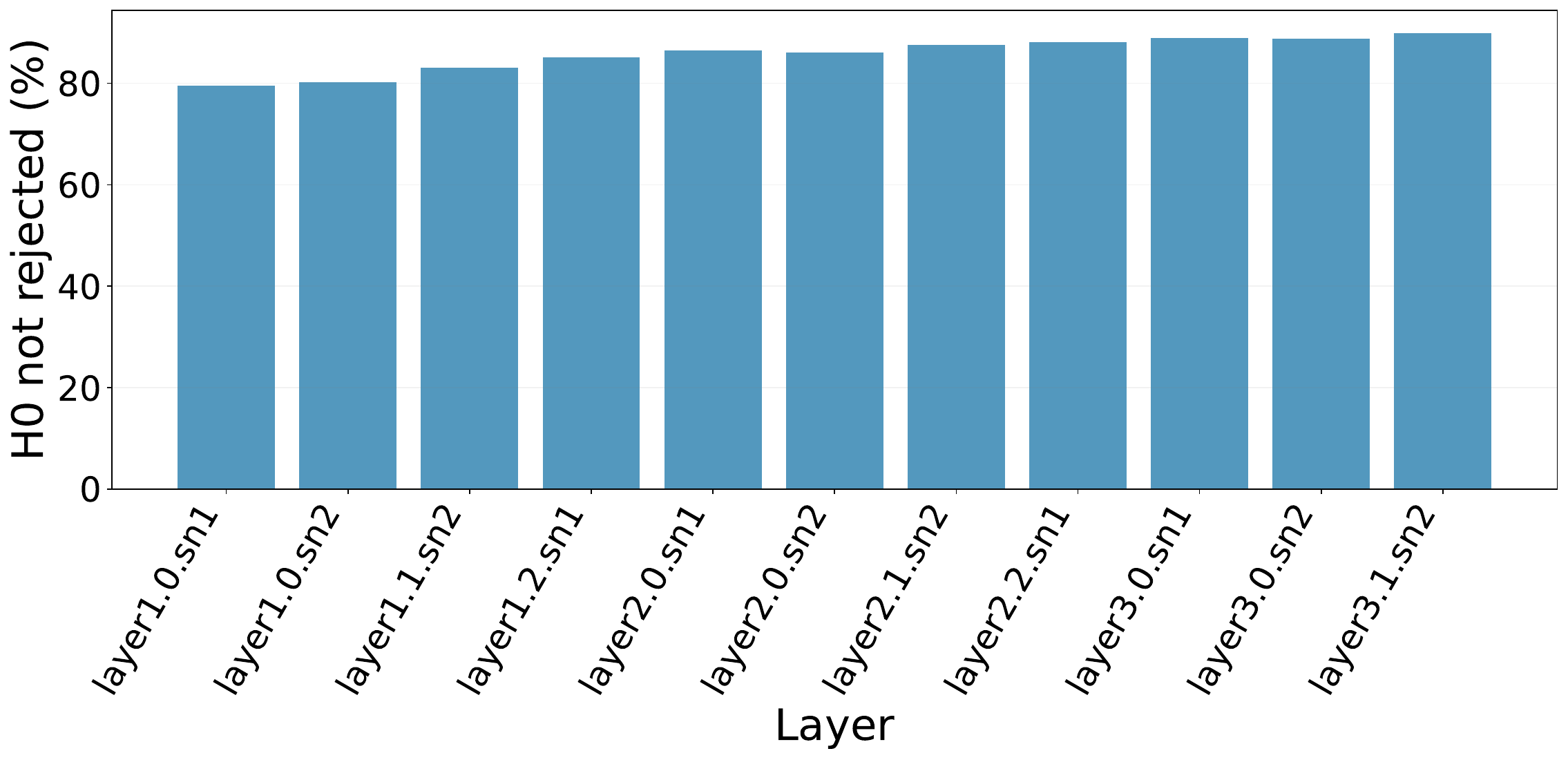}
    \caption{Ljung-Box test at lag=1. Bars show the percentage of neurons for which independence could not be rejected.}
    \label{fig:ljung-box}
  \end{subfigure}
  \hfill
  \begin{subfigure}[b]{0.99\linewidth}
    \centering
    \includegraphics[width=\linewidth]{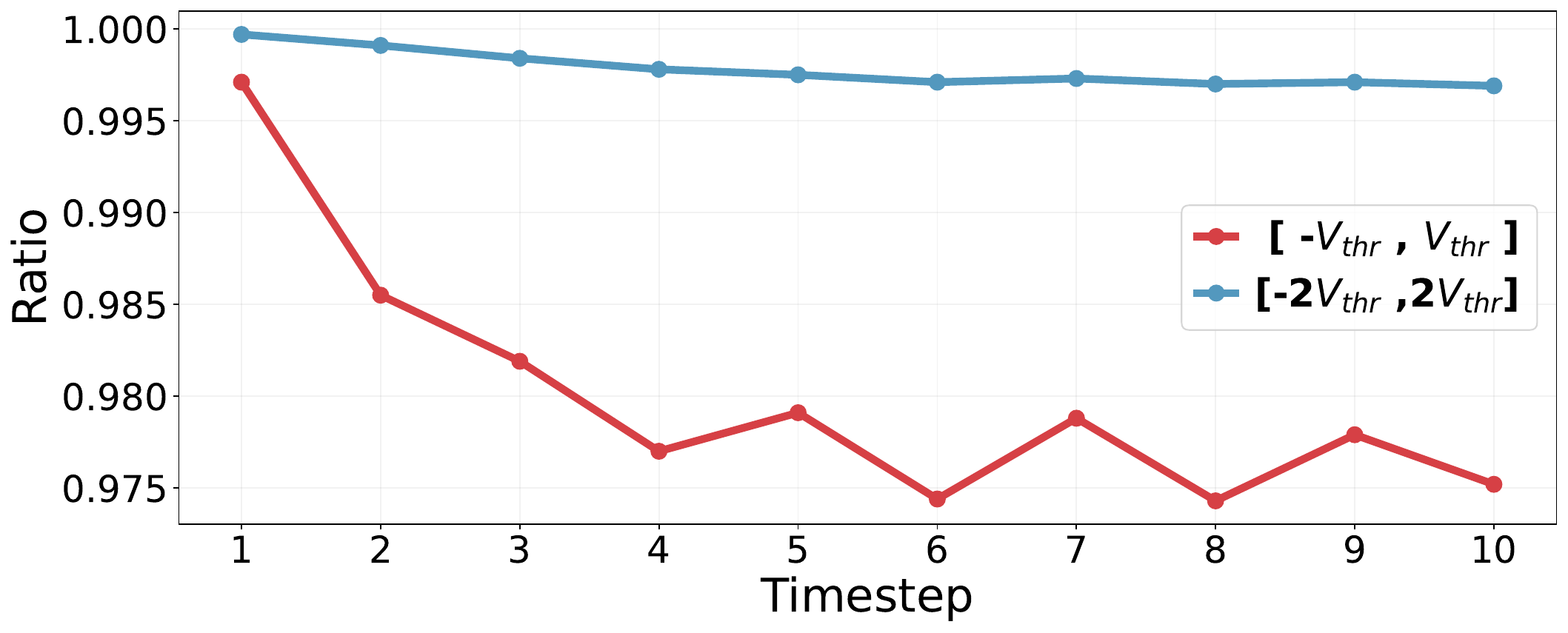}
    \caption{Boundedness ratio of membrane potentials $U[t]$.}
    \label{fig:ratio-bounded}
  \end{subfigure}

  \caption{Empirical validation of Assumption~1 using ResNet-19 trained on \textbf{CIFAR100}. 
  (a) TV distance across timesteps demonstrates near-identical input distributions. 
  (b) Ljung-Box test confirms approximate input independence. 
  (c) Boundedness ratio shows that membrane potentials remain within symmetric ranges relative to the threshold.}
  \label{fig:validation-assumption}
\end{figure}

\subsection{Validation of Assumption 1}
\label{subsec:validation_of_assumption_1}

We empirically verify that inputs are approximately \emph{i.i.d.} and that the membrane potentials of active neurons remain bounded, in line with Assumption~1. Using a trained ResNet-19 on CIFAR100, we report three complementary metrics:

\begin{itemize}
    \item \textbf{Identical distribution (\cref{fig:tv-distance}).} We compute the TV distance between input distributions at timesteps $t=1\ldots9$ and the reference at $t=10$. Distances are consistently small ($<0.1$ after $t=2$), indicating that the distribution quickly stabilizes and remains nearly identical across timesteps.
    \item \textbf{Independence (\cref{fig:ljung-box}).} We applied the Ljung-Box test for autocorrelation with significance level $\alpha=0.05$ using $\text{lag}=1$. It illustrates 80--90\% of neurons do not reject $H_0$ (independence). This suggests that temporal correlations are minor, supporting practical independence.
    \item \textbf{Boundedness (\cref{fig:ratio-bounded}).} Over 97\% of membrane potentials remain within $\pm V_{\text{thr}}$ at all timesteps, and nearly all values lie within $\pm 2V_{\text{thr}}$, confirming that the dynamics are effectively bounded in practice.
\end{itemize}

\begin{figure}[h]
  \centering
  \begin{subfigure}[b]{0.99\linewidth}
    \centering
    \includegraphics[width=\linewidth]{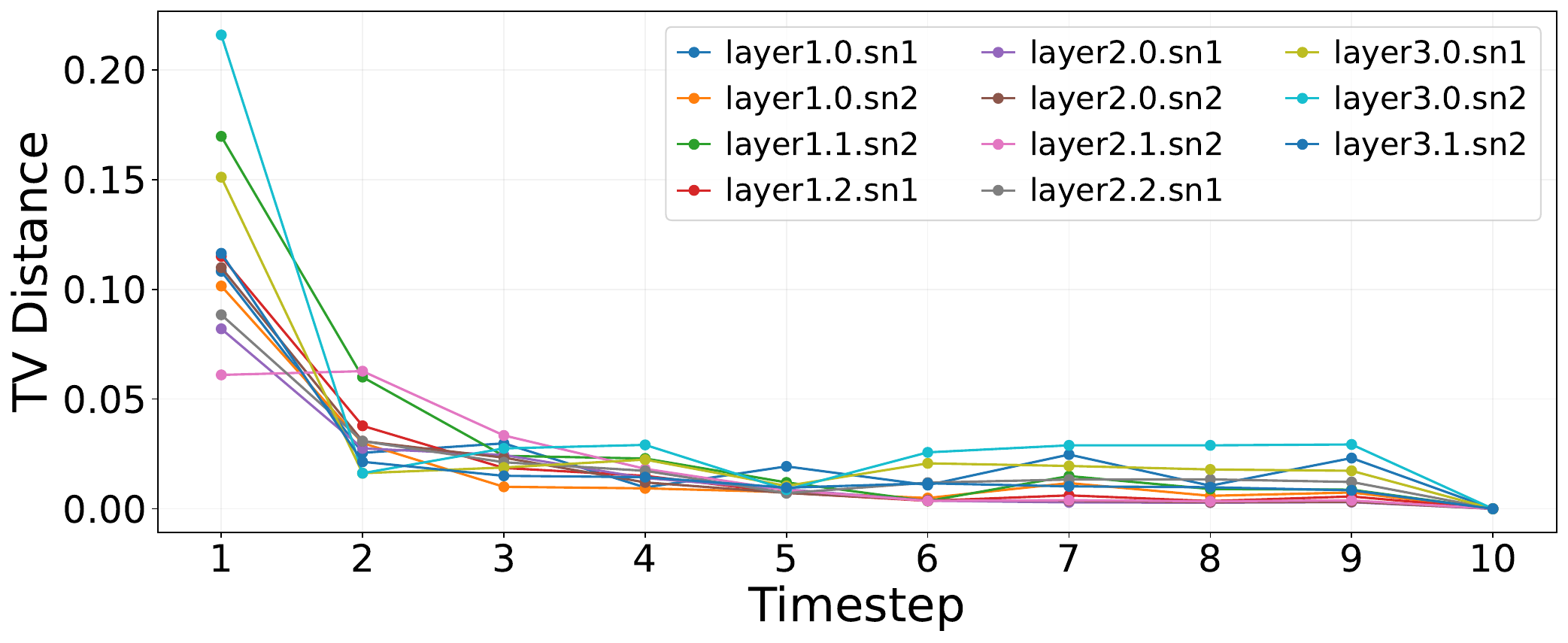}
    \caption{TV distance between input distributions at different timesteps. Each curve compares $t=1\ldots9$ against the reference at $t=10$.}
    \label{fig:tv-distance-dvs}
  \end{subfigure}
  \hfill
  \begin{subfigure}[b]{0.99\linewidth}
    \centering
    \includegraphics[width=\linewidth]{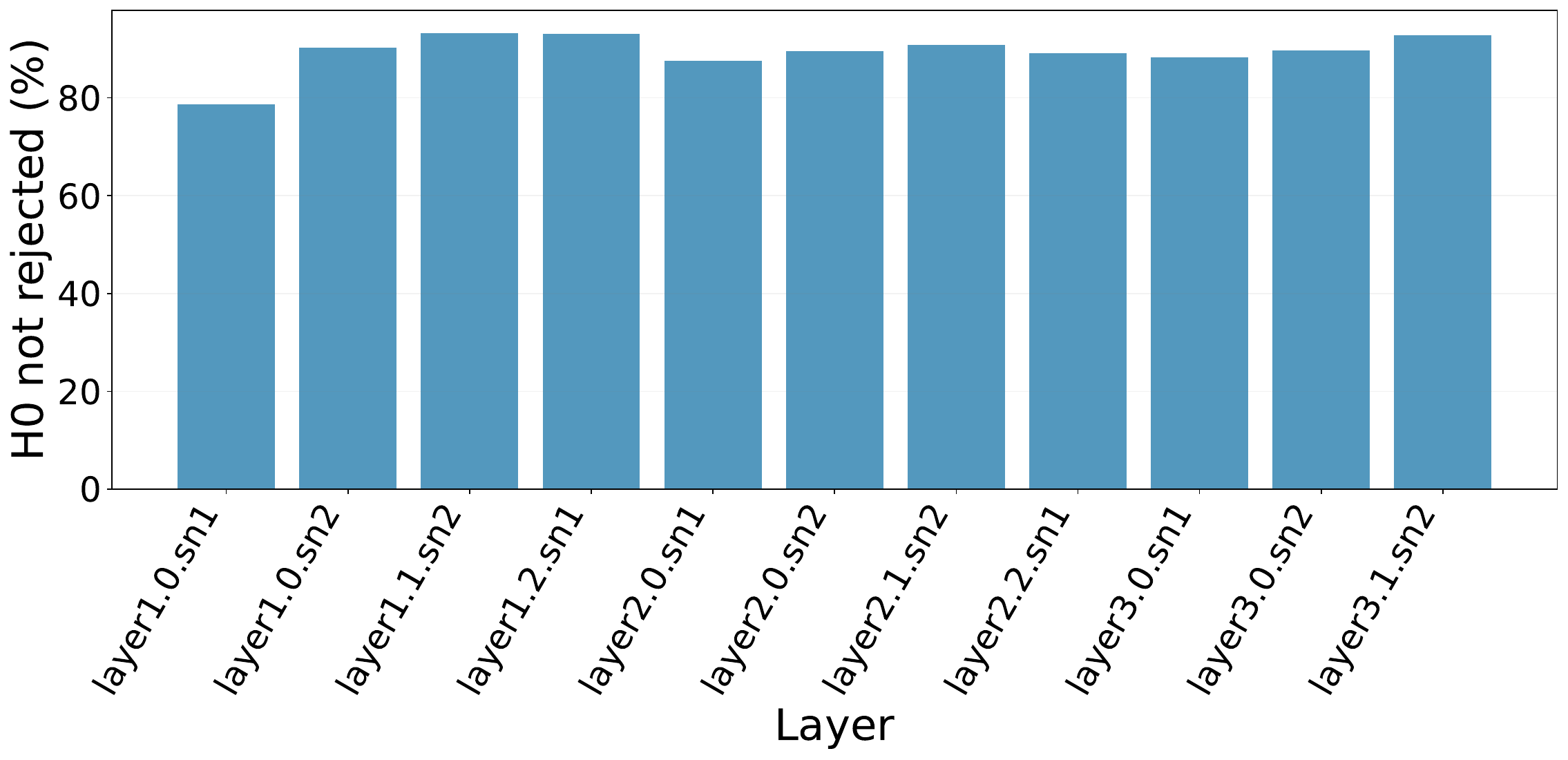}
    \caption{Ljung-Box test at lag=1. Bars show the percentage of neurons for which independence could not be rejected.}
    \label{fig:ljung-box-dvs}
  \end{subfigure}
  \hfill
  \begin{subfigure}[b]{0.99\linewidth}
    \centering
    \includegraphics[width=\linewidth]{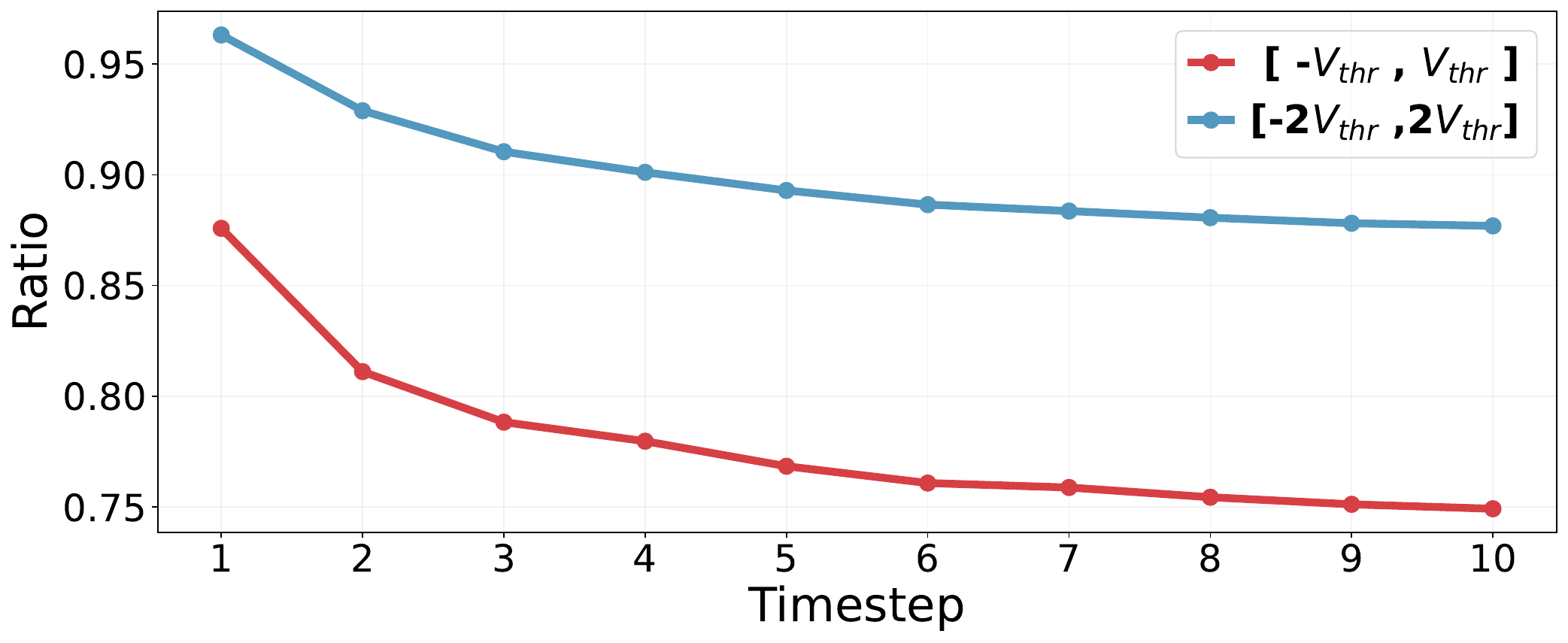}
    \caption{Boundedness ratio of membrane potentials $U[t]$.}
    \label{fig:ratio-bounded-dvs}
  \end{subfigure}

  \caption{Empirical validation of Assumption~1 using ResNet-19 trained on \textbf{DVS-CIFAR10}. 
  (a) TV distance across timesteps demonstrates near-identical input distributions. 
  (b) Ljung-Box test confirms approximate input independence. 
  (c) Boundedness ratio shows that membrane potentials remain within symmetric ranges relative to the threshold.}
  \label{fig:validation-assumption-dvs}
\end{figure}

Taken together, these results confirm that Assumption~1 holds empirically: the inputs to spiking layers are nearly i.i.d., and the membrane potentials of active neurons remain bounded, thereby validating the theoretical conditions. Consistent findings are observed on DVS-CIFAR10 as well, as shown in \cref{fig:validation-assumption-dvs}, where ResNet-19 exhibits the same behavior satisfying all parts of Assumption~1.

\begin{figure}[h]
  \centering
  \begin{subfigure}[b]{0.99\linewidth}
    \centering
    \includegraphics[width=\linewidth]{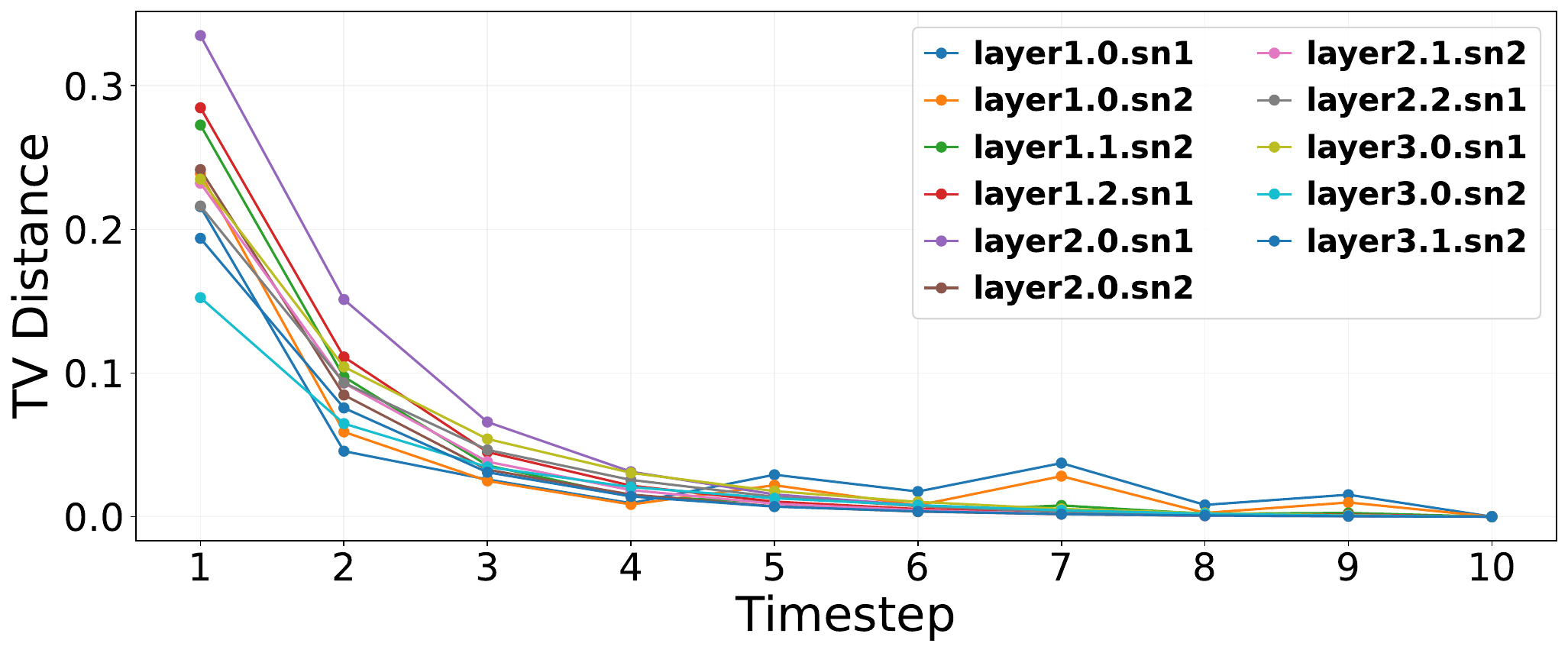}
    \caption{CIFAR100 (ResNet-19).}
    \label{fig:tv-mp-convergence-distance-cifar100}
  \end{subfigure}
  \hfill
  \begin{subfigure}[b]{0.99\linewidth}
    \centering
    \includegraphics[width=\linewidth]{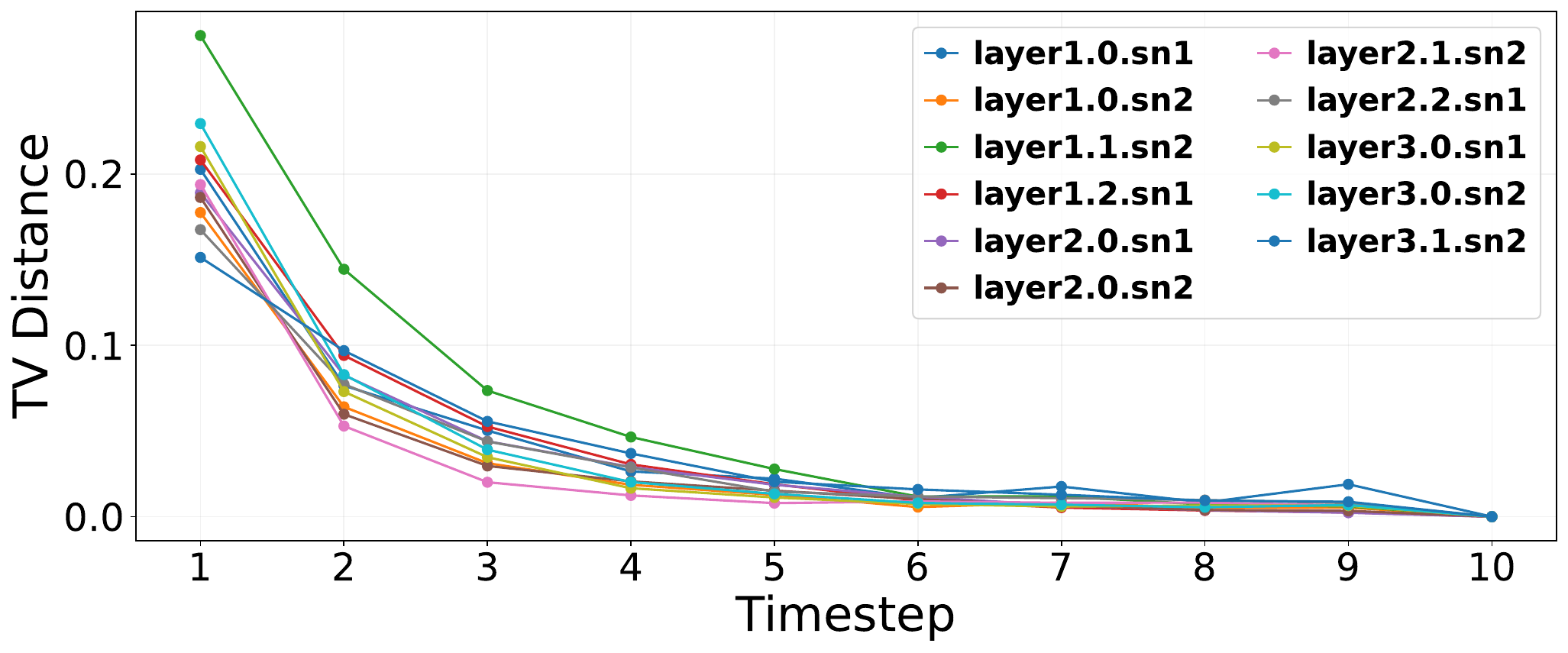}
    \caption{DVS-CIFAR10 (ResNet-19).}
    \label{fig:tv-mp-convergence-distance-dvs}
  \end{subfigure}
  \caption{TV distance between membrane potentials $U^l[t]$ at each timestep and the reference at $t=10$. Across both CIFAR100 and DVS-CIFAR10, TV distance is initially large ($>0.3$ at $t=1$) but decreases sharply by $t=2$, and falls below $0.1$ by $t=3$, indicating rapid convergence of membrane dynamics.}
  \label{fig:tv-distance-convergence}
\end{figure}

\begin{table}[h]
\centering
\small
\begin{tabular}{lcc}
\toprule
\textbf{Method} & \textbf{w/o MP-Init (\%)} & \textbf{w/ MP-Init (\%)} \\
\midrule
tdBN (timestep=2) & 72.35 $\pm$ 0.18 & \textbf{74.01 $\pm$ 0.33} \\
\bottomrule
\end{tabular}
\caption{Effect of MP-Init at early timesteps on CIFAR100 using ResNet-19}
\label{tab:convergence-mpinit}
\end{table}

\subsection{Convergence of Membrane Potentials}
\label{subsec:convergence_speed}

As predicted by Theorem~1, the empirical TV distance curves in \cref{fig:tv-distance-convergence} clearly exhibit \emph{exponential convergence}: deviations are large at $t=1$ but shrink rapidly, already becoming negligible ($<0.1$) by $t=3$. This rapid convergence is highly desirable for SNNs, since reducing the number of timesteps directly improves both energy efficiency and latency. Notably, the benefit is especially pronounced at very low timesteps: at timestep=2, MP-Init delivers a \textbf{1.66\%pt} accuracy improvement over the baseline (\cref{tab:convergence-mpinit}), highlighting that our method enables reliable inference even under aggressive latency constraints.

\subsection{Proof of Theorem 1}
\label{subsec:proof_of_theorem_1}
\textbf{Theorem 1.}
Under Assumption 1, the sequence $\{U[t]\}_{t \ge 0}$ forms a Markov chain with transition kernel $P$. This chain satisfies Doeblin’s minorization condition, ensuring the existence of a unique stationary distribution $\pi$. Moreover, the TV distance between $U[t]$ and $\pi$ decreases exponentially.

\medskip
\noindent
\textbf{Step 1: Markov Property}

We begin by confirming that \( \{U[t]\}_{t \geq 0} \) forms a Markov chain. The dynamics of \( U[t] \) are given by:
\begin{equation}
    U[t+1] = f(U[t], I_{in}[t+1]),
\end{equation}
where \( f(\cdot) \) is a deterministic function dependent on the current membrane potential \( U[t] \) and the new input \( I_{in}[t+1] \). Since the inputs \( \{I_{in}[t]\}_{t \geq 0} \) follow i.i.d, the future state \( U[t+1] \) depends solely on the current state \( U[t] \) and not on any earlier states.

Thus, the probability of transitioning to \( U[t+1] \) given the entire history \( \{U[s]\}_{s \leq t} \) depends only on \( U[t] \):
\begin{equation}
    P(U[t+1] \in A \mid U[t], \dots, U[0]) = P(U[t+1] \in A \mid U[t]),
\end{equation}
which confirms that \( \{U[t]\}_{t \geq 0} \) is indeed a Markov chain.

\medskip
\noindent
\textbf{Step 2: The Transition Kernel \( P(x, A) \) and Its Relation} 

The transition kernel \( P(x, A) \) describes the probability of moving from state \( x \) at time \( t \) to the set \( A \) at time \( t+1 \). Specifically, for the Markov chain \( \{U[t]\}_{t \geq 0} \), \( P(x, A) \) is the probability that the membrane potential at time \( t+1 \), denoted by \( U[t+1] \), falls within the set \( A \) given that the membrane potential at time \( t \) is \( U[t] = x \). This is formulated as:
\begin{equation}
    P(x, A) = P(U[t+1] \in A \mid U[t] = x).
\end{equation}

The transition kernel can be represented using the transition density \( p(x, y) \), which is the probability density function for transitioning from state \( x \) to state \( y \):
\begin{equation}
    P(x, A) = \int_A p(x, y) dy,
\end{equation}
where \( p(x, y) \) encapsulates the dynamics governed by the function \( f(U[t], I_{in}[t+1]) \) and the input \( I_{in}[t+1] \).

\medskip
\noindent
\textbf{Step 3: Doeblin's Minorization Condition}

We now show that the Markov chain \( \{U[t]\}_{t \geq 0} \) satisfies the Doeblin's minorization condition~\cite{rosenthal:1995}. We need to demonstrate the existence of a constant \( \epsilon > 0 \) and a probability measure \( \mu(\cdot) \) such that for any \( x \in S \) and any measurable set \( A \subseteq S \), the transition probability satisfies:
\begin{equation}
\label{eq:minimorization_condition}
    P(x, A) \geq \epsilon \mu(A).
\end{equation}

Given that the state space \( S \) is bounded by \( [u^{-}, u^{+}] \), and assuming the transition density \( p(x, y) \) is smooth and continuous (as the inputs are i.i.d. Gaussian), there exists a positive lower bound \( \delta \) for \( p(x, y) \) over the state space \( S \). This implies that for any measurable set \( A \subseteq S \):
\begin{equation}
    P(x, A) = \int_A p(x, y) dy \geq \delta \cdot \mu(A),
\end{equation}
where \( \mu(\cdot) \) is a probability measure (e.g., uniform measure) over \( S \). Setting \( \epsilon = \delta \), we can validate that \cref{eq:minimorization_condition} is satisfied.

\medskip
\noindent
\textbf{Step 4: Exponential Decrease in TV Distance}

As a result of the Doeblin's minorization condition, the TV distance between the distribution \( P^n(x, \cdot) \) of the Markov chain after \( n \) steps, starting from any state \( x \), and the stationary distribution \( \pi(\cdot) \) satisfies:
\begin{equation}
\label{eq:convergence_speed}
    \| P^n(x, \cdot) - \pi(\cdot) \|_{TV} \leq C(1 - \epsilon)^n,
\end{equation}
for some constant \( C > 0 \). This inequality shows that the TV distance decreases exponentially as \( n \) increases, implying that the Markov chain converges rapidly to the stationary distribution \( \pi \). This also implies the uniqueness and existence of the stationary distribution.

\subsection{Proof of Lemma 1}
\label{subsec:proof_of_lemma_1}
\textbf{Lemma 1.} Among all constants \( c \in \mathbb{R} \), \( c = E[\pi] \) minimizes expected square difference \(E[(\pi-c)^{2}]\).

\medskip

We begin by calculating the expected value of the L2 norm between \( U^{*} \) and a constant \( c \):

\begin{equation}
    E[|U^{*} - c|^{2}] = E[(U^{*} - c)^{2}] = E[U^{*2}] - 2E[U^{*}]c + c^2.
\end{equation}

Next, we take the derivative of this expression with respect to \( c \):

\begin{equation}
    \frac{\partial E[|U^{*} - c|^{2}]}{\partial c} = -2E[U^{*}] + 2c.
\end{equation}

Setting the derivative to zero to find the minimum, we obtain:

\begin{equation}
    c = E[U^{*}].
\end{equation}

Thus, the L2 norm is minimized when \( c = E[U^{*}] \), proving the lemma.

\section{Additional Details: TrSG}
\label{sec:appendix_trsg}

\subsection{How does TrSG impact training?}
\label{subsec:how_trsg_impacts_training}

\paragraph{Protocol.}
We analyze gradient flow of ResNet-19 on CIFAR100 with $\text{timestep}{=}4$ for 200 epochs. To isolate the effect of the SG formulation, we \emph{fix} $\tau{=}2.0$ and do not train $V_{\text{thr}}$ or $\tau$. We use a rectangular surrogate with $\gamma{=}1$ and the soft-reset mechanism.

\paragraph{Metrics.}
We quantify stability with three complementary metrics computed on convolutional weights of a given layer:
\[
\text{AbsStr}=\operatorname*{mean}_i |g_i|,\quad
\]
\[
\text{RatioAG}=\frac{1}{N}\sum_i \mathds{1}\!\left(|x_i|<\tfrac{\gamma}{2}\right),\quad
\]
\[
\text{GradCV}=\frac{\operatorname{std}_i(|g_i|)}{\operatorname*{mean}_i(|g_i|)},
\]
where $g_i=\partial \mathcal{L}/\partial W_i$ and $x_i$ is the surrogate argument. 

\textbf{AbsStr} measures the average gradient magnitude. Very low values indicate vanishing gradients, where updates stall, 
while excessively high values point to exploding gradients, leading to erratic updates. 
Stable training requires AbsStr to remain in a moderate range. 
\textbf{RatioAG} quantifies the fraction of neurons receiving nonzero gradients. 
If RatioAG is too low, most neurons fail to receive updates (gradient starvation); 
if too high, too many neurons receive updates, causing gradient flood and poor selectivity. 
An intermediate range is thus ideal. 

\textbf{GradCV} captures the variability of gradient magnitudes. 
Low variance (CV $<1$) implies smooth, balanced updates that can be handled by a single learning rate, 
whereas high variance (CV $\gg 1$) means some weights update disproportionately faster than others, destabilizing training.  
TrSG is designed to maintain AbsStr at moderate levels, keep RatioAG in the healthy range, 
and reduce GradCV, together ensuring stable optimization.

\paragraph{Reference case ($V_{\text{thr}}{=}1.0$,~\cref{tab:metrics_ref}).}
This widely used threshold serves as a baseline where all SG variants behave same (though $1.0$ is not necessarily optimal). From epoch 1 to 100 we observe steadily improving yet stable gradients.

\begin{table}[t]
\centering
\small
\begin{tabular}{ccccc}
\toprule
\textbf{Epoch} & \textbf{$V_{\text{thr}}{=}1.0$} & \textbf{AbsStr} & \textbf{RatioAG (\%)} & \textbf{GradCV} \\
\midrule
1   & any SG & 0.0122 & 6.72  & 0.818 \\
100 & any SG & 0.0211 & 18.97 & 0.571 \\
\bottomrule
\end{tabular}
\caption{Reference case where all SG variants behave same (ResNet‑19, CIFAR100, $\text{timestep}{=}4$, $\tau{=}2.0$).}
\label{tab:metrics_ref}
\end{table}

\paragraph{Low threshold ($V_{\text{thr}}{=}0.1$,~\cref{tab:metrics_low}).}
This stress test exposes small-threshold failure modes. At epoch 1, AS‑SG floods (wide window), RS‑SG explodes due to the $1/V_{\text{thr}}$ factor, while TrSG remains stable. By epoch 100, AS‑SG still shows instability and RS‑SG diverges; TrSG trains smoothly. These trends align with accuracy outcomes in \cref{tab:no_training_results}.

\begin{table}[t]
\centering
\small
\begin{tabular}{ccccc}
\toprule
\textbf{Epoch} & \textbf{$V_{\text{thr}}{=}0.1$} & \textbf{AbsStr} & \textbf{RatioAG (\%)} & \textbf{GradCV} \\
\midrule
\multirow{3}{*}{1}   & AS-SG & 0.2756 & 31.25 & 5.01  \\
& RS-SG & 7.8191 & 0.01  & 37.11 \\
& TrSG  & 0.0046 & 3.05  & \textbf{0.861} \\
\addlinespace[2pt]
\midrule
\multirow{3}{*}{100} & AS-SG & 0.0145 & 90.20 & 1.22  \\
& RS-SG & NaN    & 0.00  & NaN   \\
& TrSG  & 0.0153 & 21.45 & \textbf{0.865} \\
\bottomrule
\end{tabular}
\caption{Low‑threshold stress test (CIFAR100, $\text{timestep}{=}4$, $\tau{=}2.0$). RS‑SG explodes (diverges), AS‑SG floods, while TrSG remains stable. See \cref{tab:no_training_results} for accuracy.}
\label{tab:metrics_low}
\end{table}

\paragraph{High threshold ($V_{\text{thr}}{=}2.0$,~\cref{tab:metrics_high}).}
Here the absolute window of AS‑SG is too narrow and RS‑SG’s $1/V_{\text{thr}}$ suppresses signals. AS‑SG starves, RS‑SG vanishes with high variance, while TrSG stays trainable. The accuracy pattern is consistent with \cref{tab:no_training_results}.

\begin{table}[t]
\centering
\small
\begin{tabular}{ccccc}
\toprule
\textbf{Epoch} & \textbf{$V_{\text{thr}}{=}2.0$} & \textbf{AbsStr} & \textbf{RatioAG (\%)} & \textbf{GradCV} \\
\midrule
\multirow{3}{*}{1} & AS-SG & 0.0058 & 0.78 & 17.14 \\
& RS-SG & 0.0066 & 1.64 & 0.639 \\
& TrSG  & 0.0092 & 1.62 & \textbf{0.961} \\
\midrule
\addlinespace[2pt]
\multirow{3}{*}{100} & AS-SG & 0.0000 & 0.00 & 0.000 \\
& RS-SG & 0.0021 & 0.10 & 2.247 \\
& TrSG  & 0.0323 & 6.90 & \textbf{0.731} \\
\bottomrule
\end{tabular}
\caption{High‑threshold stress test (CIFAR100, $\text{timestep}{=}4$, $\tau{=}2.0$). AS‑SG starves, RS‑SG vanishes, TrSG preserves stable updates. See \cref{tab:no_training_results} for accuracy.}
\label{tab:metrics_high}
\end{table}

\begin{table}[t]
\centering
\small
\begin{tabular}{cc|ccccc}
\toprule
\textbf{Reset} & \textbf{SG} & \multicolumn{4}{c}{\(\boldsymbol{V_{\text{thr}}}\)} \\
\cmidrule(lr){3-7}
\multicolumn{2}{c|}{} & 0.1 & 0.5 & 1.0 & 1.5 & 2.0 \\
\midrule
\multirow{3}{*}{\textbf{Soft}} 
  & AS-SG        & 61.59 & 75.96 & ---  & 69.97  & 18.09 \\
  & RS-SG        & \emph{div.} & 76.12 & ---   & 71.43  & 5.43 \\
  & \textbf{TrSG}  & \textbf{73.99} & \textbf{76.82} & \textbf{75.56} & \textbf{71.97}  & \textbf{64.27} \\
\midrule
\multirow{3}{*}{\textbf{Hard}} 
  & AS-SG        & 67.71   & \textbf{76.34} & ---   & 68.42  & 12.27 \\
  & RS-SG        & 10.36   & 75.85 & ---   & 72.51  & 52.10 \\
  & \textbf{TrSG}  & \textbf{68.94}   & 76.33 & \textbf{75.72} & \textbf{73.86}  & \textbf{68.42} \\
\bottomrule
\end{tabular}
\caption{
Accuracy (\%) with fixed $\tau{=}2.0$ and \emph{frozen} $V_{\text{thr}}$ and $\tau$ to isolate SG effects. 
We vary $V_{\text{thr}}\!\in\!\{0.1,0.5,1.0,1.5,2.0\}$ under soft/hard resets. 
Conventional SGs fail outside a narrow range (explosion at $0.1$, starvation/vanishing at $2.0$), while \textbf{TrSG} remains robust across thresholds.}
\label{tab:no_training_results}
\end{table}

\begin{figure}[h]
  \centering
  \begin{subfigure}[b]{0.99\linewidth}
    \centering
    \includegraphics[width=\linewidth]{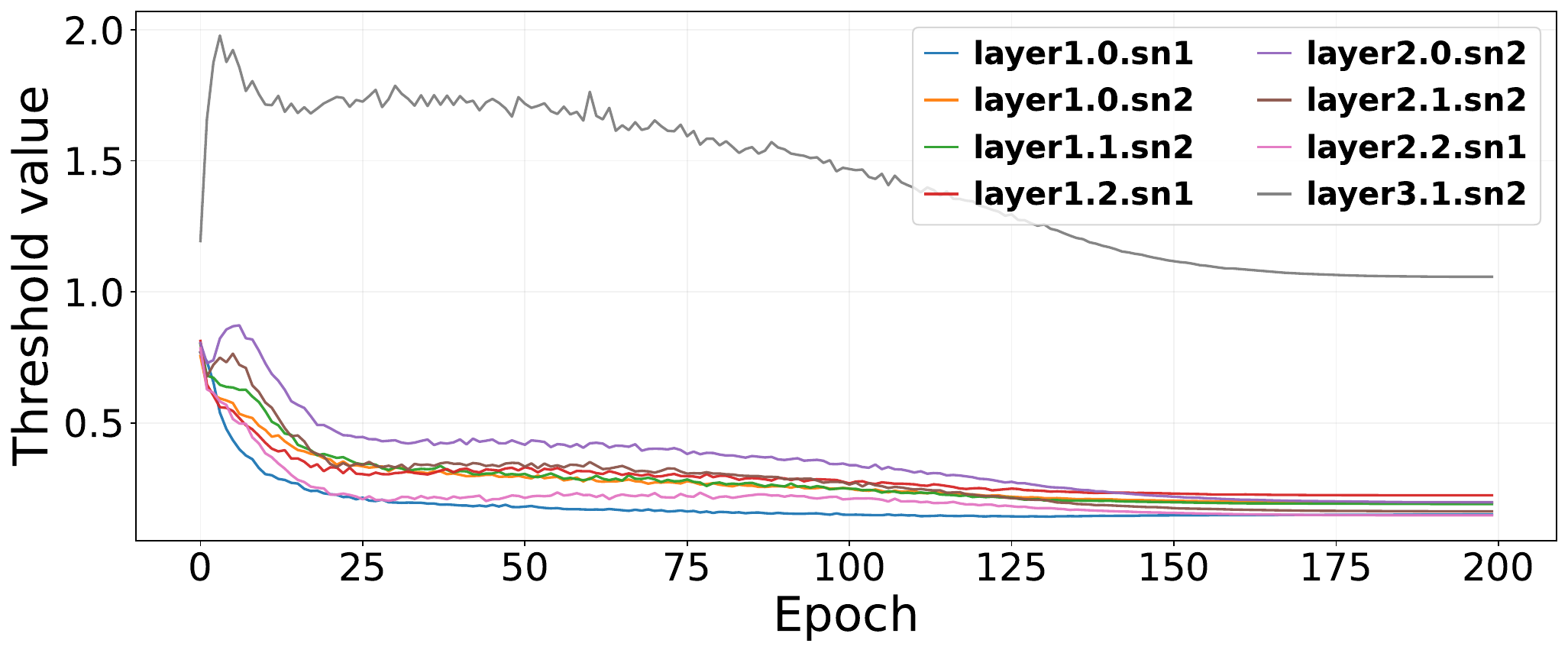}
    \caption{ResNet-19 on CIFAR100}
  \end{subfigure}
  \hfill
  \begin{subfigure}[b]{0.99\linewidth}
    \centering
    \includegraphics[width=\linewidth]{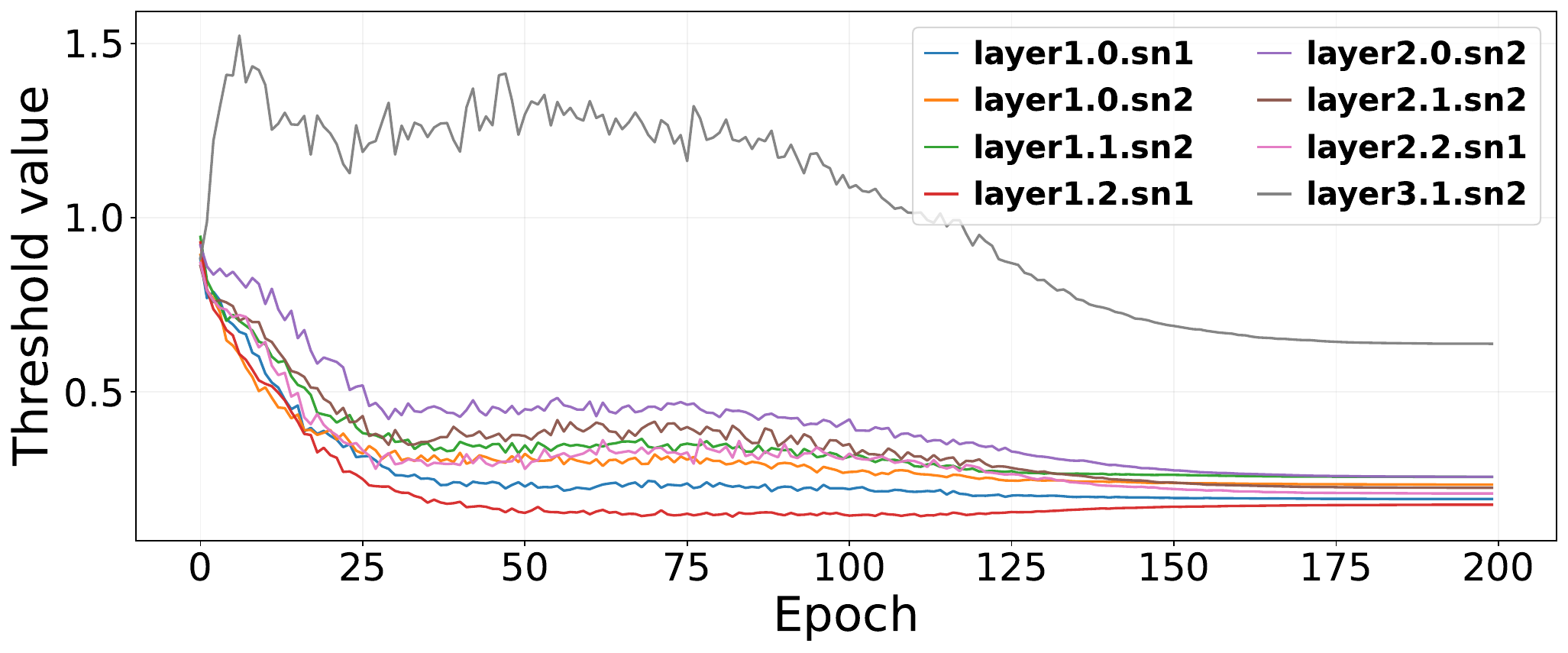}
    \caption{ResNet-19 on DVS-CIFAR10}
  \end{subfigure}
  \hfill
  \begin{subfigure}[b]{0.99\linewidth}
    \centering
    \includegraphics[width=\linewidth]{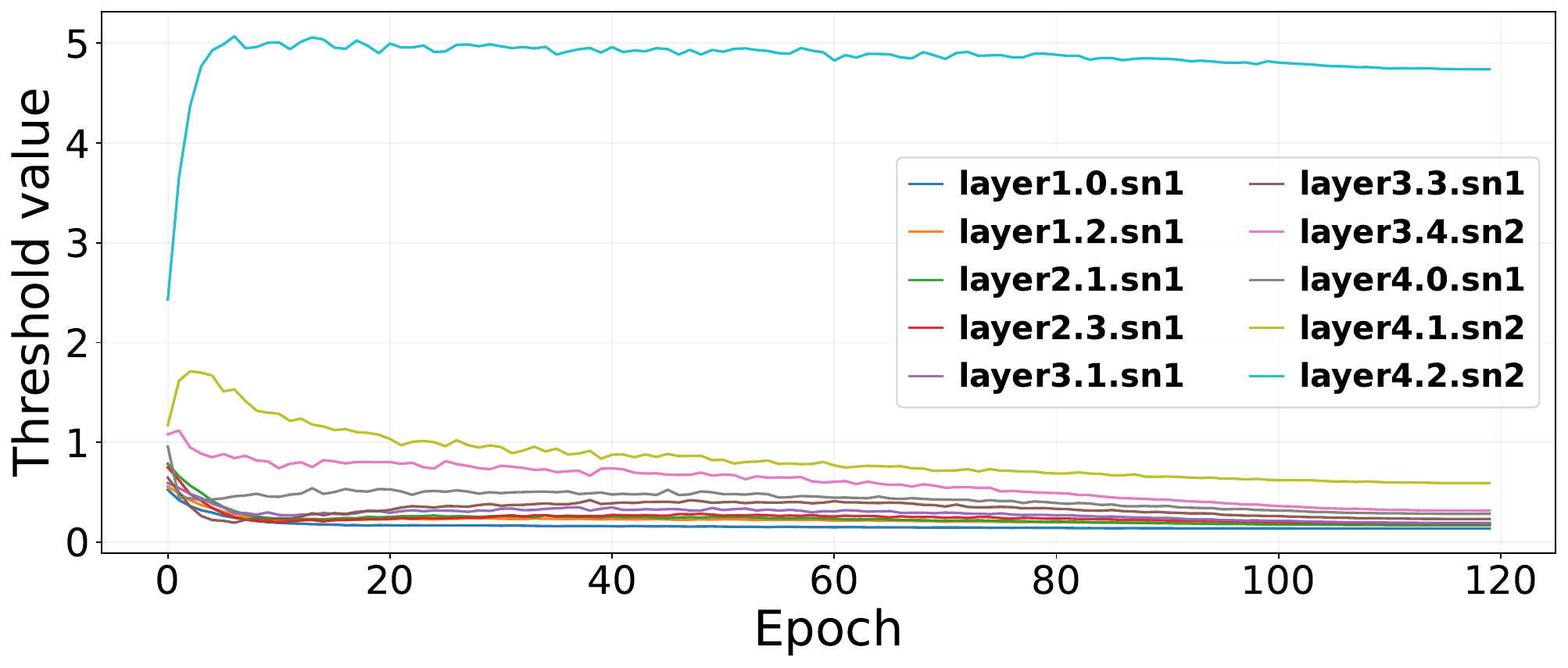}
    \caption{ResNet-34 on ImageNet}
  \end{subfigure}
  \caption{\textbf{Trajectories of trainable $V_{\text{thr}}$ during training.} 
  Early layers tend to settle at lower thresholds, while late/deep layers increase toward larger thresholds.}
  \label{fig:vthr_dynamics}
\end{figure}

\paragraph{Do real trainings visit such extreme thresholds?}
Yes. When $V_{\text{thr}}$ is trainable, many layers drift toward \emph{lower} thresholds (often $\ll1$), while late/deep layers move \emph{upward} toward larger values (around $\sim 2$ on CIFAR100 and even higher on ImageNet). 
\Cref{fig:vthr_dynamics} summarizes these trajectories: CIFAR100 and DVS-CIFAR10 show several early layers decreasing below $1$ and the last layer increasing, whereas ImageNet exhibits a wider spread with some deep layers rising well beyond $2$. 
These observations confirm that the brittle regimes identified for AS-SG/RS-SG are indeed encountered during training, precisely where TrSG’s threshold-robustness is most beneficial.

\paragraph{Implications:}

\begin{itemize}
\item \cref{tab:metrics_ref,tab:metrics_low,tab:metrics_high} explain \emph{why} TrSG is superior: it maintains moderate magnitude (AbsStr), healthy selectivity (RatioAG), and low update variance (GradCV), avoiding flood/starvation and explosion/vanishing. 

\item The threshold trajectories in \cref{fig:vthr_dynamics} show that real trainings \emph{do} visit the very regimes that break AS-SG/RS-SG, so threshold-robustness is not a corner case but a practical necessity. 

\item \textbf{Even when $V_{\text{thr}}$ and $\tau$ are not trained}, TrSG remains the best choice across thresholds (\cref{tab:no_training_results}); using TrSG is strictly beneficial, with no downside in practice.
\end{itemize}

\subsection{Ablation on Advanced Architecture}
\label{subsec:ablation_large_transformer}

\begin{table}[h]
\centering
\small
\begin{tabular}{lc}
\toprule
\textbf{Method} & \textbf{Top-1 (\%)} \\
\midrule
QKFormer (HST-10-384) & 78.80 \\
\quad + AS-SG         & 77.09 \\
\quad + RS-SG         & 78.70 \\
\quad \textbf{+ TrSG} & \textbf{79.90} \\
\bottomrule
\end{tabular}
\caption{Ablation on surrogate gradient choice for QKFormer (HST-10-384) on ImageNet at 4 timesteps. 
All runs follow the official training recipe; only the SG variant differs. 
MP-Init is enabled in all cases.}
\label{tab:qkformer_ablation}
\end{table}

We ablate the surrogate gradient formulation on a large-scale Transformer SNN, QKFormer~\cite{zhou:2024_2}, using the official training protocol while keeping all settings fixed and changing only the SG type. With MP-Init enabled across the board, \textbf{TrSG} is the only variant that reliably improves over the baseline ($+1.10$ \%pt), whereas \textbf{AS-SG} underperforms ($-1.71$ \%pt) and \textbf{RS-SG} essentially matches or slightly underperforms ($-0.10$ \%pt). 

These results demonstrate that even on advanced Transformer-based SNNs, TrSG provides consistent benefits. More importantly, they show that using conventional SGs not only fails to deliver improvements but can actually degrade performance, underscoring TrSG as the superior and safer choice for training modern large-scale SNN architectures.

\begin{figure}[t]
  \centering
  \begin{subfigure}{0.99\linewidth}
    \centering
    \includegraphics[width=0.99\linewidth]{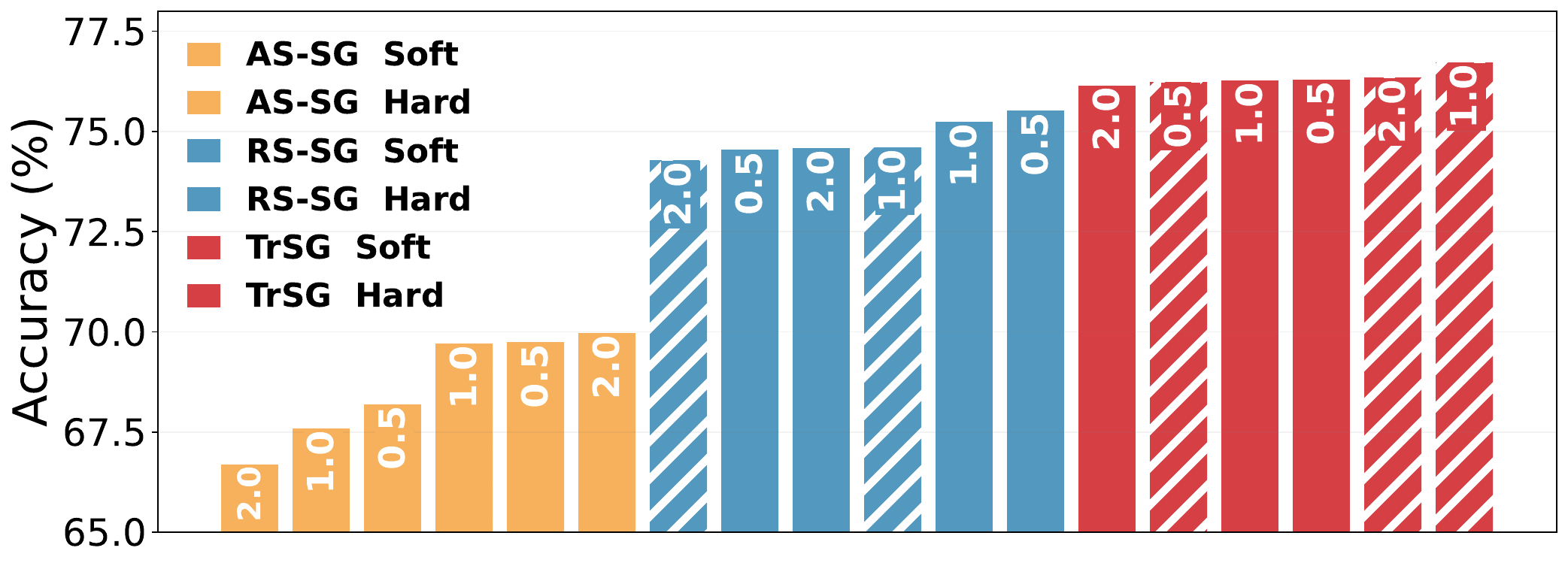}
    \caption{Arctan: init \(\tau=2.0,V_{\text{thr}}\in\{0.5,1.0,2.0\}\)}
  \end{subfigure}
  \begin{subfigure}{0.99\linewidth}
    \centering
    \includegraphics[width=0.99\linewidth]{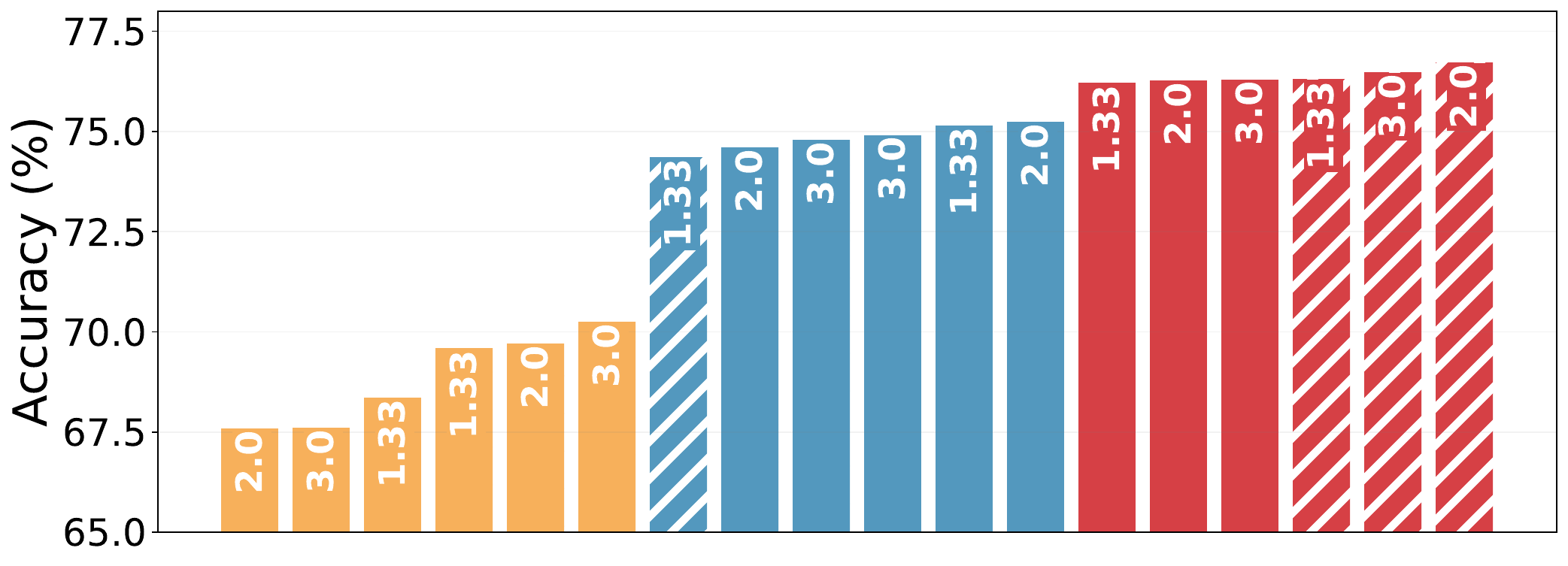}
    \caption{Arctan: init \(V_{\text{thr}}=1.0,\tau\in\{1.33,2.0,3.0\}\)}
  \end{subfigure}
  \begin{subfigure}{0.99\linewidth}
    \centering
    \includegraphics[width=0.99\linewidth]{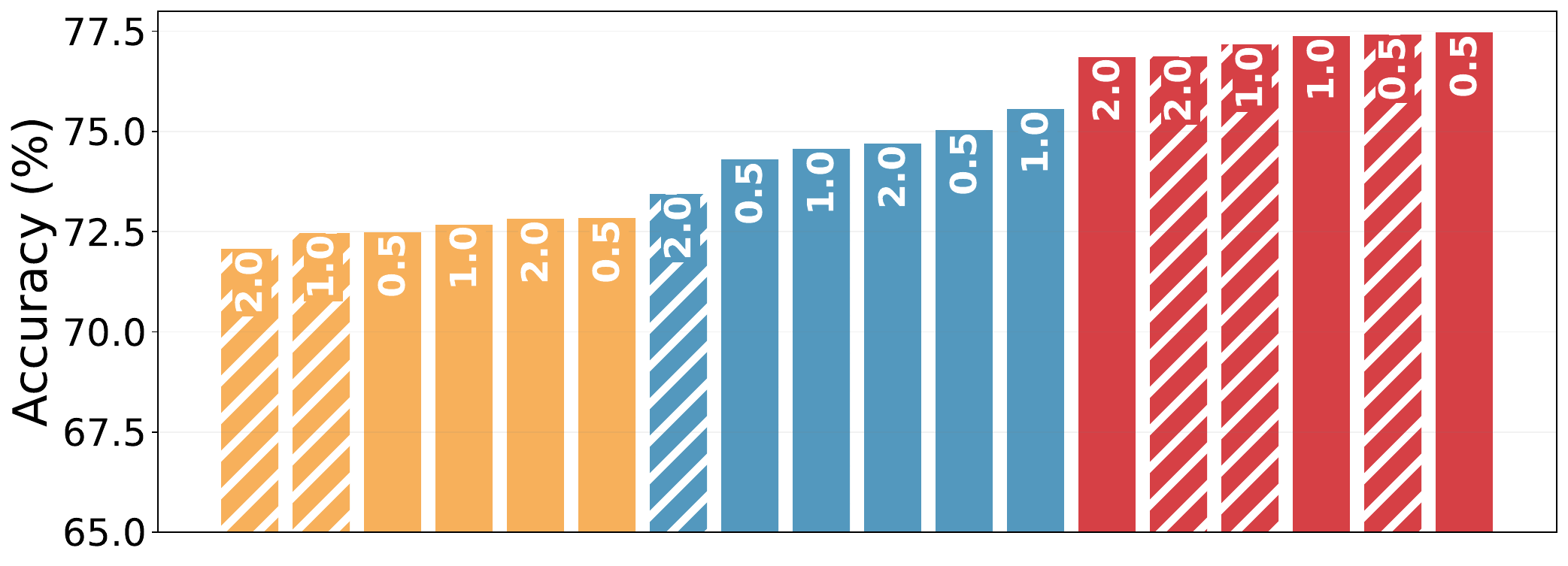}
    \caption{Sigmoid: init \(\tau=2.0,V_{\text{thr}}\in\{0.5,1.0,2.0\}\)}
  \end{subfigure}
  \begin{subfigure}{0.99\linewidth}
    \centering
    \includegraphics[width=0.99\linewidth]{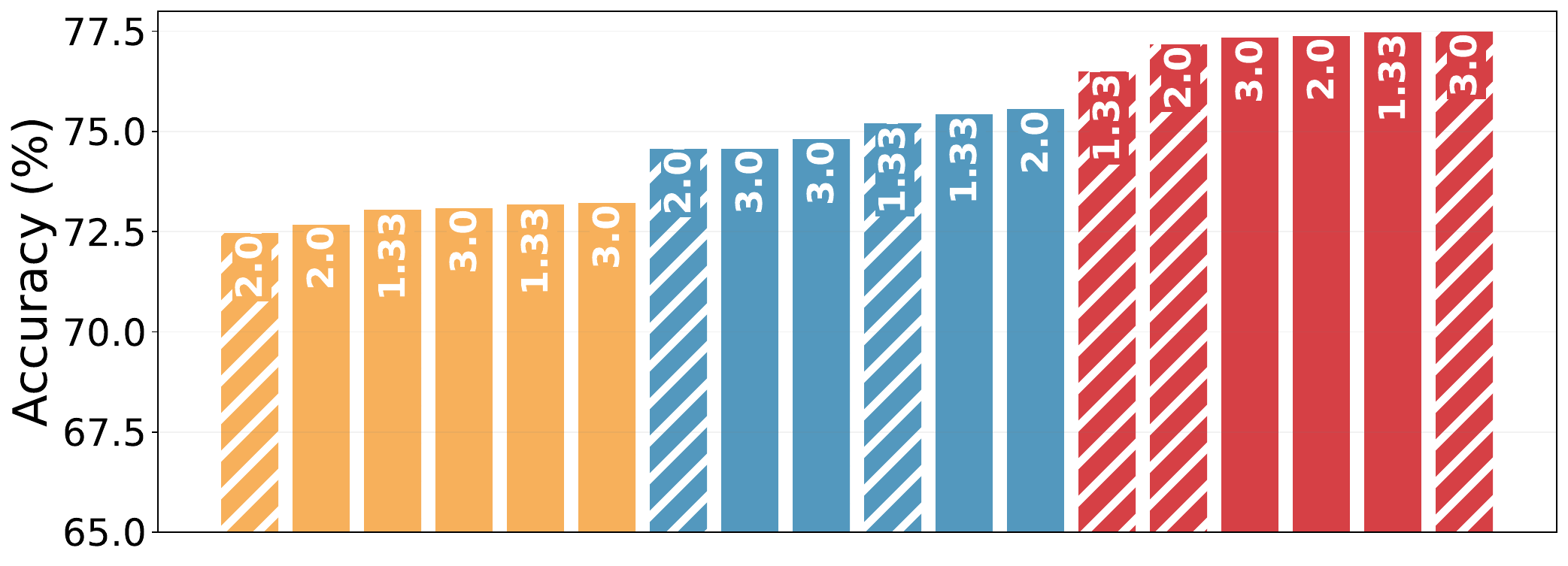}
    \caption{Sigmoid: init \(V_{\text{thr}}=1.0,\tau\in\{1.33,2.0,3.0\}\)}
  \end{subfigure}
  \caption{\textbf{Accuracy under different initial values of \(\tau\) and \(V_{\text{thr}}\) when learning both parameters, using the \emph{Arctan} and \emph{Sigmoid} surrogate function.} \textbf{(Top)}: The results using Arctan surrogate function. \textbf{(Bottom)}: The results using Sigmoid surrogate function. We compare AS-SG, RS-SG, and \textbf{TrSG} on a ResNet-19 (CIFAR100, 4 timesteps), evaluating both soft- and hard-reset LIF neurons.}
  \label{fig:acc_different_atan}
\end{figure}

\subsection{Validation on More Surrogate Functions}
\label{subsec:validation_of_atan}

In \cref{sec:second_method}, we introduced \textbf{TrSG (Threshold-robust Surrogate Gradient)} using a Rectangular-shaped surrogate as our primary example. In \cref{subsec:analysis_of_ta}, we extended experiments to include a Triangular-shaped surrogate. Here, we further demonstrate TrSG’s generality by evaluating it with an Arctan-based~\cite{fang:2021} and Sigmoid-based~\cite{wu:2018} surrogate functions.

\paragraph{Arctan Surrogate Function}
For an Arctan-based neuron, the derivative of the surrogate firing function is expressed as:
\[
    \frac{\partial f_{sr}(x)}{\partial x}
    \;=\; 
    \frac{\gamma}{\,2 \Bigl(1 + \bigl(\tfrac{\pi}{2}\,\gamma\,x\bigr)^{2}\Bigr)\!}.
\]

\paragraph{Sigmoid Surrogate Function}
For a Sigmoid-based neuron, the derivative of the surrogate firing function is expressed as:
\[
    \frac{\partial f_{sr}(x)}{\partial x}
    \;=\; 
    \frac{1}{\gamma}\frac{e^{\frac{x}{\gamma}}}{(1+e^{\frac{x}{\gamma}})^2}.
\]

where \(\gamma\) is the steepness parameter. Similar to our earlier analyses, we compare three SG methods (AS-SG, RS-SG, and \textbf{TrSG}) under both soft and hard reset neuron models.

\cref{fig:acc_different_atan} shows that \textbf{TrSG} consistently outperforms the conventional \textbf{absolute-scale} (AS-SG) and \textbf{relative-scale} (RS-SG) approaches, regardless of the initial threshold or time constant. This robust behavior parallels our findings in \cref{subsec:analysis_of_ta} for Rectangular and Triangular surrogates. TrSG maintains reliable performance across a wide range of initial conditions, emphasizing its capacity to provide stable gradient flow even for varied neuron parameters.

\subsection{Surrogate Gradient Through \texorpdfstring{\(V_{\text{thr}}^l\) and \(\tau^l\)}{}}
\label{subsec:grad_vthr_tau}

In the main text, our analysis of threshold-related instabilities centered primarily on 
\[
\frac{\partial S^l[t]}{\partial M^l[t]},
\]
i.e., how the spike output changes with respect to the membrane potential. 
However, similar issues arise when considering partial derivatives w.r.t.\ \(V_{\text{thr}}^l\) and \(\tau^l\). 
Below, we illustrate how AS-SG and RS-SG can each fail to provide stable gradients in these pathways and then show how \textbf{TrSG} maintains a consistent and well-scaled gradient flow.

\paragraph{1. Gradient w.r.t.\ \texorpdfstring{\(V_{\text{thr}}^l\)}{Vthr}.}
Recall from \cref{eq:lif_discrete_firing} that
\(
S^l[t] = \mathcal{H}(\,M^l[t] - V_{\text{thr}}^l\,)
\).
When approximating \(\mathcal{H}\) by a differentiable surrogate \(f_{sr}\), existing methods define the argument \(x\) in either \textbf{absolute-scale} or \textbf{relative-scale} form:

\begin{itemize}
    \item \textbf{AS-SG (Absolute-Scale).}  
    \(
    x = M^l[t] - V_{\text{thr}}^l.
    \)
    Then, 
    \[
    \frac{\partial S^l[t]}{\partial V_{\text{thr}}^l}
    \;=\;
    f_{sr}'(x)\,\times \Bigl(\tfrac{\partial M^l[t]}{\partial V_{\text{thr}}^l} - 1\Bigr),
    \]
    whose update does not adapt its magnitude to \(V_{\text{thr}}^l\). Consequently, if \(|V_{\text{thr}}^l|\) becomes very large or very small, the gradient can vanish or dominate excessively.

    \item \textbf{RS-SG (Relative-Scale).}  
    \(
    x = \tfrac{M^l[t]}{V_{\text{thr}}^l} - 1.
    \)
    By the chain rule,
    \[
    \frac{\partial S^l[t]}{\partial V_{\text{thr}}^l}
    \;\approx\;
    f_{sr}'\!\bigl(\tfrac{M^l[t]}{V_{\text{thr}}^l}-1\bigr)\,\times\,\tfrac{\partial}{\partial V_{\text{thr}}^l}\!\Bigl(\tfrac{M^l[t]}{V_{\text{thr}}^l}-1\Bigr),
    \]
    which can introduce factors proportional to \(\tfrac{1}{V_{\text{thr}}^l}\). If \(V_{\text{thr}}^l\) is tiny, the gradient risk is explosion; if \(V_{\text{thr}}^l\) is huge, updates can vanish.
\end{itemize}

\textbf{TrSG} circumvents both extremes by using a relative-scale definition for \(x\) and multiplying \(V_{\text{thr}}^l\) forward during training:
\[
O^l[t] \;=\; V_{\text{thr}}^l\,S^l[t].
\]
As discussed in the main text, this multiplication cancels out the problematic \(\tfrac{1}{V_{\text{thr}}^l}\) factor during backpropagation, keeping gradients w.r.t.\ \(V_{\text{thr}}^l\) stable throughout training.

\paragraph{2. Gradient w.r.t.\ \texorpdfstring{\(\tau^l\)}{tau}.}
Next, consider the discrete LIF update from \cref{eq:lif_discrete_potential}:
\(
M^l[t]
\;=\;
\Bigl(1 - \tfrac{1}{\tau^l}\Bigr)\,U^l[t-1]
\;+\;
\tfrac{1}{\tau^l}\,I_{\text{in}}^l[t]
\).
Since \(\tau^l\) appears in denominators, it affects \(S^l[t]\) indirectly via \(M^l[t]\). Hence,
\[
\frac{\partial S^l[t]}{\partial \tau^l}
\;=\;
\frac{\partial S^l[t]}{\partial M^l[t]}
\;\times\;
\frac{\partial M^l[t]}{\partial \tau^l}.
\]
If \(\tfrac{\partial S^l[t]}{\partial M^l[t]}\) is not well-scaled (as in standard AS-SG or RS-SG), the \(\tfrac{\partial S^l[t]}{\partial \tau^l}\) may yield exploding or vanishing updates. 
\textbf{TrSG}, by ensuring a stable gradient path for \(\tfrac{\partial S^l[t]}{\partial M^l[t]}\), similarly prevents instability when learning \(\tau^l\).

In conclusion, TrSG’s threshold-multiplication mechanism and relative-scale formulation ensure that gradients remain appropriately scaled not only w.r.t.\ \(M^l[t]\) but also along the \(V_{\text{thr}}^l\) and \(\tau^l\) pathways. This provides a broad foundation for stable, end-to-end training of spiking neurons, even when multiple internal parameters are learnable.

\section{Additional Experimental Results}
\label{sec:suppl_additional_exp}

\begin{figure}[h]
  \centering
  \begin{subfigure}{0.99\linewidth}
    \centering
    \includegraphics[width=0.99\linewidth]{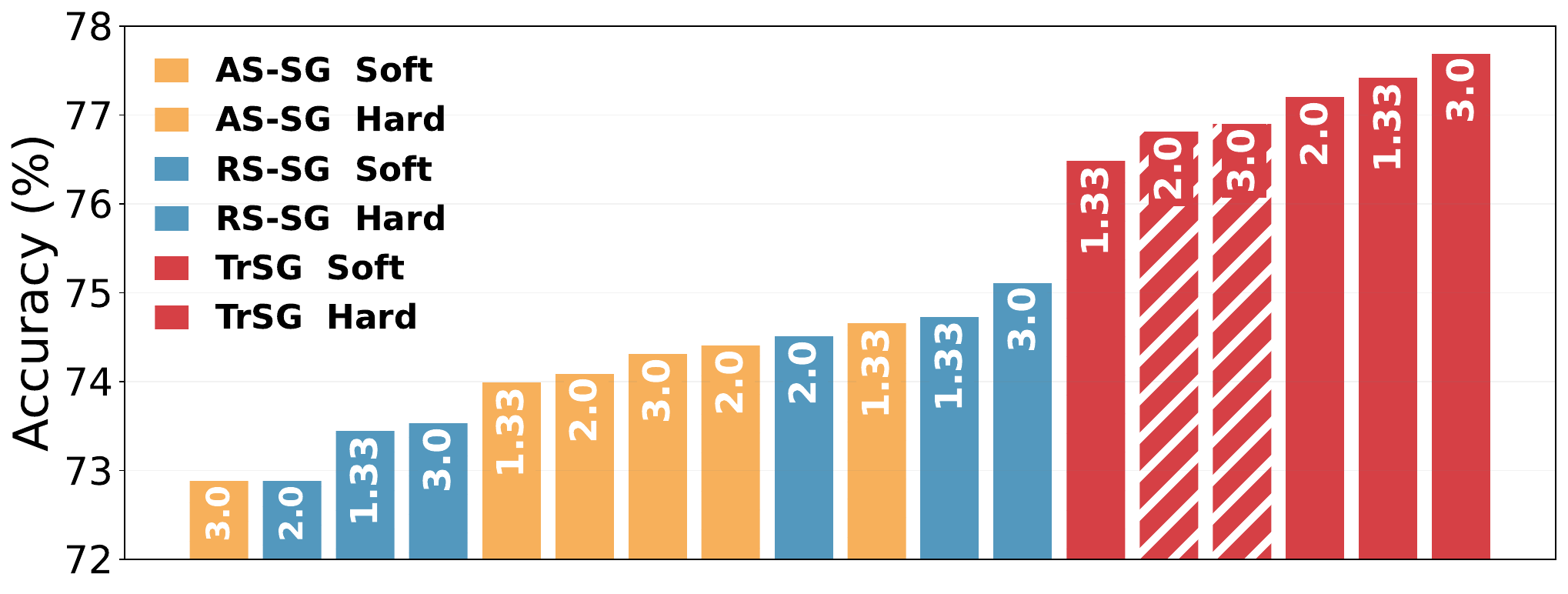}
    \caption{Rectangular-shaped Surrogate Gradient}
    \label{fig:acc_different_tau_rect}
  \end{subfigure}
  \vspace{0.4em}
  \begin{subfigure}{0.99\linewidth}
    \centering
    \includegraphics[width=0.99\linewidth]{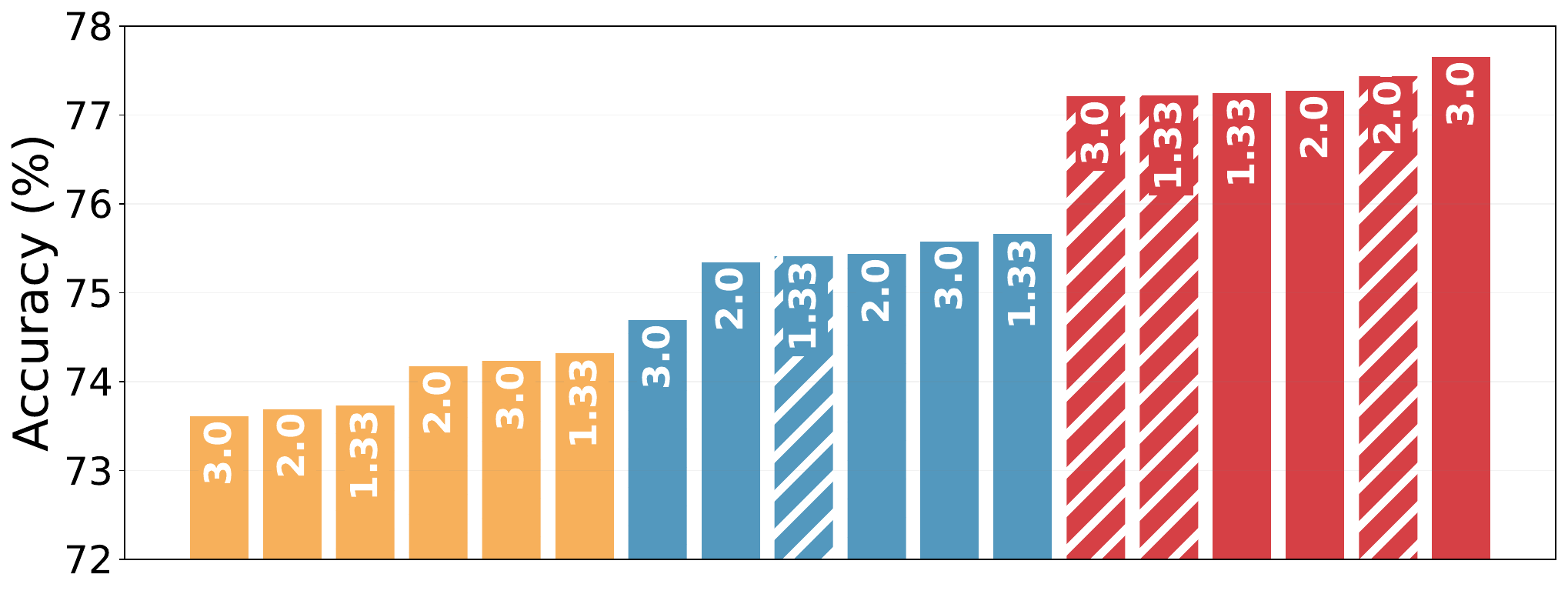}
    \caption{Triangular-shaped Surrogate Gradient}
    \label{fig:acc_different_tau_tri}
  \end{subfigure}
  \caption{
    \textbf{Accuracy results with different initial \(\tau\) values with initial \(V_{\text{thr}} = 1.0\).} We compare three SG types \textbf{(i) AdSG} \textbf{(i) RdSG} \textbf{(i) TrSG} under two shapes
    \textbf{(a)}~Rectangular and \textbf{(b)}~Triangular, each with both soft and hard resets. We vary initial \(\tau \in \{1.33,2.0,3.0\}\), indicated in the bars, and train ResNet-19 on CIFAR100 using 4 timesteps.
  }
  \label{fig:acc_different_tau_rect_tri}
\end{figure}

\subsection{Varying Initial \texorpdfstring{\(\tau\)}{} with a Fixed Initial \texorpdfstring{\(V_{\text{thr}}\)}{}}
\label{subsec:tau_var_supp}

To further complement the experiments in \cref{subsec:analysis_of_ta}, 
we conduct additional tests where we fix the initial threshold \(\,V_{\text{thr}}=1.0\) and instead vary the initial time constant \(\,\tau \in \{1.33,\,2.0,\,3.0\}\).  In \cref{fig:acc_different_tau_rect_tri}, the top panel 
(\cref{fig:acc_different_tau_rect}) reports accuracy under the Rectangular-shaped surrogate gradient, while the bottom panel (\cref{fig:acc_different_tau_tri}) corresponds to the Triangular-shaped surrogate. Each panel compares three SG types (AdSG, RdSG, and TrSG) and two reset modes (soft vs.\ hard). 

Our proposed \textbf{TrSG} consistently outperforms AS-SG and RS-SG across all initial \(\tau\) values. TrSG maintains high accuracy in every setting, demonstrating its robust learning dynamics and further validating the overall trend observed in \cref{subsec:analysis_of_ta}.

\begin{table}[t]
    \centering
    \begin{tabular}{ccc}
        \toprule
        \(\tau\) & \(V_{\text{thr}}\) & Accuracy (\%) \\
        \midrule
        \ding{55} & \ding{55} & 75.56 \\
        \ding{51} & \ding{55} & 74.61 \\
        \ding{55} & \ding{51} & 76.98 \\
        \ding{51} & \ding{51} & 77.22 \\
        \bottomrule
    \end{tabular}
    \caption{Ablation study of trainable parameters on ResNet-19 trained on CIFAR-100 with a timestep of 4. The initial (or fixed) values for \(V_{\text{thr}}\) and \(\tau\) are set to 1.0 and 2.0, respectively. \(\checkmark\) indicates trainable, while \(\text{\texttimes}\) means not trainable.}
    \label{tab:ablation_study}
\end{table}

\subsection{Impact of Trainable Parameters on Accuracy}

To analyze the impact of training the threshold potential \(V_{\text{thr}}\) and the time decay constant \(\tau\), we conduct an ablation study. The results are summarized in \cref{tab:ablation_study}.

From \cref{tab:ablation_study}, we observe that training only \(\tau\) results in a slight degradation in accuracy compared to fixed parameters. In contrast, training \(V_{\text{thr}}\) alone leads to a noticeable improvement in accuracy. Moreover, training \(\tau\) alongside \(V_{\text{thr}}\) yields the best results, with a final accuracy of 77.22\%. 

This indicates that while training \(\tau\) alone may not be beneficial, its effect becomes synergistic when combined with \(V_{\text{thr}}\) training. By allowing both parameters to be learned, the model can dynamically adjust them to achieve a balance between convergence speed and network stability, thereby enhancing performance.

\begin{figure}[h]
  \centering
  \includegraphics[width=0.99\linewidth]{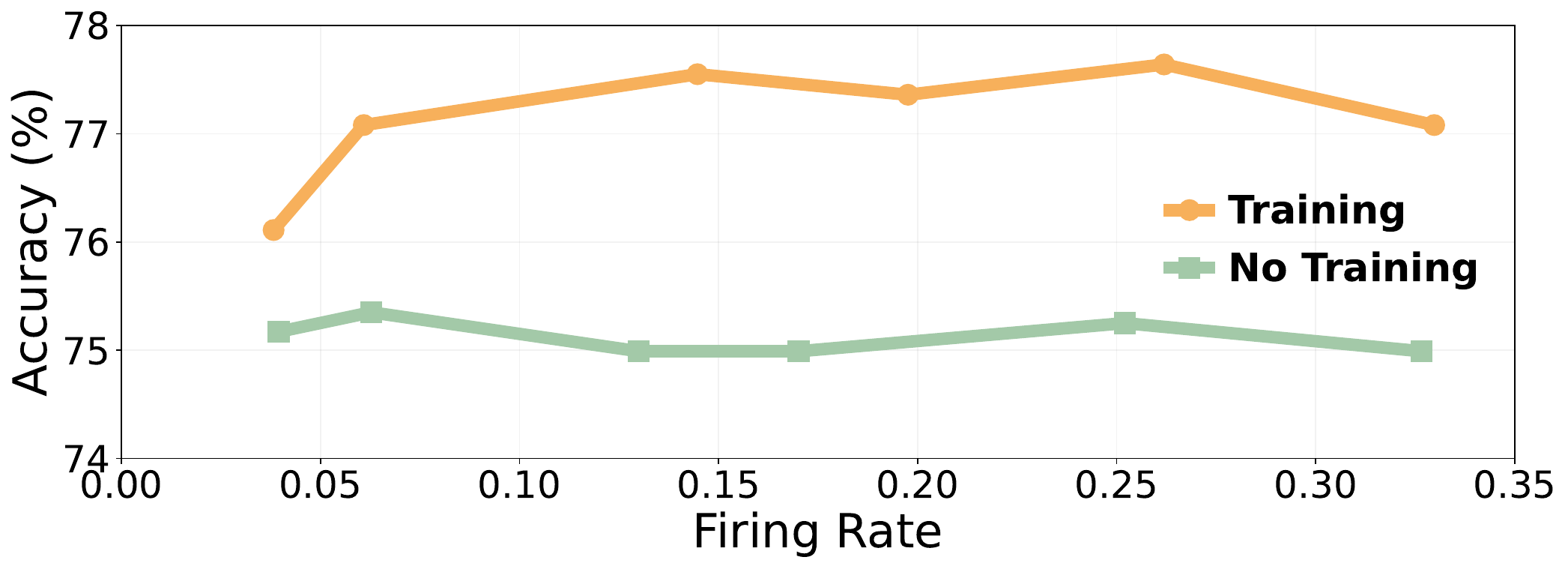}
  \caption{\textbf{Accuracy vs.\ firing rate under TrSG.}
  We sweep a sparsity loss~\cite{yan:2022} to match global firing rates and compare two settings:
  \emph{Training} (learn $V_{\text{thr}}$ and $\tau$ with TrSG) vs.\ \emph{No Training} (both frozen).}
  \label{fig:firing_rate_vs_accuracy_trsg}
\end{figure}

\subsection{Accuracy vs.\ Firing Rate}
\label{sec:acc_vs_firing_rate_trsg}

As shown in \cref{subsec:analysis_of_mpinit}, MP‑Init forms a Pareto front over firing rate.
A natural question is whether training internal parameters ($V_{\text{thr}}$, $\tau$) with TrSG merely increases spikes to boost accuracy.
To decouple these effects, we vary a sparsity regularization loss~\cite{yan:2022} that directly controls spike counts and evaluate accuracy across matched firing rates, comparing \emph{Training} (TrSG updates $V_{\text{thr}}$, $\tau$) to \emph{No Training} (both frozen).
As seen in \cref{fig:firing_rate_vs_accuracy_trsg}, the \textbf{Training} curve strictly dominates the \textbf{No Training} curve at all firing rates, indicating that the improvement stems from well‑adapted neuron dynamics rather than excessive spikes.
This is consistent with the threshold trajectories in \cref{fig:vthr_dynamics}, where layers evolve toward operating points that favor stable, informative spikes. 

\begin{table}[h]
\centering
\footnotesize
\begin{tabular}{ccccc}
\toprule
\textbf{MP-Init} & \textbf{TrSG} & \textbf{ACs / MACs(G)} & \textbf{Energy (mJ)} & \textbf{Acc (\%)} \\
\midrule
\ding{55} & \ding{55} & 1.03 / 0.14 & 1.57 & 75.58 \\
\ding{51} & \ding{55} & 1.01 / 0.14 & 1.57 & 76.20 \\
\ding{55} & \ding{51} & 1.13 / 0.14 & 1.68 & 77.13 \\
\ding{51} & \ding{51} & 1.12 / 0.14 & 1.67 & 77.68 \\
\bottomrule
\end{tabular}
\caption{\textbf{Energy at comparable firing rates.}
We report accumulate operations (ACs), multiply–accumulate operations (MACs), and the calculated energy costs.}
\label{tab:energy_calculation}
\end{table}

\subsection{Energy Cost}
Since firing rate tightly correlates with ACs (and thus energy), the analysis in \cref{subsec:analysis_of_mpinit,sec:acc_vs_firing_rate_trsg} can be viewed as an energy–accuracy trade‑off as well.
To quantify this, we measure ACs and MACs and compute energy following \cite{energy}; the results in \cref{tab:energy_calculation} show that, at similar firing rates (hence similar energy), training internal neuron parameters with TrSG yields consistently higher accuracy.
In short, TrSG improves accuracy at nearly the same energy cost, reinforcing that the gains come from better adaptation of $V_{\text{thr}}$ and $\tau$, not from increasing spike counts.

\begin{figure*}[ht]
  \centering
  \begin{subfigure}{0.995\linewidth}
      \begin{subfigure}{0.49\linewidth}
        \centering
        \includegraphics[width=0.99\linewidth]{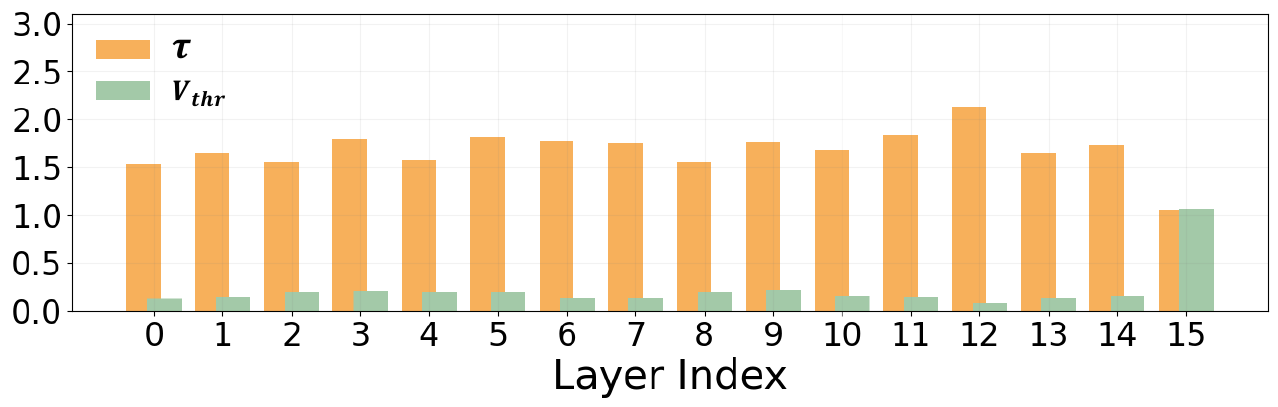}
      \end{subfigure}
      \begin{subfigure}{0.49\linewidth}
        \centering
        \includegraphics[width=0.99\linewidth]{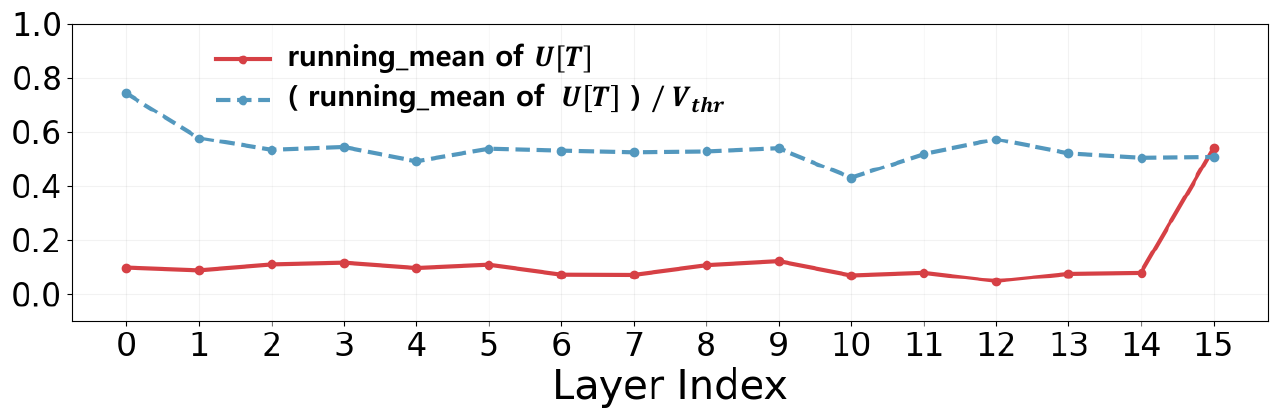}
      \end{subfigure}
      \caption{CIFAR100 (soft reset)}
      \label{fig:final_values_cifar100_softreset}
  \end{subfigure}
  \begin{subfigure}{0.995\linewidth}
      \begin{subfigure}{0.49\linewidth}
        \centering
        \includegraphics[width=0.99\linewidth]{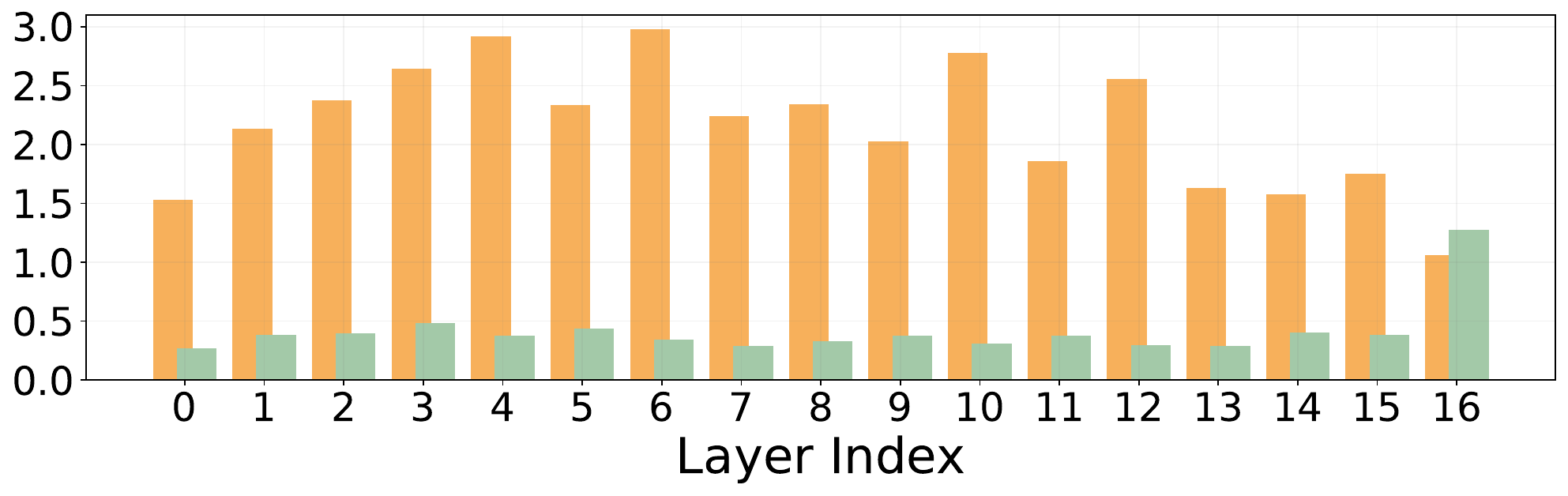}
      \end{subfigure}
      \begin{subfigure}{0.49\linewidth}
        \centering
        \includegraphics[width=0.99\linewidth]{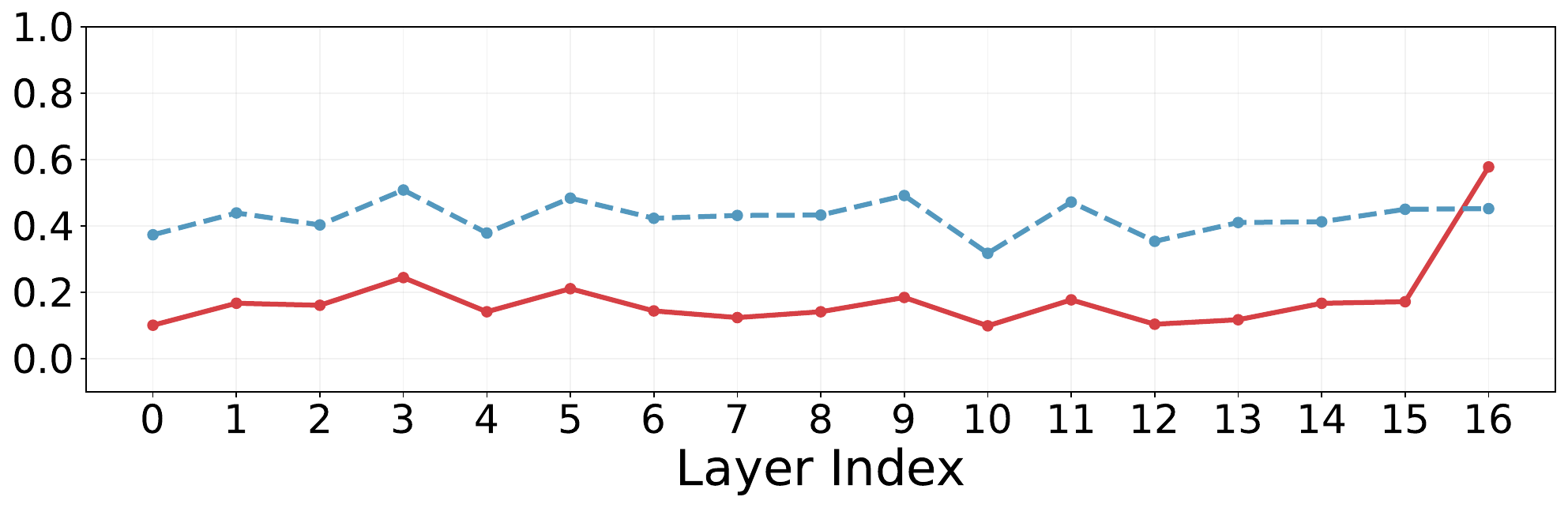}
      \end{subfigure}
      \caption{DVS-CIFAR10 (soft reset)}
      \label{fig:final_values_dvscifar10_softreset}
  \end{subfigure}
  \begin{subfigure}{0.995\linewidth}
      \begin{subfigure}{0.49\linewidth}
        \centering
        \includegraphics[width=0.99\linewidth]{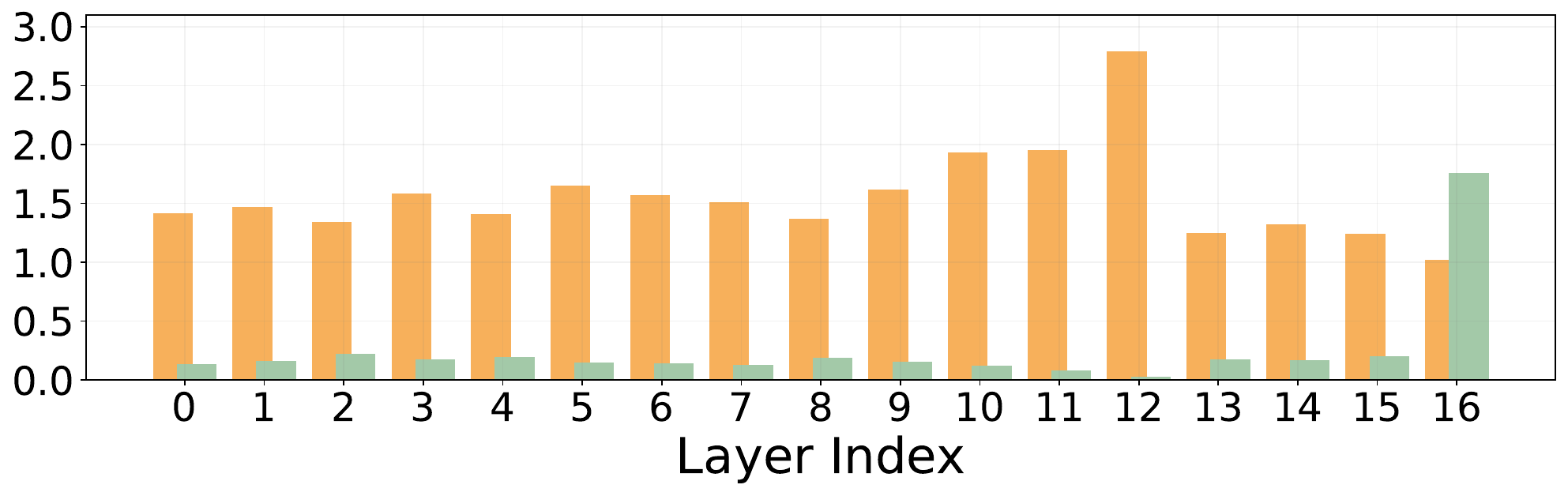}
      \end{subfigure}
      \begin{subfigure}{0.49\linewidth}
        \centering
        \includegraphics[width=0.99\linewidth]{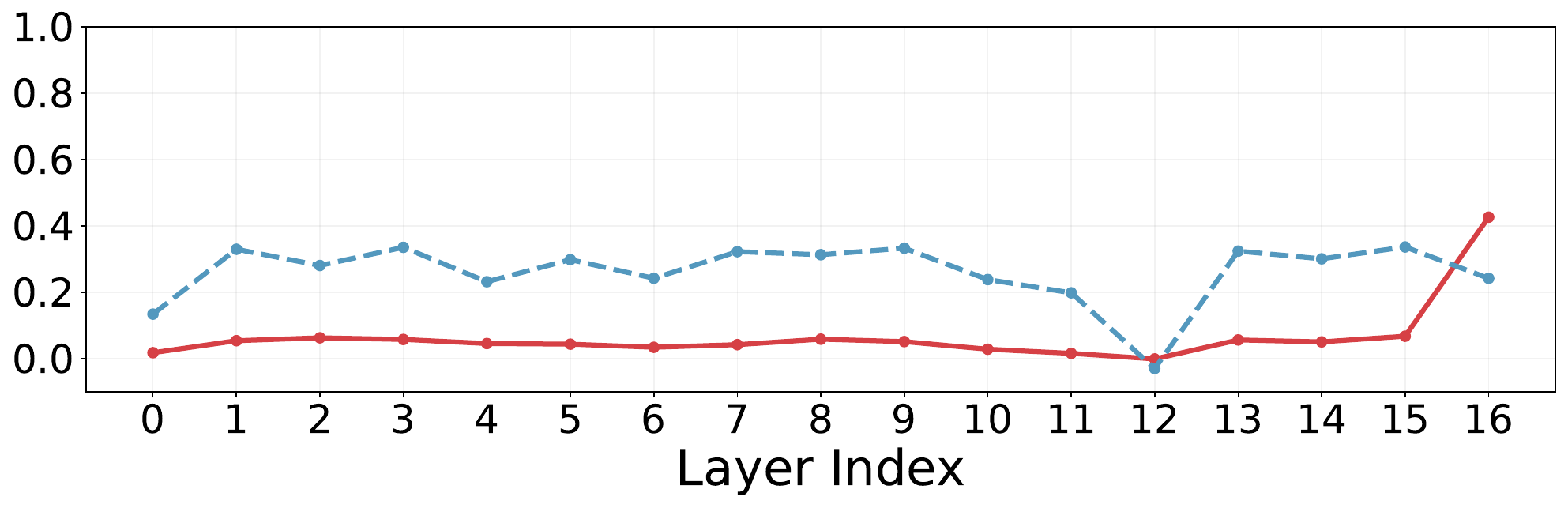}
      \end{subfigure}
      \caption{CIFAR100 (hard reset)}
      \label{fig:final_values_cifar100_hardreset}
  \end{subfigure}
  \begin{subfigure}{0.995\linewidth}
      \begin{subfigure}{0.49\linewidth}
        \centering
        \includegraphics[width=0.99\linewidth]{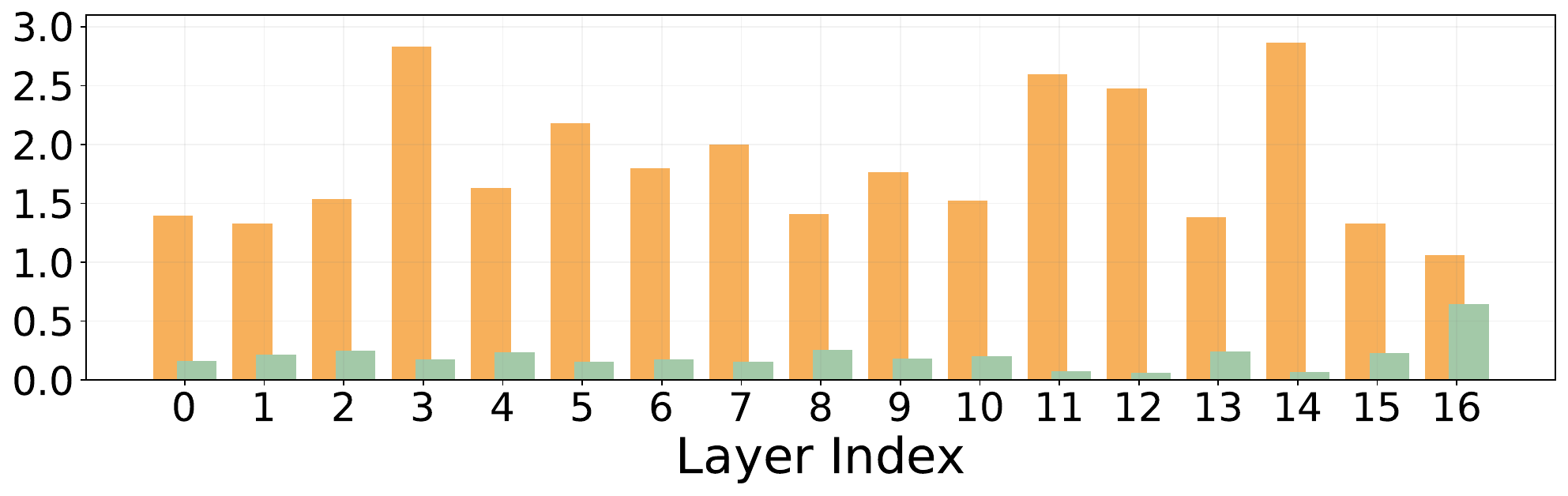}
      \end{subfigure}
      \begin{subfigure}{0.49\linewidth}
        \centering
        \includegraphics[width=0.99\linewidth]{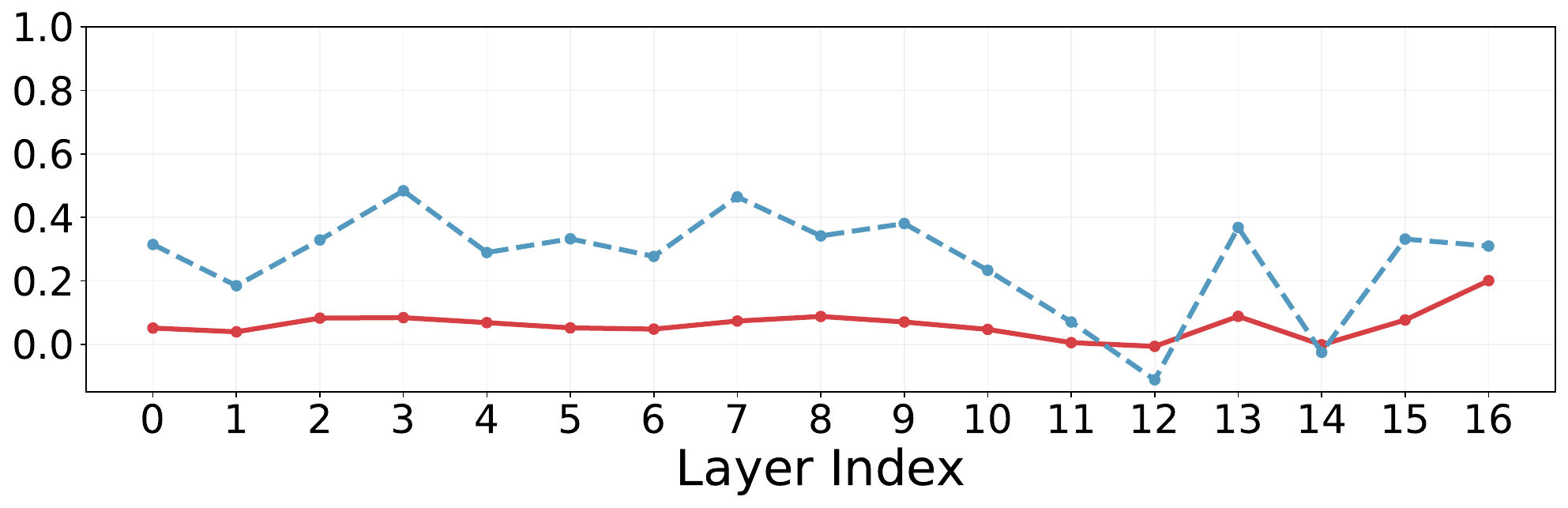}
      \end{subfigure}
      \caption{DVS-CIFAR10 (hard reset)}
      \label{fig:final_values_dvscifar10_hardreset}
  \end{subfigure}
  \caption{The final values of \(V_{\text{thr}}\), \(\tau\), and the running mean of the membrane potential at the last timestep (\(U[T]\)) after training. ResNet-19 is trained on CIFAR100 (timestep = 4) and DVS-CIFAR10 (timestep = 10).}
  \label{fig:final_values}
\end{figure*}

\subsection{Final Values of Neuron Parameters}
\label{subsec:final_values_of_parameters}

In addition to learning the network weights, our approach also optimizes the neuron parameters \(\tau\) and \(V_{\text{thr}}\), while MP-Init tracks an internal running mean of the membrane potential at the last timestep, \(U[T]\). To investigate how these three quantities evolve through training, we visualize their final layer-wise values for both CIFAR100 and DVS-CIFAR10 in \cref{fig:final_values}, considering soft vs.\ hard resets.

\paragraph{Running Mean of \(U[T]\).}
The running mean of \(U[T]\) typically settles at about half of \(V_{\text{thr}}\) in the soft-reset case (\cref{fig:final_values_cifar100_softreset,fig:final_values_dvscifar10_softreset}) and roughly a quarter of \(V_{\text{thr}}\) for hard-reset neurons (\cref{fig:final_values_cifar100_hardreset,fig:final_values_dvscifar10_hardreset}). 
This pattern indicates that the average post-synaptic potential remains below the learned threshold in most layers yet remains sufficiently high to generate spikes as needed.

\paragraph{Threshold \(\boldsymbol{V_{\text{thr}}}\).}
Across both datasets and reset types, the learned \(V_{\text{thr}}\) commonly falls below 0.5 in earlier and intermediate layers, suggesting that moderate thresholds facilitate stable spiking activity.

\paragraph{Decay Constant \(\boldsymbol{\tau}\).}
We observe a consistent trend of \(\tau\) converging around 1.5--2.0 in most layers for CIFAR100, while on DVS-CIFAR10, \(\tau\) tends to reach even larger values (e.g., near 2.5--3.0). 
This discrepancy implies that dynamic event-driven data may benefit from neurons accumulating input more slowly, whereas CIFAR100’s static images require a shorter effective integration time.

\paragraph{Last-Layer Behavior.}
Interestingly, in the final layer on both datasets, \(\tau\) commonly drops close to 1, which makes it become a binary layer. The layer may act as a gating mechanism, helping the model to classify features decisively without excessive temporal integration.

\begin{figure}[h]
  \centering
  \begin{subfigure}{0.99\linewidth}
    \includegraphics[width=0.99\linewidth]{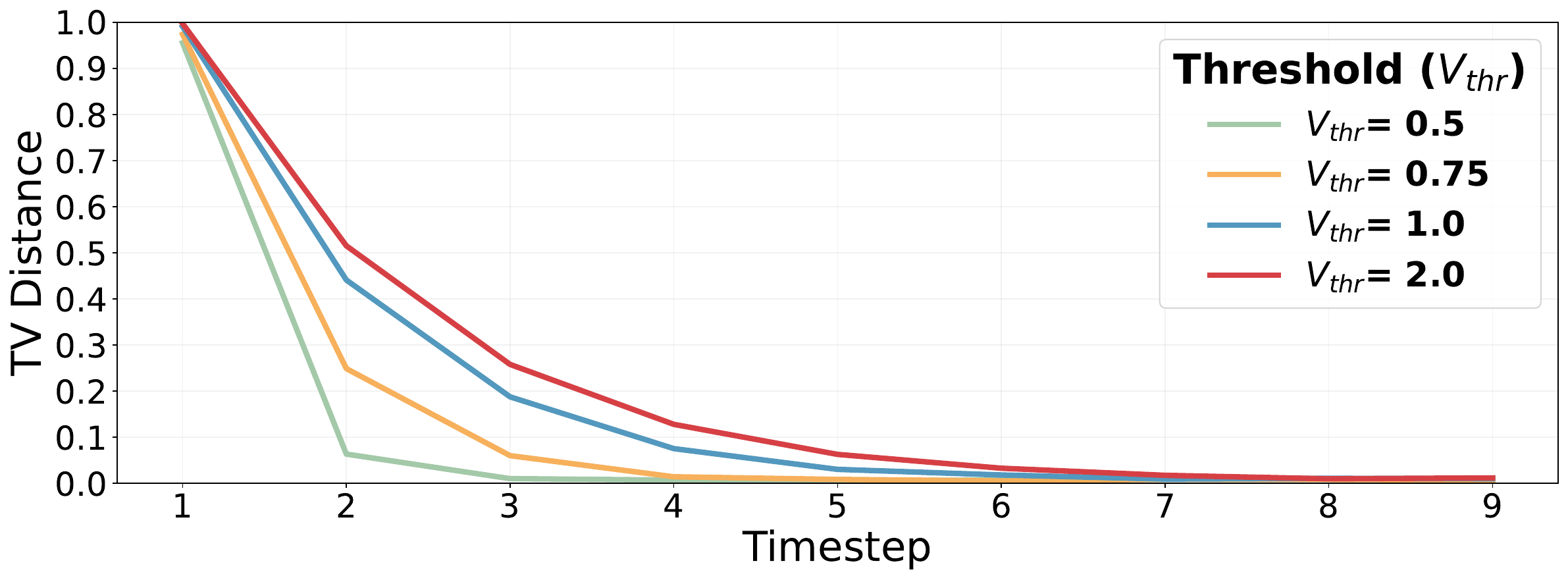}
    \caption{Impact of varying \(V_{\text{thr}}\) on convergence speed.}
  \end{subfigure}
  \hfill
  \begin{subfigure}{0.99\linewidth}
    \includegraphics[width=0.99\linewidth]{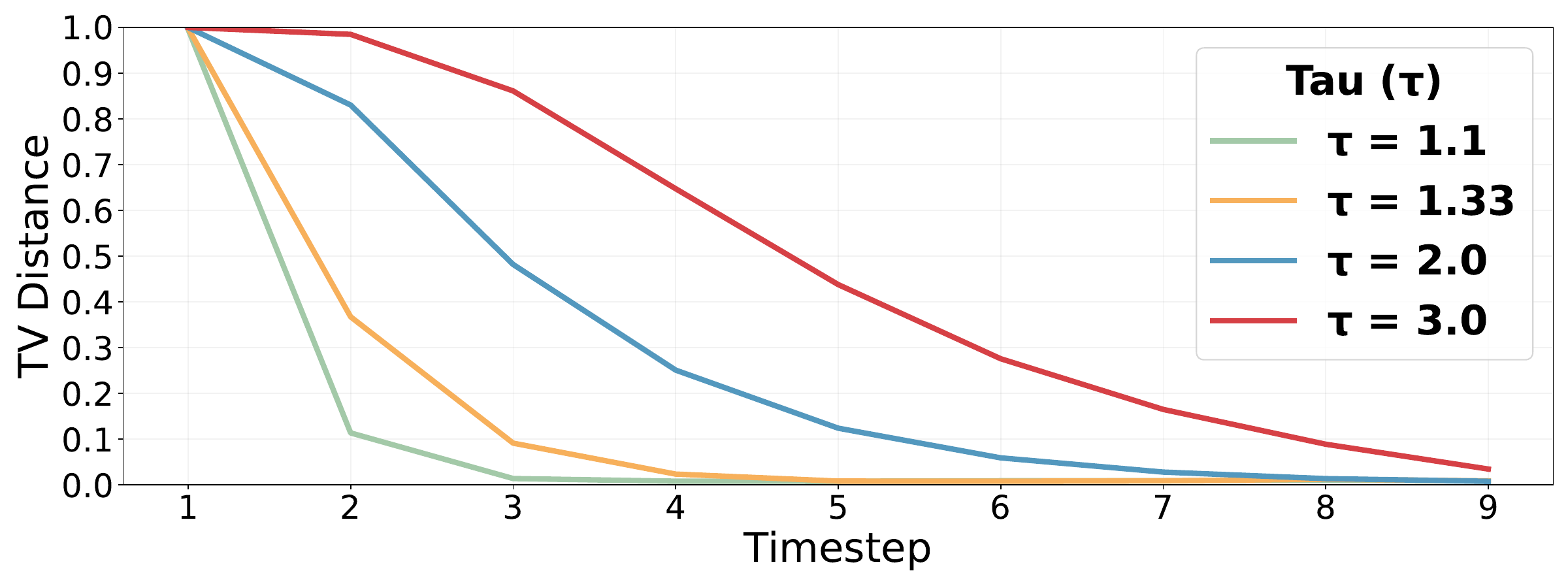}
    \caption{Impact of varying \(\tau\) on convergence speed.}
  \end{subfigure}
  \caption{TV distance of membrane potential distribution between timestep 10 and previous timesteps, illustrating the effects of altering the threshold \(V_{\text{thr}}\) (Left) or the time decay constant \(\tau\) of the LIF neuron (Right). The input is randomly generated from a Gaussian distribution.}
  \label{fig:tv_norm}
\end{figure}

\subsection{Convergence Rate of Membrane Potential}

To analyze the factors influencing the convergence rate described in Theorem 1, we conduct a mathematical analysis complemented by simple experiments, revealing that the time decay constant \(\tau\) and the threshold voltage \(V_{\text{thr}}\) play pivotal roles in determining the convergence speed, thereby influencing the TCS problem.

As outlined in \cref{eq:minimorization_condition} and \cref{eq:convergence_speed}, Doeblin’s Minorization Condition ensures not only the convergence of distributions but also provides insights into the convergence speed. Consider the uniform measure \(\mu\), which represents a uniform distribution over the interval \([u^{-}, u^{+}]\). For simplicity, let \(U[t-1] = 0\), such that \(U[t] = \frac{1}{\tau} I_{\text{in}}[t]\). Under these assumptions, \cref{eq:minimorization_condition} can be expressed as:

\begin{equation}
    P\left(U[t] = \frac{1}{\tau} I_{\text{in}}[t] \in A\right) \geq \epsilon \mu(A),
\end{equation}

From this expression with the Assumption. 1, it becomes evident that decreasing \(\tau\) increases the maximum permissible value of \(\epsilon\), which, in turn, accelerates the convergence speed as per \cref{eq:convergence_speed}. Similarly, reducing the threshold \(V_{\text{thr}}\) results in the membrane potential reaching the threshold more frequently, triggering resets. This frequent resetting narrows the range \([u^{-}, u^{+}]\) of the membrane potential, thereby increasing \(\epsilon\) and further accelerating convergence.

We conduct experiments utilizing soft reset LIF neurons with Gaussian input to validate the theoretical findings. The parameters \(V_{\text{thr}}\) and \(\tau\) are varied, and the TV distance~\cite{rosenthal:1995} is computed between the distribution at each timestep and the final distribution, as illustrated in \cref{fig:tv_norm}. The results confirm that smaller values of \(V_{\text{thr}}\) and \(\tau\) can alleviate the TCS problem. However, it is crucial to note that setting these parameters to excessively low values is not universally advantageous.

For instance, a threshold \(V_{\text{thr}}\) that is too low can result in excessive firing rates, leading to unstable network dynamics. Similarly, an overly small \(\tau\) degrades the neuron’s ability to accumulate membrane potential over time, effectively reducing its behavior to binary quantization. Therefore, the selection of \(V_{\text{thr}}\) and \(\tau\) requires a careful balance to achieve both rapid convergence and stable network dynamics.

This analysis might establish a connection between MP-Init and TrSG. Rather than relying on manual tuning, training \(V_{\text{thr}}\) and \(\tau\) as learnable parameters within the network can be more effective. As shown in \cref{subsec:final_values_of_parameters}, TrSG successfully optimizes \(V_{\text{thr}}\) and \(\tau\), leading them to moderately low values and achieving superior accuracy.

\end{document}